\documentclass[final,leqno,onefignum,onetabnum]{siamltex1213}

\usepackage{dsfont}
\usepackage{amssymb,amsmath}
\usepackage{graphicx}
\usepackage[export]{adjustbox}
\usepackage{mdframed}
\usepackage[draft]{fixme}

\usepackage[siims]{optional}
\usepackage{color}

\newcommand{\mn}[1]{{\color{black}{#1}}}

\begin{document}


\newcommand{\cf}{\S}
\newcommand\eg{\textit{e.g.~}}
\newcommand\etal{\textit{et~al.~}}
\newcommand\ie{\textit{i.e.}}
\newcommand\viz{\textit{viz.~}}
\newcommand{\App}[1]{Appendix~\ref{#1}}
\newcommand{\Algo}[1]{Algorithm~\ref{#1}}
\newcommand{\Eqn}[1]{Eqn.~\eqref{#1}}
\newcommand{\eqn}[1]{eqn.~\eqref{#1}}
\newcommand{\Chap}[1]{Chapter~\ref{#1}}
\newcommand{\Fig}[1]{Fig.~\ref{#1}}
\newcommand{\Sec}[1]{Section~\ref{#1}}
\newcommand{\Table}[1]{Table~\ref{#1}}

\newcommand\ba{\mathbf{a}}
\newcommand\bb{\mathbf{b}}
\newcommand\bB{\mathbf{B}}
\newcommand\bc{\mathbf{c}}
\newcommand\bC{\mathbf{C}}
\newcommand\bd{\mathbf{d}}
\newcommand\bD{\mathbf{D}}
\newcommand\be{\mathbf{e}}
\newcommand\bff{\mathbf{f}}
\newcommand\bg{\mathbf{g}}
\newcommand\bK{\mathbf{K}}
\newcommand\bm{\mathbf{m}}
\newcommand\bn{\mathbf{n}}
\newcommand\bp{\mathbf{p}}
\newcommand\br{\mathbf{r}}
\newcommand\bs{\mathbf{s}}
\newcommand\bt{\mathbf{t}}

\newcommand\bu{\mathbf{u}}
\newcommand\bU{\mathbf{U}}
\newcommand\bv{\mathbf{v}}
\newcommand\bw{\mathbf{w}}
\newcommand\bx{\mathbf{x}}
\newcommand\by{\mathbf{y}}
\newcommand\bz{\mathbf{z}}
\newcommand\balpha{\boldsymbol{\alpha}}
\newcommand\bbeta{\boldsymbol{\beta}}
\newcommand\bepsilon{\boldsymbol{\epsilon}}
\newcommand\blambda{{\boldsymbol{\lambda}}}
\newcommand\bLambda{\boldsymbol\Lambda}
\newcommand\bmu{\boldsymbol{\mu}}
\newcommand\bsigma{\boldsymbol{\sigma}}
\newcommand\bSigma{\boldsymbol{\Sigma}}
\newcommand\btheta{\boldsymbol{\theta}}

\newcommand\bbE{\mathbb{E}}
\newcommand\bbG{\mathbb{G}}
\newcommand\bbR{\mathbb{R}}
\newcommand\bbV{\mathbb{V}}
\newcommand\bbZ{\mathbb{Z}}

\newcommand\cA{\mathcal{A}}
\newcommand\cB{\mathcal{B}}
\newcommand\cD{\mathcal{D}}
\newcommand\cE{\mathcal{E}}
\newcommand\cF{\mathcal{F}}
\newcommand\cG{\mathcal{G}}
\newcommand\cI{\mathcal{I}}
\newcommand\cJ{\mathcal{J}}
\newcommand\cL{\mathcal{L}}
\newcommand\cM{\mathcal{M}}
\newcommand\cN{\mathcal{N}}
\newcommand\cO{\mathcal{O}}
\newcommand\cP{\mathcal{P}}
\newcommand\cT{\mathcal{T}}
\newcommand\cV{\mathcal{V}}
\newcommand\cU{\mathcal{U}}
\newcommand\cX{\mathcal{X}}
\newcommand\cY{\mathcal{Y}}
\newcommand\cZ{\mathcal{Z}}

\newcommand\dsE{\mathds{E}}
\newcommand\dsG{\mathds{G}}
\newcommand\dsH{\mathds{H}}
\newcommand\dsV{\mathds{V}}

\newcommand\rmA{\mathrm{A}}
\newcommand\rmB{\mathrm{B}}
\newcommand\rmd{\mathrm{d}}
\newcommand\rmD{\mathrm{D}}
\newcommand\rmE{\mathrm{E}}
\newcommand\rmF{\mathrm{F}}
\newcommand\rmH{\mathrm{H}}
\newcommand\rmI{\mathrm{I}}
\newcommand\rmJ{\mathrm{J}}
\newcommand\rmK{\mathrm{K}}
\newcommand\rmM{\mathrm{M}}
\newcommand\rmT{\mathrm{T}}
\newcommand\rmU{\mathrm{U}}
\newcommand\rmV{\mathrm{V}}
\newcommand\rmW{\mathrm{W}}
\newcommand\rmy{\mathrm{y}}
\newcommand\rmz{\mathrm{z}}
\newcommand\rmZ{\mathrm{Z}}

\newcommand{\argmax}{\operatornamewithlimits{argmax}}
\newcommand{\argmin}{\operatornamewithlimits{argmin}}
\newcommand{\distributionequal}{\operatornamewithlimits{\sim}}
\opt{arxiv}
{
  \newcommand{\diag}{\textrm{diag}}
  \newcommand{\const}{\textrm{const}}
  \newcommand\rank{{\mathrm{Rank}}}
  \newcommand{\myabstract}[1]{\abstract{#1}}
}
\opt{siims}
{ 
  \newcommand{\myabstract}[1]{\begin{abstract}#1\end{abstract}}
}

\newcommand{\KLD}[2]{{\mathds{KL}}[#1 || #2]}
\newcommand{\Cov}{\mathds{C}\text{ov}}
\newcommand{\Corr}{\mathds{C}\text{orr}}
\newcommand{\E}[1]{{\dsE\left\{#1\right\}}}
\newcommand{\Ent}[1]{{\dsH\left\{#1\right\}}}
\newcommand{\Exp}{\dsE}
\newcommand{\half}{\frac{1}{2}}
\newcommand\Hes{\nabla\nabla\tr}
\newcommand{\inv}{^{-1}}
\newcommand{\IP}[1]{{\left\langle{#1}\right\rangle}} 
\newcommand\I{\rmI}
\newcommand\ind{\mathds{1}}
\newcommand\KL{\mathds{KL}}
\newcommand\one{\mathds{1}}
\newcommand{\p}[1]{{p\left(#1\right)}}
\newcommand{\pinv}{^{-}}
\newcommand{\Prec}{\bLambda}
\newcommand{\Prr}{\mathrm{Pr}}
\newcommand\R{\bbR}
\newcommand{\tr}{^\top}
\newcommand{\Trace}{\mathrm{Tr}}
\newcommand\Var[1]{{\mathds{V}\mathrm{ar}\left\{#1\right\}}}
\newcommand\vol{\mathrm{vol}}


\newcommand{\TV}{\mathrm{TV}}
\newcommand{\img}{\mathrm{y}}
\renewcommand{\ij}{{i,j}}
\newcommand{\ioj}{{i+1,j}}
\newcommand{\ijo}{{i,j+1}}
\newcommand{\iojo}{{i+1,j+1}}
\newcommand{\x}{{x}}
\newcommand{\y}{{\rmy}}
\newcommand{\z}{{\rmz}}
\newcommand{\q}{{q}}
\newcommand{\Q}{{Q}}
\newcommand{\Py}{{P}}
\renewcommand{\E}[1]{{\dsE_\Q\left\{#1\right\}}}

\newcommand{\EV}[2]{{\dsE}_{#1}\left[#2\right]}
\newcommand{\PD}{~.}
\newcommand{\CM}{~,}
\newcommand{\zvar}{\rmz}
\newcommand{\area}{}
\newcommand{\wrt}{wrt.~}
\newcommand{\ase}{\overset{a.s.}{=}}
\newcommand\where{\text{where }}
\newcommand\thrfr{\text{therefore }}

\newcommand{\indx}{\mn{{i\in{\mathbb I}_{\cX}}}}

\graphicspath{{Figures/}{Figures/santner/}}

\title{Active Mean Fields for Probabilistic Image Segmentation: Connections with Chan-Vese and Rudin-Osher-Fatemi Models}
\author{Marc Niethammer\footnotemark[1] \and Kilian M. Pohl\footnotemark[2] \and Firdaus Janoos\footnotemark[3] \and \mbox{William M. Wells III}\footnotemark[4]}

\maketitle

\renewcommand{\thefootnote}{\fnsymbol{footnote}}

\footnotetext[1]{University of North Carolina at Chapel Hill, Department of Computer Science and Biomedical Research Imaging Center (BRIC), (\email{mn@cs.unc.edu}).}
\footnotetext[2]{Center for Health Sciences, SRI International and Department of Psychiatry and Behavioral Sciences, Stanford University, (\email{kilian.pohl@stanford.edu}).}
\footnotetext[3]{Two Sigma Investments, (\email{firdaus.janoos@gmail.com}).}
\footnotetext[4]{Harvard Medical School and Brigham and Women's Hospital, (\email{sw@bwh.harvard.edu}).}

\renewcommand{\thefootnote}{\arabic{footnote}}


\myabstract{
Segmentation is a fundamental task for extracting semantically meaningful regions from an image. \mn{The goal of segmentation algorithms} is to \mn{accurately} assign object labels to each image location. \mn{However,} image-noise, shortcomings of algorithms\mn{, and image ambiguities cause uncertainty in label assignment. \mn{Estimating the uncertainty in label assignment is} important in multiple application domains, such as segmenting tumors from medical images} for radiation treatment planning. \mn{One way to estimate these uncertainties is through the computation of posteriors of Bayesian models, which is computationally prohibitive for many practical applications}. On the other hand, \mn{most computationally efficient methods fail to estimate label uncertainty.} \mn{We therefore propose in this paper the Active Mean Fields (AMF) approach, a technique based on Bayesian modeling} that uses a mean-field approximation \mn{to efficiently compute a segmentation and its corresponding uncertainty}. \mn{Based on a variational formulation, the resulting convex model combines} any label-likelihood measure with a \mn{prior on the length of the segmentation boundary}. A specific implementation of that model is the Chan--Vese segmentation model (CV), \mn{in which the binary segmentation task is defined by a Gaussian likelihood and a prior regularizing the length of the segmentation boundary}. Furthermore, the Euler--Lagrange equations derived from the AMF model are equivalent to those of the popular Rudin-Osher-Fatemi (ROF) model for image denoising. Solutions to the AMF model can thus be implemented by directly utilizing highly-efficient ROF solvers on log-likelihood ratio fields. We \mn{qualitatively asses the approach on synthetic data as well as on real natural and medical images. For a quantitative evaluation, we apply our approach to the \texttt{icgbench} dataset.}}

\opt{arxiv}
{
  \setcounter{tocdepth}{10}
  \tableofcontents
}

\opt{siims}
{
  \begin{AMS}\end{AMS}
}

\section{Introduction}

Image segmentation approaches rarely provide measures of \mn{segmentation label uncertainty.} In fact, most existing and probabilistically-motivated segmentation approaches only compute \mn{the} \emph{maximum a posteriori} (MAP) \mn{solution}~\cite{kolmogorov2004,li2009,bresson2007,cremers2007review,paragios2002geodesic}. \mn{Using} these models to segment ambiguous boundaries is troublesome especially for applications where confidence in object  boundaries impacts analysis. \mn{For example, many radiation treatment plans base dose distribution on the boundaries of tumors segmented from medical images with low contrast~\cite{martin2015}. This can be problematic, as segmentation variability can have a substantial effect on radiation treatment; Martin et al.~\cite{martin2015} report that such variability \mn{caused mean observer} tumor control probability (\ie, the probability to control or eradicate a tumor at a given dose) to range from (22.6 $\pm$11.9)\% to (33.7 $\pm$ 0.6)\% \mn{between six participating physicians} in a study of intensity-modulated radiation therapy (IMRT) of 4D-CT-based non-small cell lung cancer radiotherapy.} The precision of the planning could 
be improved around highly-confident tumor boundaries~\mn{\cite{martin2015,jameson2014}} thereby reducing the risk of damaging healthy tissue in those areas. As significant information about label uncertainty is contained in the posterior {\it distribution}, it is natural to go beyond \mn{determining}  a MAP solution and instead to either compute the posterior distribution itself or a computationally efficient approximation.

This paper develops \mn{such} a method for an efficient approximation of the posterior distribution on labels. Furthermore, it connects this method to the Rudin-Osher-Fatemi (ROF) model for image-denoising~\cite{rudin1992nonlinear,vogel1996iterative,beck2009} and previously existing level-set segmentation approaches~\cite{osher2003}, in particular the Chan-Vese segmentation model~\cite{chan2000active}. Due to these connections we can (i) make use of the efficient solvers for the ROF model to approximate the posterior distribution on labels \mn{and} (ii) compute the solution to the Chan-Vese model through the MAP realization of our approximation to the posterior distribution\mn{, \ie}, our model is more general and subsumes the Chan-Vese model. In contrast to the implicit style of active-contour methods that represent labels by way of zero level-sets, \mn{such as the classical formulation of the Chan-Vese model}, we use a dense logit (``log odds''), representation of label probabilities. This is akin to the convex approaches for active contours~\cite{appleton2006}, but in a probabilistic formulation.

\subsection{Motivations}

Beyond optimal labelings, posterior distributions on labelings offer some advantages. For example, in many instances, one wishes to obtain information about segmentation confidence; or in change detection, distributions can help to determine whether an observed apparent change may be due to chance. Furthermore, probabilistic models on latent label fields can be useful for constructing more ambitious systems that, e.g., perform simultaneous segmentation and atlas registration~\cite{pohl2006bayesian}. \mn{However, the computation of posterior distributions is typically costly. Conversely, the computation of deterministic segmentation results, as for example by the popular active-contour approaches, is inexpensive and has shown to be an effective approach. Hence, we were motivated to merge both technologies, to obtain an active-contour inspired segmentation approach capable of estimating posterior distributions efficiently.} 

In previous work~\cite{PohlIPMI2007}, we described an {\em Active Mean Fields} (AMF) approach to image segmentation that used a variational mean field method (VMF) approach along with a logit representation to construct a segmentation system similar to the one described here. This method empirically \mn{generated accurate segmentations and converged well,}  \mn{but used a different, and more awkward, approximation of the expected value of the length functional. In this present work, we use a linearization approach via the Plefka approximation. Using this approximation has profound consequences as it allows to make connections to the Chan-Vese~\cite{chan2000active} segmentation model and the ROF denoising model~\cite{rudin1992nonlinear} in the continuous space. This connection in turn makes possible the efficient implementation of the AMF model through approaches used for ROF denoising. Hence, the overall model is convex, easy to implement and fast.} \mn{Furthermore,} we \mn{show good approximation properties in comparison to the ``exact'' distribution.}

\subsection{Contributions}
{

The main contributions of this article are:
\begin{itemize}
  \item It derives an \mn{AMF} approach \mn{that allows a computationally efficient estimation of the posterior distribution of the segmentation label map based on the} \mn{VMF} approximation for binary image segmentation \mn{regularized via} a  boundary length prior.
  \item It establishes strong connections between the proposed AMF model, active-contour models and total-variation (TV) denoising. In particular, the model retains the global optimality of convex active-contours while estimating a level-set function that has a direct interpretation as an approximate posterior on the segmentation. This is in contrast to level-set techniques which use the zero level-set only as a means for representing the object boundary with no (probabilistic) interpretation of the non-zero level-sets.
  \item It demonstrates how the Rudin-Osher-Fatemi (ROF) TV denoising model can be used to efficiently compute solutions of the AMF model. Hence, given the widespread availability of high-performance ROF-solvers, the AMF model is very simple to implement and will be immediately usable by the community with little effort.
\end{itemize}
}

\subsection{Background}

The earliest and simplest probabilistic image segmentation approaches frequently used pixel-wise independent Maximum Likelihood (ML) or MAP classifiers \cite{vannier1985multispectral}, that could be as simple as image thresholding.  Better performance, in the face of noise, motivated the use of regularization, or prior probability models on the label fields that discouraged fragmentation \cite{Besag1986}, leading to the wide-spread application of Markov Random Field (MRF) models \cite{Held-Wells-1997,zhang2001segmentation}.  Image segmentation with MRF models was initially thought to be computationally complex, which motivated approximations, including the {\em mean field} approach from statistical physics~\cite{kapur1998,chandler1987}.  Moreover, recently, fast solvers have appeared using graph-cuts, belief propagation or linear programming techniques \mn{that yield globally optimal solutions for certain} energy functions~\cite{Szeliski2008}.

Typically, the segmentation problem is posed as the minimization of an energy or negative \mn{log-likelihood} that incorporates an image likelihood and a regularization term on the boundaries of segmented objects. This regularization may be specified either:  (i) directly on the boundary (explicitly as a parametric curve or surface, or implicitly through the use of level-set functions); or (ii) by representing objects via indicator functions, where discontinuities in those functions identify boundaries. The direct boundary representation is attractive because it reduces complexity as only objects of co-dimension one need to be dealt with (\ie, a curve in 2D, a surface in 3D, etc.). The price for this reduction in complexity is that, usually, minimization becomes non-convex, and hence can get trapped in poor local minima in the absence of good initializations. In the {\em snakes} approach~\cite{kass1988}, a popular example of explicit boundary representation, the boundary curve represented by 
\mn{control points} is evolved such that it captures the object of interest (for example, by getting attracted to edges) while assuring regularity of the boundary by penalizing rapid boundary changes through elasticity and rigidity terms. Although computationally efficient, explicit parametric representations cannot easily deal with topological changes and have numerical issues due to their fixed object parameterization (\eg when an initial curve grows or shrinks drastically). Furthermore, though not an intrinsic problem of explicit parameterizations, such methods are typically not geometric, making evolutions dependent on curve parameterizations.

In contrast, level-set representations~\cite{osher2003,malladi1995shape} \mn{of} active-contour methods~\cite{caselles1997,kichenassamy1995} do not suffer from these topological and parameterization issues. These methods use implicit representations of the label-field, where an object's boundary is, for example represented through the zero level-set of a function. \mn{A parametric boundary representation is evolved {\it directly}, for example by moving its associated control points. For a level-set representation the {\it level-set function} is evolved, which {\it indirectly} implies an evolution of the segmentation boundary. Specifically, an evolution equation is imposed on the level-set function such that its zero level-set moves as desired.} As the level-set function is \mn{(by construction)} either strictly positive or negative (depending on convention used) inside the object and strictly negative or positive on the outside of the object, a labeling can be obtained by simple thresholding. Level-set approaches for image segmentation make use of advanced numerical methods to solve the associated partial differential equations~\cite{osher2003,sapiro2001}. To assure boundary regularity, segmentation energies typically penalize boundary length \mn{or} surface area.

While initial curve and surface evolution methods focused on energy minimization based on boundary regularity and boundary misfit, later approaches such as the Chan-Vese model~\cite{chan2000active}, added terms that encoded statistics about the regions inside and outside the segmentation boundary. \mn{Such \mn{region-based} models can be as simple as \mn{homoscedastic (\ie, same variance)} Gaussian likelihoods with \mn{specified (but distinct) means for foreground and background respectively}, as in the Chan-Vese case.} \mn{They can also be much more complex such as trying to maximally separate intensity or feature distributions inside and outside an object~\cite{georgiou2007}. \mn{Overall, a large variety of region-based approaches exist, providing great modeling freedom~\cite{cremers2007review}}}. While region-based models are less sensitive to initialization, they are still non-convex when combined with weighted curve-length terms for regularization. Hence, a global optimum cannot be guaranteed by numerical optimization for such formulations. The dependency on curve \mn{and} surface initializations popularized the formulation of energy minimization methods \mn{which can find a global energy optimum. One such approach is to reformulate an energy minimization problem as a problem defined over an appropriately chosen graph.}

In the context of image segmentation, the idea is to create a graph with added source and sink nodes in such a way that a minimum cut of the graph implies a variable configuration which minimizes the original image segmentation energy~\cite{boykov2001}. For a large class of binary segmentation problems\mn{, these graph-cut approaches} allow for the efficient computation of {\it globally} optimal solutions through max-flow algorithms~\cite{kolmogorov2004}. In particular, discrete versions of the active-contour and Chan-Vese models (with fixed means) can be formulated. To avoid computing trivial solutions for the boundary-only active contours, graph-cut formulations add seed-points, specifying fixed background and foreground pixels \mn{or} voxels \mn{(in 3D)}. While conceptually attractive, graph-cut approaches suffer from the need to build the graphs and the necessity to specify discrete neighborhood structures which may negatively affect the regularity of the obtained solution by creating so-called metrication artifacts.

Recently, the focus has shifted away from level-set and graph representations to formulations of active contours and related models by means of indicator functions~\cite{appleton2006,bresson2007} defined in the continuum and allowing for convex formulations. These methods are closely related to segmentation via graph-cuts, but avoid the construction of graphs and can alleviate metrication artifacts. A key insight here is that the boundary-length \mn{or} area term can \mn{formulated through the total variation of an} indicator function. \mn{This regularization formulation} becomes convex when followed by relaxation of the indicator function to the \mn{interval} $[0,1]$. Hence these approaches strike an attractive balance between Partial Differential Equation (PDE)-based level-set formulations and the global properties of graph-cut methods. \mn{As} they are specified via PDEs, \mn{highly accurate and fast implementations for these} solvers are available~\cite{pock2008fast}.
As these convex formulations make use of \mn{TV} terms, they are conceptually related to \mn{TV} image-denoising. The use of TV regularization for denoising was pioneered by Rudin, Osher and Fatemi (the ROF model \cite{rudin1992nonlinear}). The ROF model uses quadratic (\ie, $\ell_2$) coupling to the image intensities and TV for edge-preserving noise-removal~\cite{burger2013guide}. Approaches with $\ell_1$ coupling yielding a form of geometric scale-space have also been proposed~\cite{chan2005}. \mn{As we will see, our proposed approach will be closely related to these modern TV regularization and denoising approaches.}

Segmentation approaches based on energy-optimization as discussed above typically either have a probabilistic interpretation (as negative log-likelihoods) or have been explicitly derived from probabilistic considerations. The reader is referred to Cremers et al.  \cite{cremers2007review} for a review of recent developments in probabilistic level-set segmentation. All these techniques, while probabilistic in nature, seek optimal labels and do not directly provide information about the posterior distribution or uncertainty in their solutions. \mn{In contrast, our proposed AMF approach will approximate posterior distributions from which one can infer a segmentation and corresponding uncertainty.}

\subsection{AMF Segmentation Approach}
\label{sec:amf_segmentation_approach_preview}

\mn{AMF segmentation is a Bayesian approach, which results in a posterior distribution on the label map. The AMF approach combines explicit representations of label likelihoods with a boundary length prior}. As we \mn{will} show, our approach makes strong connections to ROF-denoising, and convex active-contour as well as probabilistic active-contour formulations.

\mn{In prior work, Monte-Carlo approaches have been used to characterize posterior distributions on segmentations, which require sampling~\cite{fan2007,chang2011,petersen2010}.  In addition, the Monte-Carlo approach is quite general about statistical modeling assumptions so that it could be applied to the likelihood and regularity terms of our segmentation tasks. Approximations are then only caused by the sampling.
A potential drawback of such a Monte-Carlo approach is that an accurate estimation might require the generation of a large number of samples, which can be time consuming. \\
 
In contrast to the Monte-Carlo approach, our mean-field approximation is based on a factorized distribution that is quick to compute, but which is a relatively severe approximation. A potential drawback of our method is that samples drawn from the approximated posterior can have an un-natural fragmented appearance. However, our experimental results reveal that the approximation is surprisingly accurate (in terms of correlation of the posterior probabilities and the segmentation area), when compared to the exact model using much slower Gibbs sampling.
}
~\\
\begin{mdframed} 
\mn{In summary, the primary advantages of our approach are speed, simplicity, and leverage of existing convex solver technology. We show in~\Sec{SecROF} that using ROF-denoising on the logit field of label probabilities results in a ``denoised'' logit transform from which a label probability function can easily be obtained through a sigmoid transformation. Given an ROF solver, the AMF model can thus be implemented in one line of source-code. Furthermore, the AMF model provides a good approximation of the posterior of the segmentation under a curve-length prior as we experimentally show in~\Sec{SecAgreementWithOriginalModel}. }
\end{mdframed}


\subsection{Structure of the Article}

In \Sec{SecAMF}, we specify a discrete-space probabilistic formulation of
segmentation with the goal of finding the posterior distribution of
labels, given an input image.  \mn{We use the VMF approach}, along with
a linearization approximation that \mn{simplifies} the problem.
This results in an optimization problem for determining the parameters
of an approximation to the posterior distribution on pixel labels.  In
the style of Chan and Vese \cite{chan2000active} and many others, we shift
from discrete to continuous space facilitating use of the
calculus of variations for the optimization problem, yielding the
Euler-Lagrange equations for the AMF model.

\mn{In \Sec{SecConnections}, we show that the AMF Euler-Lagrange equations for the zero level-set
correspond to those of a special case of the Chan-Vese
model~\cite{chan2000active}, and that the AMF ``approximate
posterior'' has the same mode, or MAP realization, as the exact
posterior distribution. Subsequently we show that the
AMF Euler-Lagrange equations have the same form as those of the ROF model of image
denoising, and we discuss methods that may be used for solving AMF.}

\Sec{SecProperties} describes \mn{other important} properties of AMF.  We show that
for a one-parameter family of realizations, the approximated and exact
posteriors agree ratiometrically, and we explore their agreement for more
general realizations. In addition, we show that the AMF problem is
convex, and is unbiased in a particular sense.

\Sec{SecExpts} \mn{shows the}  experimental results \mn{on} examples that include intensity ambiguities. \mn{It also demonstrates} the quality of the AMF in approximating the true posterior via Gibbs sampling. \mn{Furthermore, \Sec{SecExpts} discusses AMF results for} real ultrasound images of the heart\mn{,} the prostate
\mn{, a common test image in computer vision, and on a large collection of images from the \texttt{icgbench} segmentation dataset~\cite{santner2010}.}

Finally, \Sec{SecConclusions} concludes with a summary and an outlook on future work. Detailed derivations of the approximation properties can be found in the appendix.


\section{Active Mean Fields (AMF)}
\label{SecAMF}
\label{SecAMFDiscrete}
This section introduces the basic discrete-space probabilistic model
(\Sec{SecProb}), that includes a conventional conditionally
independent likelihood term and a prior that penalizes the boundary
length of the labeling.  The \mn{VMF} approach
is used (\Sec{SecMF}), along with a Plefka approximation
(\Sec{SecPlefka}), to construct a factorized distribution that,
given image data, approximates the posterior distribution on labelings.
The resulting optimization problem for determining the
parameters of the variational distribution has a KL-divergence data
attachment term and a TV regularizer.  The objective
function is converted to continuous space (\Sec{SecTV}), yielding the Euler-Lagrange
equations of the AMF model (\Sec{SecEL}), that  involve logit label probabilities and
likelihoods along with a curvature term.

In the following sections, we use upper-case $P$ and $Q$ to represent probability mass functions and lower-case $p$ and $q$ to represent probability density functions.

\subsection{Original Probability Model}
\label{SecProb}

Define the space of images as a compact domain
\footnote{Our theory also holds for higher dimensions, \ie, $\cX\subset\R^n$. We discuss our approach in $\R^2$ for simplicity \mn{and hence talk about pixels. In 3D for example, we would deal with voxel grids and we would need to compute a 3D variant of total variation, but the overall results will hold unchanged.}}  $\cX \subset {\mathbb R}^2$ indexed by $\x \in {\mathbb R}^2$ and let  ${\mathbb I}_{\cX} \triangleq \{i: \x_i \in \cX\}$ denote the indices of the lattice of image pixels. \mn{Furthermore, $\rmZ$} denotes a binary random field defined on the pixel lattice
whose realizations $\zvar$ are the binary labelings of a real-valued image $\y$ on the pixel lattice; \mn{given the image pixel index $i \in {\mathbb I}_{\cX}$,} \mn{$\zvar_i$ and $\y_i$ are the corresponding quantities specific to pixel $\x_i  \in \cX$. For convenience, we write $p(\y_i|h) \triangleq p(\y_i|z_i=h)$ with $h \in \left\{0,1\right\}$, where the definition of $p(\y_i|\zvar_i)$ is problem specific and is assumed to be given (for example, specified parametrically or obtained through kernel density estimation on a given feature space; we will not address this issue here). Now, if we make} the usual assumption that the likelihood term, \ie, the probability density of observing intensities conditioned on labels, is conditionally \mn{independent and identically distributed (iid)},  \ie
{\opt{arxiv}{\small}
\begin{align}
 p(\y|\zvar) = \prod_{\mn{i\in {\mathbb I}_{\cX}}} p\left(\y_i~|~\zvar_i\right),
\end{align}
}
\mn{then the corresponding log-likelihood, defined with respect to the logit transform of the pixel-specific image likelihood}
{\opt{arxiv}{\small}
\begin{align}
    \psi_i \triangleq \ln\frac{ p(\y_i|1)}{ p(\y_i|0)} \CM
\end{align}
}
\mn{is}
{\opt{arxiv}{\small}
\begin{align}
\ln p(\y|\zvar) = \sum_\indx \ln p\left(\y_i~|~\zvar_i\right) = \sum_\indx \zvar_i \psi_i + \ln p(\y_i|0).
\label{eq:likely}
\end{align}
}

Next, we apply a prior that penalizes the length $L(z)$ of the boundaries of the label map,
{\opt{arxiv}{\small}
\begin{align}\label{eq:prior-def}
    P(\zvar) \propto \exp(-\lambda L(\z) ).
\end{align}
}
{Here, {\small$\lambda\in\R^+$} is a constant. The larger $\lambda$ the more irregular segmentation boundaries are penalized and therefore discouraged.} We defer discussion of the length functional $L(\cdot)$ to \Sec{SecTV}.

By Bayes' rule the posterior probability of the label map given the image is
{\opt{arxiv}{\small}
\begin{align}\label{eq:Bayes-rule}
P(\zvar | y) \propto&~~  p(y|\zvar) P(\zvar)\\
\intertext{\mn{so that}} 
\ln P(\zvar | y) =& \sum_{i \in \cX} \zvar_i \psi_i + \ln p(\y_i|0)\;  - \lambda L(\zvar)  + \const. \label{eq:Bayes-rule2}
\end{align}
}
Here the constant is equal to the log-partition function of the prior distribution. \mn{This constant is not easily computed, as it requires a sum over all of the configurations of $\zvar$.}

\subsection{\mn{Variational} Mean-Field Approximation}
\label{SecMF}

\mn{As mentioned above, the partition function cannot easily be computed.} \mn{In the variational mean-field (VMF) approach~\cite{wainwright2008graphical}, we approximate the posterior distribution $P$ via a simpler variational distribution $Q$  by minimizing the distance between $P$ and $Q$ (here, in a Kullback-Leibler sense -- see details below). The explicit computation of the integrals involved in the partition function (for continuous variables) can thereby be avoided.} \mn{Specifically, the} mean-field method approximates the joint distribution of a countable family of random-variables as a product of univariate distributions. \mn{The VMF approximation is widely used in machine learning and other fields \cite{wainwright2008graphical}.}

For the binary segmentation problem, we define the mean-field approximation \mn{$Q(\z;\theta)$} of the posterior distribution $P(\z | y)$ \mn{as a field of independent Bernoulli random variables $z_i$ defined on the lattice $\indx$ with probability $\theta_i$, which constitute the random field $Z$:}
{\opt{arxiv}{\small}
\begin{align} \label{eqn:mean-field-finite}
Q(\zvar;~{\theta}) &\triangleq \prod_\indx {\theta}_i^{\zvar_i} \left(1 - {\theta}_i\right)^{1 - \zvar_i}\\
    =&  \exp\left\{ \sum_\indx \left[ \zvar_i \phi_i  + \ln \left(1 - {\theta}_i\right)\right] \right\},
\label{eqn:mean-field-finite-2}
\end{align}
}
where $\phi_i \triangleq \ln \frac{ {\theta}_i}{1 - {\theta}_i}$. \mn{Next, \mn{$Q(\z;\mn{\cdot})$} is parameterized so that it minimizes the KL-divergence with respect to the original probability model, \ie, }
{\opt{arxiv}{\small}
\begin{align}
    \mn{\hat{\theta}_*}\mn{\triangleq}& \arg\min_\theta\KLD{Q(\zvar;\theta)}{P(\zvar| y)}\\
     =& \arg\min_\theta \EV{Q}{\ln Q(\zvar;\theta)-\ln P(\zvar| y) }\\
     =& \arg\min_\theta \EV{Q}{\ln Q(\zvar;\theta)-\ln p(y| \zvar) + \lambda L(\zvar)}.
\label{eqn:MFEQ}
\end{align}
}
With minor abuse of KL-divergence notation:
{\opt{arxiv}{\small}
\begin{equation}
\label{eqn:ct-kldiv-2}
  \mn{\hat{\theta}_*} = \arg\min_\theta \KLD{Q(\zvar;\theta)}{p(y|\zvar)} + \EV{Q}{\lambda L(\zvar)}.
\end{equation}
}
\mn{In other words, the VMF approximation selects the parameters of the factorized variational distribution $Q(\rmZ;~\theta)$ such that (i) local image likelihood information, $p(y|\zvar)$, is captured {\it while at the same time} (ii) considering the expected value of the segmentation boundary length (which is a global property that regularizes the solution).}

\subsection{Plefka's Approximation}
\label{SecPlefka}
Although minimizing the KL-divergence term in \Eqn{eqn:ct-kldiv-2} with respect to $\theta$ is relatively straightforward, minimizing {\opt{arxiv}{\small}$\EV{Q}{L(\zvar)}$} is \mn{generally} not as it entails a sum over all configurations of $\zvar$. In the mean-field literature, difficult expectations of functions of random-fields have been approximated
using Plefka's method~\cite{plefka1982}.

\mn{Noting that $\EV{Q}{\zvar}= \theta$ according to \Eqn{eqn:mean-field-finite} and that the first order Taylor expansion of the curve length function with respect to $\zvar^*$ is {\opt{arxiv}{\small} \mbox{$L(\zvar) \approx L(\zvar^*) + (\zvar-\zvar^*)\cdot \nabla L(\zvar^*) $}, then Plefka's approximation states that}
\begin{equation}
\EV{Q}{L(\z)} \approx L(\zvar^*) + (\EV{Q}{\z}-\zvar^*)\nabla L(\zvar^*) \approx L(\EV{Q}{\z}) = L(\theta)
\end{equation}
so that an approximation of \Eqn{eqn:ct-kldiv-2} is
{\opt{arxiv}{\small}\begin{align}\label{eqn:ct-amf-approx}
  \hat{\theta} \mn{\triangleq}& \arg\min_\theta \KLD{Q(\zvar;\theta)}{p(y| \zvar)} + \lambda  L(\theta),
\end{align}
}
\mn{where $\hat{\theta}_*\approx\hat{\theta}$.}

\mn{Assuming $L(\cdot)$ is convex, then the Plefka approximation of \Eqn{eqn:ct-amf-approx} is a lower bound to the original objective function of \Eqn{eqn:ct-kldiv-2} as Jensen's inequality states }$ \EV{Q}{L(\z)} \geq L(\EV{Q}{z}) = L(\theta)$}. While this is not directly useful for our purposes, there has been some work on ``converse Jensen inequalities''~\cite{dragomir2004} that may provide useful bound relationships.
In the end, approximations are justified by the quality of their results, \mn{such as the favorable properties highlighted in \Sec{SecProperties}.}

\subsection{\mn{Continuous Variant of Variational Problem}}
\label{SecTV}
In the previous section, we showed how the problem of computing the posterior distribution of a label-field under an (unspecified) boundary-length prior results in solving the optimization problem of \Eqn{eqn:ct-amf-approx}. To solve this problem using computationally efficient PDE optimization techniques, we first replace the random-field defined on a discrete lattice by one defined on continuous space.

Expanding \mn{\Eqn{eqn:ct-amf-approx} by using the definition of the log likelihood (\Eqn{eq:likely}) and of $Q(\cdot,\cdot)$ (\Eqn{eqn:mean-field-finite-2})} we get:
\opt{arxiv}
{\small
\begin{align}
\notag
\hat{\theta} =&  \argmin_\theta \Exp_{Q}\left[ \sum_\indx \zvar_i \phi_i  + \ln \left(1 - {\theta}_i\right)  -  \zvar_i \psi_i - \ln p(\y_i|0) \right]
\\\notag
&\hspace{1in} + \lambda L(\theta)
\\\notag
=&  \argmin_\theta \sum_\indx \left[ \theta_i \phi_i  + \ln \left(1 - {\theta}_i\right) -  \theta_i \psi_i - \ln p(\y_i|0)\right]
\\
&\hspace{1in}
 + \lambda L(\theta).
\end{align}
}
\opt{siims}
{
\begin{align}
\hat{\theta} =&  \argmin_\theta \Exp_{Q}\left[ \sum_\indx \zvar_i \phi_i  + \ln \left(1 - {\theta}_i\right)  -  \zvar_i \psi_i - \ln p(\y_i|0) \right]
\\
&\hspace{1in} + \lambda L(\theta)
\\
=&  \argmin_\theta \sum_\indx \left[ \theta_i \phi_i  + \ln \left(1 - {\theta}_i\right) -  \theta_i \psi_i - \ln p(\y_i|0)\right] + \lambda L(\theta).
\\
=&  \argmin_\theta \sum_\indx \left[ \theta_i \phi_i  + \ln \left(1 - {\theta}_i\right) -  \theta_i \psi_i \right] + \lambda L(\theta).
\label{eqn:discreteMFMin}
\end{align}
}

\mn{To solve the above equation by extending $\theta$ to the continuum, \mn{the logit transform of the likelihood} is now defined as
\begin{equation}
\psi(\x) \triangleq \ln\frac{ p(\y(\x)|\z(\x)=1)}{ p(\y(\x)|\z(\x)=0)},
\label{eqn:psi}
\end{equation}
\mn{where $\x$ denotes the location (\ie, the continuous equivalent of the index $\indx$), and $\y(\x)$, $\z(\x)$, and $\theta(\x)$ are the corresponding values of $\y$, $\z$, and $\theta$ at location $\x$.}
\mn{Similarly, the continuous variant of the logit transform of the variational probability function, $\theta(\x$), is defined as}
\begin{equation}
\phi(\x)\triangleq \ln\frac{\theta(\x)}{1-\theta(\x)}.
\label{eqn:phi}
\end{equation}
%
Now, if we denote with $v$ the area 
of a lattice element and replace the summation over the lattice with integration over $\cX$, then \Eqn{eqn:discreteMFMin} becomes in the continuous space,}
\opt{arxiv}
{\small
\begin{align}
\hat{\theta} =&  \argmin_\theta v^{-1}\int_\cX  \theta(\x) ( \phi(\x) - \psi(\x)) + \ln \left(1 - {\theta}(\x)\right) d\x
\\ &\hspace{1in} + \quad \lambda L(\theta) \PD
\end{align}
}
\opt{siims}
{
\begin{align}
\hat{\theta} =&  \argmin_\theta ~v^{-1}\cdot \left(\int_\cX  \theta(\x) ( \phi(\x) - \psi(\x)) + \ln \left(1 - {\theta}(\x)\right) d\x \right) + \lambda L(\theta) \PD
\end{align}
}

By the co-area formula~\cite{bethuel1994}, the length of the boundaries of a binary image defined on the continuum is equal to its total-variation:
{\opt{arxiv}{\small}
\begin{align} \label{eqn:bdry-len-prior}
    L(\zvar) =  \TV[\zvar(\x)] = \int_\cX ||\nabla \zvar(\x)||_2 ~d\x
\end{align}
}
where \mn{$||\cdot||_2$ is the 2-norm and }{\opt{arxiv}{\small}$\nabla \zvar$} is the (weak) gradient of {\opt{arxiv}{\small}$z$}.\footnote{$\nabla(z)$ is defined as $\int_\cX \langle \nabla \zvar, \eta \rangle d\x = -\int_\cX \langle \zvar,\nabla\cdot\eta \rangle~d\x$ for any test function $\eta:\cX\rightarrow \R^2$; \mn{in the case of $\z(\x)$ being an element of a convex set, $L(z)$ is convex}.}

Therefore putting it all together, the continuous variant of the variational problem is:
\opt{arxiv}
{\small
\begin{multline} \label{eq:ObjFunCont}
\hat{\theta} =  \argmin_\theta   \int_\cX  \theta(\x) (\phi(\x) - \psi(\x))  + \ln \left(1 - {\theta}(\x)\right) \\ +
 v\lambda ||\nabla \theta(\x)||_2 ~ d\x,
\end{multline}
}
\opt{siims}
{
\begin{equation} \label{eq:ObjFunCont}
\hat{\theta} =  \argmin_\theta   \int_\cX  \theta(\x) (\phi(\x) - \psi(\x))  + \ln \left(1 - {\theta}(\x)\right) +
 v\lambda ||\nabla \theta(\x)||_2 ~ d\x,
\end{equation}
}
which we call the {\it Active Mean Field approximation}. \mn{Note, that $\phi(x)$ depends on $\theta(x)$ according to \mn{\Eqn{eqn:phi}}.}

\subsection{Euler-Lagrange (EL) Equations}
\label{SecEL}

Defining the curvature operator,
\mn{
{\opt{arxiv}{\small}
\begin{align}
\kappa(\theta(x))\triangleq \nabla \cdot \left( \frac{\nabla \theta(x)}{\|\nabla \theta(x)\|_2} \right),
\end{align}
}
}
\mn{the Euler-Lagrange equation describing the stationary points of \Eqn{eq:ObjFunCont} is given by:}
{\opt{arxiv}{\small}
\begin{align}
\phi(x) - \psi(x) - v \lambda \area \kappa(\theta(x)) = 0 \PD
\end{align}
}
\mn{This can be derived as follows: Expanding $\phi(x)$ according to \mn{\Eqn{eqn:phi}}, we obtain the objective function
\begin{equation}
        E(\theta) = \int_\cX -\theta(x)\psi(x) + \theta(x)\ln(\theta(x)) + (1-\theta(x))\ln(1-\theta(x))+v\lambda\|\nabla \theta(x)\|_2 ~d\x.
\end{equation}
The variation of $E(\theta)$ is~\cite{troutman2012}
\begin{equation}
        \delta E(\theta;\delta\theta) \triangleq \frac{\partial E(\theta + \epsilon\delta\theta)}{\partial\epsilon}\arrowvert_{\epsilon=0} \CM
\end{equation}
\mn{where $\epsilon\in\mathbb{R}$, $\delta\theta$ denotes an admissible perturbation of $E(\theta)$, and \mn{$\frac{\partial}{\partial \epsilon}$ denotes the partial derivative with respect to $\epsilon$.} The variation becomes}
\begin{equation}
    \delta E(\theta;\delta\theta) = \int_\cX -\psi(x)\delta\theta(x) + \ln\frac{\theta(x)}{1-\theta(x)}\delta\theta(x) + v\lambda \frac{1}{\|\nabla \theta(x)\|_2}\nabla\theta(x)\cdot \nabla\delta\theta(x) ~d\x.
\end{equation}
Integration by parts assuming Neumann boundary conditions and using \Eqn{eqn:phi} results in
\begin{equation}
 \delta E(\theta;\delta\theta) = \int_\cX \left(\phi(x)-\psi(x) -v\lambda \nabla\cdot\left(\frac{\nabla\theta(x)}{\|\nabla \theta(x)\|_2}\right)\right)\delta\theta(x)~d\x \PD
\end{equation}
\mn{As the variation needs to vanish for all admissible perturbations $\delta \theta(x)$ at optimality, we obtain the Euler-Lagrange equation}
\mn{
\begin{equation}
        \phi(x) - \psi(x) - v \lambda \area \kappa(\theta(x)) = 0\PD\label{eq:AMF_phi_theta}
\end{equation}
}
\mn{According to~\Eqn{eqn:phi}, $\phi(x)$ is obtained from $\theta(x)$ through a logit transform. Consequentially, we can obtain $\theta(x)$ from $\phi(x)$ via the sigmoid function 
\begin{equation}
\sigma(x) \triangleq (1 + \exp(-x))^{-1}
\end{equation}
as  $\theta(x)=\sigma(\phi(x))$.  The sigmoid function, $\sigma(\cdot)$, is monotonic (\ie, $\sigma'(x)>0$) so that}
\begin{equation}
        \nabla\theta(x) = \sigma'(\phi(x))\nabla\phi(x)
\end{equation}
and
\begin{equation}
        \frac{\nabla\theta(x)}{\|\nabla\theta(x)\|_2} = \frac{\sigma'(\phi(x))\nabla\phi(x)}{\|\sigma'(\phi(x))\nabla\phi(x)\|_2} = \frac{\nabla\phi(x)}{\|\nabla\phi(x)\|_2} \PD
\end{equation}
Hence, the Euler-Lagrange equation can be rewritten as
}
{\opt{arxiv}{\small}
\begin{equation}
  \phi(x) - \psi(x) - v \lambda \area \kappa(\phi(x)) = 0\PD
  \label{EqMF}
\end{equation}
}
In summary, the distribution {\opt{arxiv}{\small}$Q(\z;\theta)$} approximates the ``exact'' distribution, {\opt{arxiv}{\small}$P(\z|y)$}, in the KL-divergence sense when $\phi$ (the logit transform of the
parameter $\theta$) satisfies the Euler-Lagrange equation of the AMF model; we will refer
to \Eqn{EqMF} as the ``AMF Equation.'' As the objective function is strictly convex (see \Sec{SecConvexity}) in {\opt{arxiv}{\small}$\theta$}, the stationary point is the unique global optimum.

\section{Connections to Chan-Vese and ROF}
\label{SecConnections}
In this section we establish the connection between the AMF model and the Chan-Vese segmentation model (\Sec{SecChanVese}) as well as the ROF denoising model (\Sec{SecROF}). In particular, we show that the Chan Vese Euler-Lagrange equations correspond to those of the zero level-set of the AMF model, so a Chan-Vese segmentation can be obtained as the zero level-set of the AMF solution. We also show that the AMF Euler-Lagrange equations (\mn{\Eqn{EqMF}}) have the same form as those of the ROF model. Therefore, the solver technologies that have been developed for the ROF model may be deployed for AMF.

\subsection{Connection to Chan-Vese}
\label{SecChanVese}

{
\mn{To derive the connection between the AMF and the Chan-Vese approach, we introduce the energy $E_{cv}(\cdot)$ for the generalized Chan-Vese model based on a relaxed indicator function (\ie, {\opt{arxiv}{\small}$\theta\in[0,1]$}), which, according to~\cite{bresson2007}, can be written as}
{\opt{arxiv}{\small}
\begin{equation}
  E_{cv}(\theta) = \int_\cX -\theta(\x)\psi(\x) + v \lambda \|\nabla\theta(\x)\|_2~d\x,
  \label{eq:generalized_cv_energy}
\end{equation}
}
\mn{with the first part of the function being the data term  and the second term regularizing the boundary length}. Such a length prior is essential to encourage large, contiguous segmentation areas. The importance of the length-prior becomes especially clear in the context of the Mumford-Shah model~\cite{mumford1989}, of which the Chan-Vese model is a special case. In the absence of a length prior, \mn{the Mumford-Shah approach will assign each pixel in regions with constant image intensity to its own (separate) parcel.}
\mn{The standard Chan-Vese model~\cite{chan2001active} (without the area prior of this model) can be recovered from \Eqn{eq:generalized_cv_energy} for the special case that the class conditional intensity model is Gaussian, \ie,}
{\opt{arxiv}{\small}$y_i|\zvar_i = 1 \sim N(\mu_1, \sigma^2_1)$} and {\opt{arxiv}{\small}$y_i|\zvar_i = 0 \sim N(\mu_0, \sigma^2_0)$}.
In this case:
{\opt{arxiv}{\small}
\begin{equation}
  \psi(x) \triangleq \mn{\ln\frac{\sigma_0}{\sigma_1}}- \frac{1}{2 \sigma_0^2}(y(x) - \mu_0)^2 + \frac{1}{2 \sigma_1^2}(y(x) - \mu_1)^2,
  \label{EqGaussLikelihood}
\end{equation}
}
\mn{and the corresponding Chan-Vese energy becomes:
\begin{equation}
        E_{cv}^{gauss}(\theta)=\int_\cX -\theta(\x)\left(\ln\frac{\sigma_0}{\sigma_1}- \frac{1}{2 \sigma_0^2}(y(x) - \mu_0)^2 + \frac{1}{2 \sigma_1^2}(y(x) - \mu_1)^2\right) + v \lambda \|\nabla\theta(\x)\|_2~d\x.\label{EqChanVeseGauss}
\end{equation}
}
\mn{The means of the Gaussians ($\mu_1$, $\mu_2$) are estimated jointly in the standard Chan-Vese model~\cite{chan2000active} \mn{and the standard deviations are assumed to be fixed and identical.}} \mn{In contrast, in the generalized Chan-Vese model~(\Eqn{eq:generalized_cv_energy}), parameters of $\psi(\x)$ are typically assumed to be fixed and are not jointly estimated. This assures the convexity of the overall model. However, if desired, these parameters can also be estimated. A simple approach would be an alternating optimization strategy.} \mn{Note that the Chan-Vese segmentation model of~\Eqn{EqChanVeseGauss} becomes Otsu-thresholding~\cite{otsu1975} if the length prior is disregarded ($\lambda=0$). Hence, unlike Chan-Vese segmentation, Otsu-thresholding cannot suppress image fragmentation and irregularity.}

The Euler-Lagrange equations of the generalized Chan-Vese energy~\mn{(\Eqn{eq:generalized_cv_energy})} \mn{are:}
\begin{equation}
  -\psi(\x) -v \lambda \kappa(\theta(\x)) = 0.
\end{equation}
This is identical to the AMF Euler-Lagrange equation \mn{(\Eqn{eq:AMF_phi_theta})} at the zero level-set
$\phi(x) = 0$. By construction, the zero level-set of a {\it level-set implementation for the generalized Chan-Vese model} has to agree with the solution obtained from the {\it Euler-Lagrange equations of the generalized Chan-Vese model using indicator functions} as both minimize the same energy function just using different parameterizations. Consequentially, also, the zero level-sets of both the AMF model and the level-set implementation of Chan-Vese need to agree.

In contrast to the generalized Chan-Vese model described above, the original Chan-Vese model \mn{of~\cite{chan2000active}, formulated as a curve evolution approach,} \mn{is characterized by an energy function (penalizing segmentations with large, continuous areas)} \mn{with an additional term of the form
\begin{equation}
  E_{area}(C)=\nu {\rm Area}(inside (C)),
\end{equation}
\mn{where $C$ denotes the curve defining the boundary of the segmentation, $\nu\in\mathbb{R}_+^0$ is a non-negative constant to weight the area influence, and ${\rm Area}(inside (C))$ simply denotes the area enclosed by $C$. For implementation purposes $C$ is implicitly represented by the zero level-set of a {\it level set function} $\phi$.} The corresponding Euler-Lagrange equation is, on the zero level-set~\cite{chan2000active},
\begin{equation}
  -\psi(\x) - v \lambda \kappa(\phi(\x)) + \nu = 0 \label{eq:chan_vese_el_with_area}\PD
\end{equation}
}
\mn{Examining} the $\nu$ level-set of the AMF model \mn{(\Eqn{EqMF})}, so that $\phi(x) = \nu$, we notice that this level-set satisfies the same Euler-Lagrange equation as the zero level-set of the Chan-Vese model with a specified non-zero value of $\nu$. In other words, the level-sets of the dense AMF solution provide a family of solutions for the Chan-Vese problem for a continuum of values of the area penalty.}

\mn{Note that such area penalties cannot effectively be added in the indicator-function based approaches to the Chan-Vese active-contour models proposed by Appleton et al.~\cite{appleton2006} and Bresson et al.~\cite{bresson2007}. The goal of these models is to capture a binary segmentation result through a relaxed indicator function, (\ie, $\theta\in[0,1]$ instead of $\theta\in\{0,1\}$). However, it can be shown~\cite{niethammer2013} that in certain instances this relaxation produces undesirable segmentation results when combined with an area penalty.}


\subsection{Connection between AMF and ROF Models}
\label{SecROF}

In their seminal paper, Rudin, Osher and Fatemi \cite{rudin1992nonlinear} proposed a denoising
method for, e.g., intensity images $u_0(\x)$,
\opt{arxiv}
{
\begin{multline}
  u^*(\x) = \arg \min_u  \int_\cX \|\nabla u(\x)\|_2~d\x \\ \mbox{s.t.} \quad \int_\cX (u(\x) - u_0(\x))^2~d\x = \sigma^2 \CM
\end{multline}
}
\opt{siims}
{
\begin{equation}
  u^*(\x) = \arg \min_u  \int_\cX \|\nabla u(\x)\|_2~d\x \quad \mbox{s.t.} \quad \int_\cX (u(\x) - u_0(\x))^2~d\x = \sigma^2 \CM
\end{equation}
}
\mn{where $\sigma>0$.} As discussed by Vogel and Oman \cite{vogel1996iterative}, this is equivalent to the following unconstrained problem,
\opt{arxiv}
{
\begin{multline}
  u^*(\x) = \arg \min_u \left[ \int_\cX (u(\x) - u_0(\x))^2 d\x \right. \\ \left. + \frac{\alpha}{2} \int_\cX \|\nabla u(\x)\|_2 d\x \right] \CM
\end{multline}
}
\opt{siims}
{
\begin{equation}
  u^*(\x) = \arg \min_u \left[ \int_\cX (u(\x) - u_0(\x))^2 d\x \right. \left. + \frac{\alpha}{2} \int_\cX \|\nabla u(\x)\|_2 d\x \right] \CM
\end{equation}
}
for a suitable choice of \mn{$\alpha>0$}. They refer to this formulation as ``TV penalized least squares.''

The corresponding Euler-Lagrange equation is
\begin{equation}
  u(\x) - u_0(\x) - \alpha \kappa(u(\x)) = 0 \PD\label{eq:el_rof}
\end{equation}

\mn{For $\alpha=v\lambda$}, this equation has the same form as the Euler-Lagrange equations of the
AMF model \mn{of~\Eqn{EqMF}, which is}  
\begin{equation}
\phi(\x) - \psi(\x) - v \lambda \area \kappa(\phi(\x)) = 0\PD
\label{EqAMF2}
\end{equation}  In this equivalence, the denoised intensity \mn{image} of the ROF
model, $u$, corresponds to the logit parameter field of the AMF distribution, $\phi$,
while the noisy input intensity \mn{image} of \mn{the ROF model}, $u_0$, corresponds to the \mn{logit-transformed
label probabilities} in the AMF problem, $\psi$.  Furthermore, if the class conditional
intensity model is homoscedastic Gaussian, then (from \mn{\Eqn{EqGaussLikelihood}})
$\psi(\x)$ is \mn{linear} in the observed intensity. \mn{Furthermore,} the AMF solution is equivalent to solving an ROF problem that is effectively denoising the \mn{logit-transformed approximation of the posterior label likelihoods.}


Because of the equivalence of the Euler-Lagrange equations of the AMF and the ROF models, \mn{the considerable technology developed for solving the ROF model may be applied to the AMF model}.
In particular, \mn{a globally optimal solution (see \Sec{SecConvexity} for a proof of the convexity of this model) of the AMF model can be computed by the ROF denoising approach.} 
\mn{In other words,} given an ROF solver (ROFsolve) that minimizes
\begin{equation}
  E_{ROF}(u;u_0,\mn{\alpha}) = \int_\cX (u(\x)-u_0(\x))^2 + \frac{\mn{\alpha}}{2}\|\nabla u\|_2~d\x
\end{equation}
such that
\begin{equation}
  u^* \mn{\triangleq} \underset{u}{\text{arg min}}~E_{ROF}(u;u_0,\mn{\alpha}) = \mn{\text{ROFsolve}( u_0,\mn{\alpha} )} ,
\end{equation}
solving the AMF problem for a given $\psi$ and $\lambda$ then simply becomes
\begin{equation}
  \theta^* = \sigma( \text{ROFsolve}( \psi,\mn{v \lambda} ) ). \label{eq:AMF_by_ROF}
\end{equation}
\mn{\Eqn{eq:AMF_by_ROF}} is {\it the central result concerning the implementation of our method } as it connects the optimal AMF solution to a straightforward ROF denoising problem. 


\section{Additional Properties of AMF}
\label{SecProperties}

We \mn{now} summarize some approximation properties of AMF (\Sec{SecApproximation}), show the objective function to be convex (\Sec{SecConvexity}), and show that AMF is unbiased in a specific sense (\Sec{SecUnbiased}).

\subsection{Approximation Properties}
\label{SecApproximation}

Our goal is an efficient yet accurate approximation, $Q(\zvar;\theta)$, to the exact posterior distribution $P(\zvar|y)$ for general realizations of $\zvar$. \mn{To show that $Q(\zvar;\theta)$ is in fact a good approximation, we study its properties here.} For convenience, we only summarize the results of some of the approximation properties of the AMF model and refer to the appendix for mathematical details. In particular, the appendix shows that
\begin{itemize}
  \item[a)] The zero level-set of $\phi$ is the boundary of the most probable realization $\zvar_0$ of $Q(\zvar;\theta)$ \mn{{\it and}} \mn{is} the MAP realization under $P(\zvar|y)$.  This is not generally the case for mean field approximations.
\item[b)] Because the \mn{log partition function of the prior is not easily computed we compare} {\opt{arxiv}{\small}$\ln\frac{Q(\zvar;\theta)}{ Q(\zvar_0 ;\theta)}$} with {\opt{arxiv}{\small}$\ln\frac{P(\zvar|y)}{P(\zvar_0|y)}$}, where $\zvar_0$ is the most probable realization under both distributions according to a). \mn{These} probability ratios are not only in agreement for the zero level-set, but \mn{also for realizations} that are bounded by any level-set of $\phi$.
\item[c)] The probability ratios approximately agree for realization whose boundary normals are close in direction to $\nabla \phi$.
\item[d)] If neither a) nor b) hold, the probability ratio for \mn{$Q(\zvar,\theta)$} will be larger than that for $P(\zvar|y)$, \mn{\ie}, it underestimates the length penalty associated with the prior.
\end{itemize}

\subsection{Convexity}
\label{SecConvexity}

A nice property of the AMF model is that its energy is strictly convex and therefore we can find a unique global minimizer. This is in contrast to the TV based segmentation models~\cite{appleton2006,bresson2007} which are generally convex (but not strictly so) and therefore may have multiple non-unique optima.

To show convexity, we consider the continuum formulation of  AMF  which can be rewritten as a function of {\opt{arxiv}{\small}$\theta(\x)\in[0,1],$} as:
\opt{arxiv}
{\small
\begin{multline}
  \label{eq:constant_AMF_energy}
  E_{\mn{\text{amf}}}(\theta) = \int_\cX - \theta \psi + v \lambda \area\|\nabla\theta\|_2\\ +(1-\theta)\mn{\ln}(1-\theta) +\theta \mn{\ln} \theta~d\x
\end{multline}
}
\opt{siims}
{
\begin{equation}
  \label{eq:constant_AMF_energy}
  E_{\mn{\text{amf}}}(\theta) = \int_\cX - \theta \psi + v \lambda \area\|\nabla\theta\|_2 +(1-\theta)\mn{\ln}(1-\theta) +\theta \mn{\ln} \theta~\mn{d\x}
\end{equation}
}
{where dependencies on space are dropped only for notational convenience (\ie, $\theta=\theta(\x)$ and $\psi = \psi(\x)$) and we expressed \mn{$\phi$} in terms of $\theta$.}
The term
{\opt{arxiv}{\small}
\begin{equation}
\int_\cX - \theta\psi + \lambda \area\|\nabla\theta\|_2~d\x \label{eq:standard_TV}
\end{equation}
}
is convex in $\theta$ as the first summand is linear in $\theta$, the 2-norm is convex, $\nabla$ is a linear operator and both terms are summed with a positive weight $\lambda$. To see that the rest of the integrand is also convex, consider a function of the form
{\opt{arxiv}{\small}
\begin{align}
  f(\theta) =& (1-\theta)\mn{\ln}(1-\theta) + \theta \mn{\ln}(\theta).
\end{align}
}
which implies that {\opt{arxiv}{\small}
\begin{align}
 f''(\theta) =& \frac{1}{\theta(1-\theta)} > 0  \quad\text{for}\quad \theta\in(0,1).
\end{align}}
Therefore, {\opt{arxiv}{\small}$\int_\cX (1-\theta)\mn{\ln}(1-\theta)+\theta\mn{\ln}(\theta)~d\x$} is strictly convex. Because the sum of convex and strictly convex functions is strictly convex, the overall AMF energy is strictly convex in $\theta$ and therefore has a unique global minimizer (see~\cite{boyd2004} for details on convexity preserving operations). In particular, we note that for a non-informative data term, \mn{\ie}, pixels are locally equally likely to be foreground or background ($\psi=0$), $\theta(x)=\frac{1}{2}$ is the globally optimal solution. For the related standard TV segmentation model~\cite{appleton2006}, which would only minimize \mn{\Eqn{eq:standard_TV}}, any constant solution would be a global minimizer.

\subsection{Unbiased in Homogeneous Regions}
\label{SecUnbiased}

In this section we analyze the behavior of the AMF estimator over homogeneous (\ie, constant intensity) patches of an image. The AMF objective function, \Eqn{eq:ObjFunCont}, can be written:
\opt{arxiv}
{\small
\begin{multline}
\hat{\theta} =  \arg\min_\theta  \int_\cX  \KLD{Q(\zvar(x);\theta(\x))}{p(y(\x)| \zvar(\x))} d\x + v \lambda \TV[\z] \PD
\end{multline}
}
\opt{siims}
{
\begin{equation}
\hat{\theta} =  \arg\min_\theta  \int_\cX  \KLD{Q(\zvar(x);\theta(\x))}{p(y(\x)| \zvar(\x))} d\x + v \lambda \TV[\z] \PD
\end{equation}
}
\mn{Now, for a patch $\cX$ of constant intensity, \ie, {\opt{arxiv}{\small}$\psi(\x) = \psi_0$}, the optimum will be attained
at {\opt{arxiv}{\small}$\mn{\ln ( \theta(x)/(1-\theta(x)) )} = \phi(\x) = \psi_0$} as both the KL and TV terms vanish}. This in turn implies that the regularizer does not interact in homogeneous regions and an unbiased probability  estimate is obtained.

In contrast, other probabilistic segmentation approaches, \eg the Ising model \cite{ising1925beitrag}, lack this ``unbiased in homogeneous regions'' property and  because of this interaction with the regularizer, setting the regularization parameter $\lambda$ in such cases can be tricky. To illustrate this point, consider a VMF treatment of the Ising model that parallels the approach and notation used for AMF. Defining an Ising model where \mn{$N(i)$ are the neighbors of $\x_i$ and the neighborhood potential term is}
{\opt{arxiv}{\small}
\begin{align}
U(\zvar) \triangleq  \lambda \sum_{\indx} \sum_{j\in N(i)} \zvar_i (1 - \zvar_j) + (1 - \zvar_i)\zvar_j,
\end{align}
}
then
\begin{align}
P(\zvar | y) \propto p(y|\zvar) P(\zvar) \text{\;where\;} P(\zvar) \propto  e^{-U(\zvar)} \PD
\end{align}
Using the VMF approximation, we obtain:
\opt{arxiv}
{\opt{arxiv}{\small}
\begin{align}
\hat{\theta} &= \arg \min_\theta \left\{\KLD{Q(\rmZ;\theta)}{p(y|\rmz)} + \lambda \EV{Q(\rmZ;\theta)}{U(\rmZ)}  \right\}
\\
 &= \arg \min_\theta \Bigg\{\KLD{Q(\rmZ;\theta)}{p(y|\rmz)}
 \\ &\qquad + \lambda \sum_i \sum_{j\in N(i)} \theta_i (1 - \theta_j) + (1 - \theta_i)\theta_j   \Bigg\} \CM
\end{align}
}
\opt{siims}
{
\begin{align}
\hat{\theta} &= \arg \min_\theta \left\{\KLD{Q(\rmZ;\theta)}{p(y|\rmz)} + \lambda \EV{Q(\rmZ;\theta)}{U(\rmZ)}  \right\}
\\
 &= \arg \min_\theta \Bigg\{\KLD{Q(\rmZ;\theta)}{p(y|\rmz)} + \lambda \sum_i \sum_{j\in N(i)} \theta_i (1 - \theta_j) + (1 - \theta_i)\theta_j   \Bigg\} \CM
\end{align}
}
which yields the following stationary-point equation:
{\opt{arxiv}{\small}
\begin{align}
\label{vmf-ising}
\phi_i - \psi_i - 4 n \lambda \sum_{j \in N(i)} \left[\theta_j - \frac{1}{2} \right] = 0,\quad n\triangleq |\{j\in N(i)\}|.
\end{align}
}
\mn{This consistency  equation characterizes }the solution of
the VMF approximation to the Ising model.  It is clear from \mn{\Eqn{vmf-ising}}
that the regularization term will only be zero \mn{when the neighborhood average of $\theta_i$ equals $\frac{1}{2}$,} while in other cases the \mn{unbiased property} will not apply.


\section{Experiments}
\label{SecExpts}

This section illustrates the behavior of the proposed AMF model. \mn{\Sec{SecNumericalSolution} describes our numerical solution approach for the ROF model. \Sec{SecAmbiguity} compares the AMF model to the standard Chan-Vese approach when dealing with ambiguous boundaries}. \Sec{SecAgreementWithOriginalModel} investigates how well the AMF model agrees with the original probability model without approximations. \mn{\Sec{SecQualitativeAMF} qualitatively assesses the AMF model on real ultrasound data of the heart and the prostate, as well as on the Fabio image often used for testing in computer vision.} \mn{\Sec{SecQuantitativeAMF} quantitatively analyzes AMF by applying it to the images from the \texttt{icgbench} segmentation benchmark dataset.}


\subsection{\mn{Numerical Solution}}
\label{SecNumericalSolution}

\mn{We indirectly solve the AMF model by relating it to the ROF problem as discussed in \Sec{SecROF}}. \mn{The ROF model was initially solved~\cite{rudin1992nonlinear} using a gradient descent method, and this may still be a reasonable option if AMF solving is embedded in an outer iteration, e.g. expectation-maximization~\mn{\cite{dempster1977}}. The difficulty in computing the optimum of the ROF energy is due to the \mn{TV} term that is not differentiable everywhere. The very first solver changed the  optimization problem by replacing the \mn{TV} term with $\sqrt{|\nabla u|^2+\beta^2}$~\cite{chan1999}, which made the energy function differentiable everywhere. To allow for better discretization of the \mn{TV} term, primal-dual~\cite{chan1999}, and fully dual methods~\cite{chambolle2004} have been explored. More recently, methods based on accelerated proximal gradient descent (\texttt{FISTA})~\cite{beck2009} and split Bregman iterations~\cite{goldstein2009} have been applied to solve the ROF model. \mn{See~\cite{chambolle2016} for a comprehensive overview of recent continuous optimization strategies for the ROF model.}} \mn{We use \texttt{FISTA} for all our following experiments on synthetic and real data.} \mn{To avoid computational issues in our experiments, probabilities were clamped to be in $[10^{-5},1-10^{-5}]$. We used the Matlab \texttt{FISTA} implementation by Amir Beck and Marc Teboulle~\cite{beck2009}. Convergence for \texttt{FISTA} was left at the default value of $10^{-4}$. The maximum number of iteration steps was set to 10,000 but was never reached.}

\subsection{Segmentation with Ambiguity}
\label{SecAmbiguity}

\begin{figure}
  \centering
  \begin{tabular}{cccc}
  \opt{arxiv}
    {
    \includegraphics[height=1.65in]{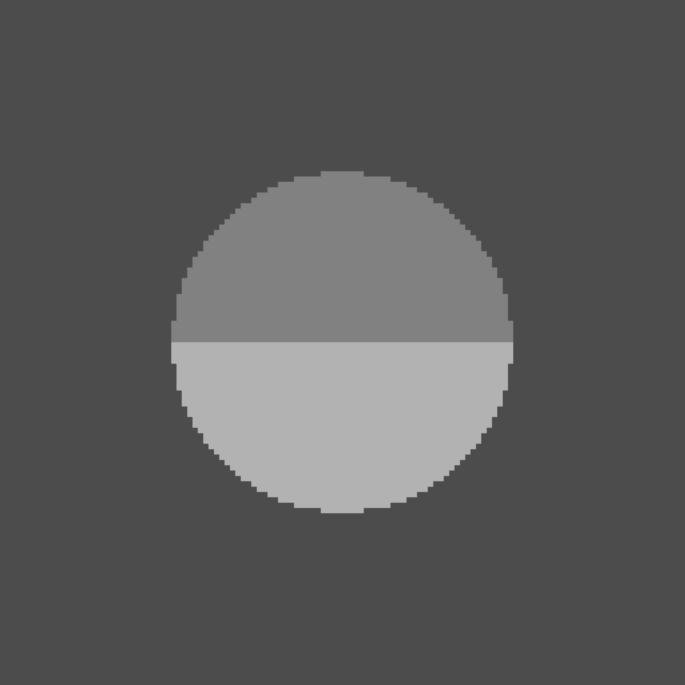} &
    \includegraphics[height=1.65in]{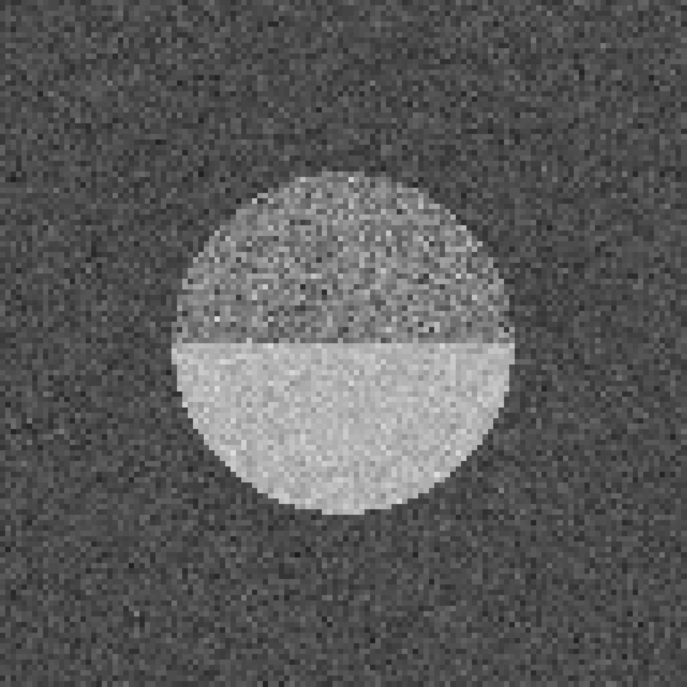} &
    \includegraphics[height=1.65in]{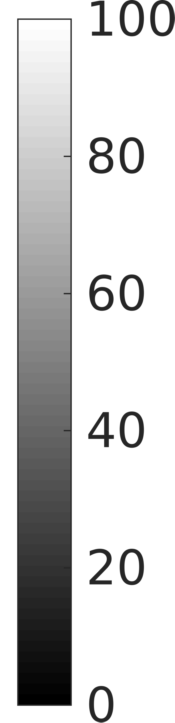} &
    \includegraphics[height=1.65in]{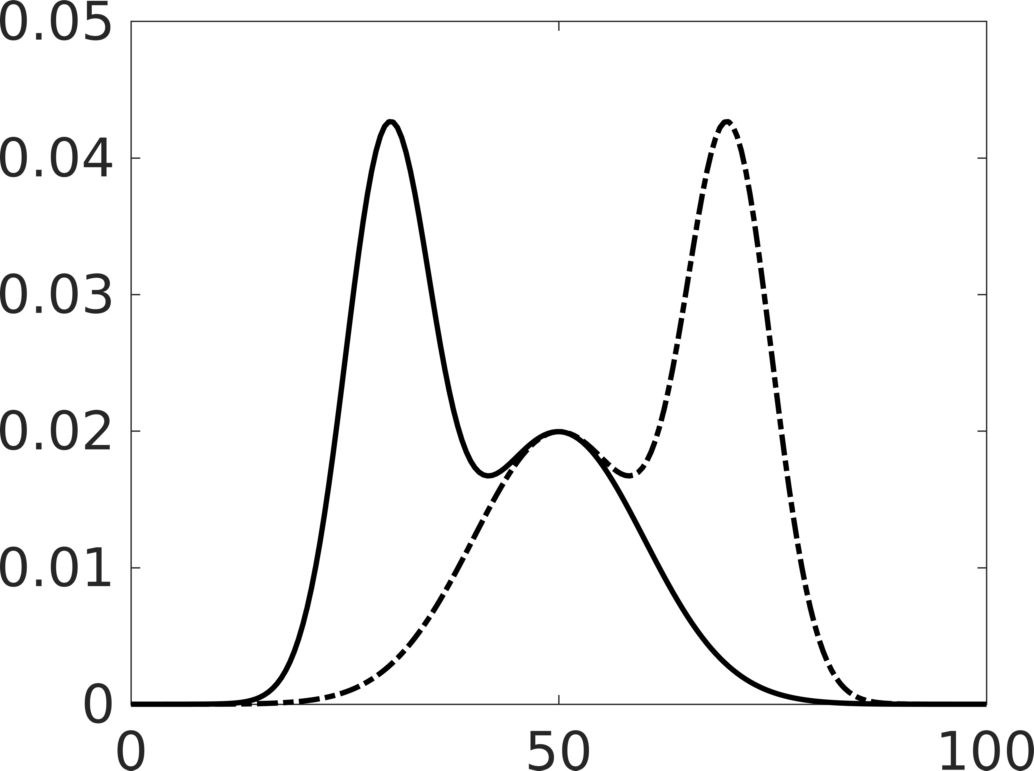}
    }
    \opt{siims}
    {
    \includegraphics[height=1.3in]{cleanImage.png} &
    \includegraphics[height=1.3in]{noisyImage.png} &
    \includegraphics[height=1.3in]{colorbar_for_synthetic_image_gray.png} &
    \includegraphics[height=1.3in]{distribution.png}
    }
  \end{tabular}
  \caption{{Ambiguous segmentation scenario. Left: original image, middle: noisy image, right: class conditional distributions. Distributions clearly overlap which should result in a segmentation ambiguity for the upper part of the circle which was deliberately chosen to have intensities in between the  background and the  foreground (bottom part of the circle). \mn{Background class conditional distribution displayed as a solid black line, foreground class conditional distribution displayed as a dash-dotted black line.}}}
  \label{figAmbiguousImagesAndDistribution}
  ~\\
%
  \centering
\renewcommand{\tabcolsep}{1pt}
\begin{tabular}{ccccc}
    \includegraphics[height=1.2in]{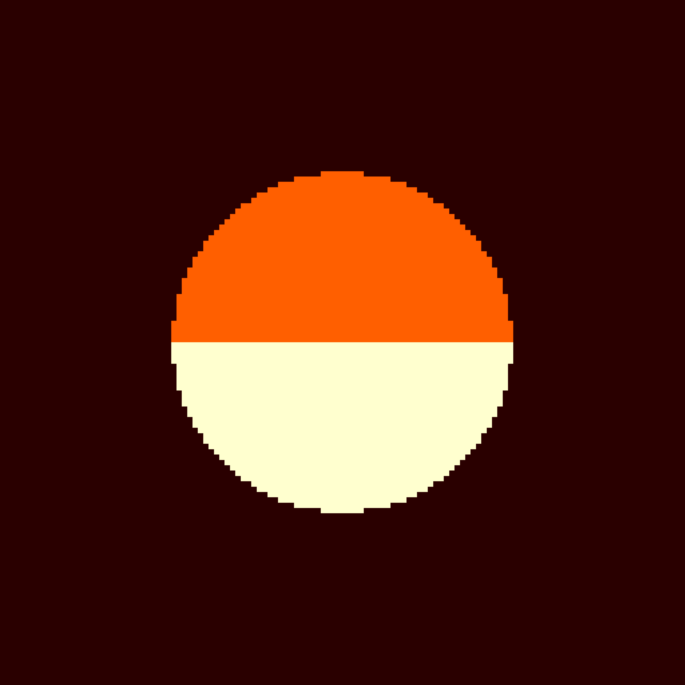} &
    \includegraphics[height=1.2in]{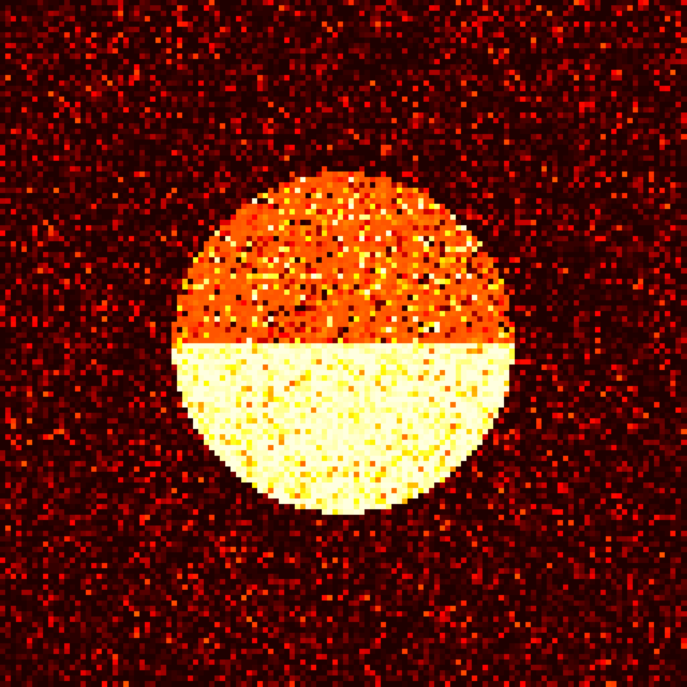} &
    \includegraphics[height=1.2in]{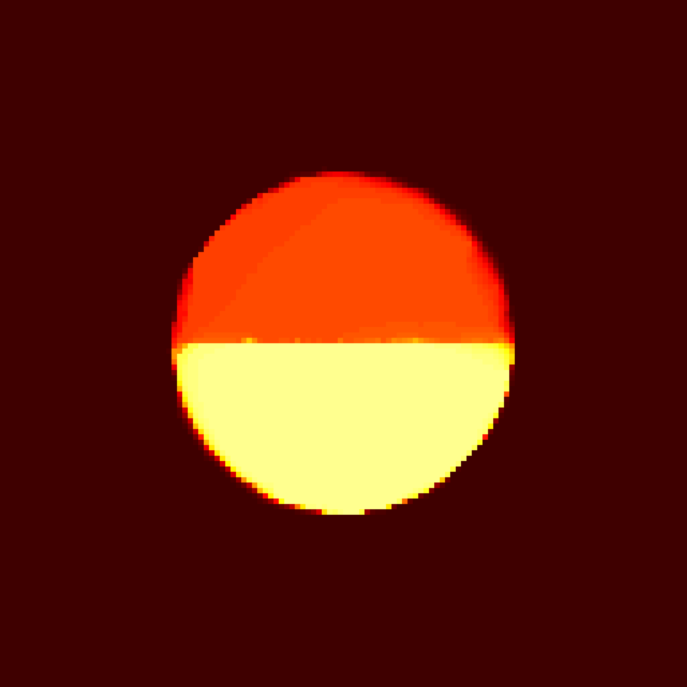} &
    \includegraphics[height=1.2in]{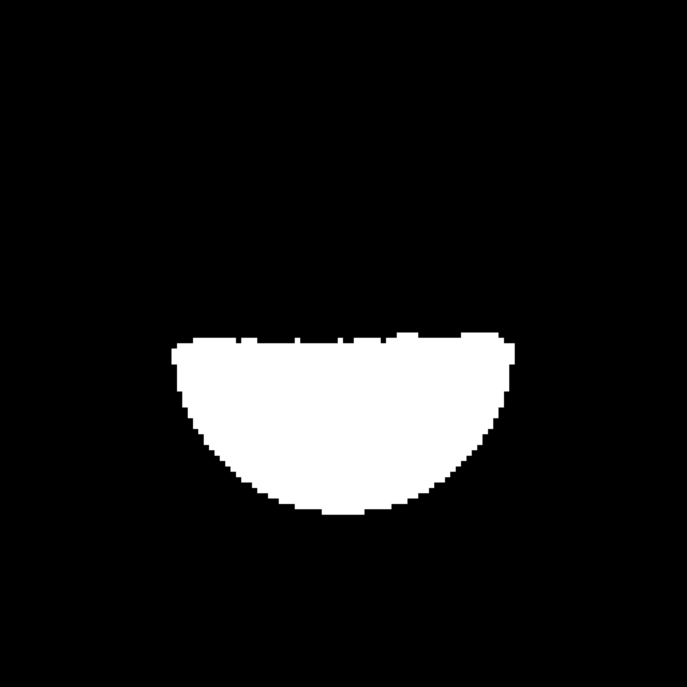} &
    \includegraphics[height=1.2in]{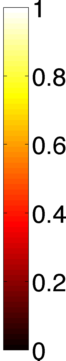}\\
    (a) & (b) & (c) & (d) &
  \end{tabular}
  \renewcommand{\tabcolsep}{6pt}
  \caption{{\mn{(a)} \mn{Noise-free foreground label probabilities based on the noise-free image of~\Fig{figAmbiguousImagesAndDistribution}} (which is not available in practice). (b) \mn{Noisy label probabilities based on the noisy image of~\Fig{figAmbiguousImagesAndDistribution}}. The upper part of the circle is clearly ambiguous with foreground label probability of $P=0.5$.} (c) Estimated label probabilities using the AMF model. (d) Estimated MAP solution (binarization at $P=0.5$) from the AMF-estimated label probabilities. Clearly, the AMF model captures more information -- the MAP solution completely loses the ambiguity of the upper part of the circle.}
  \label{figProbabilities}
  ~\\
%
  \centering
  \begin{tabular}{ccc}
  \opt{arxiv}
  {
    \includegraphics[height=1.65in]{theta.png} &
    \includegraphics[height=1.65in]{p1Clean.png} &
    \includegraphics[height=1.65in]{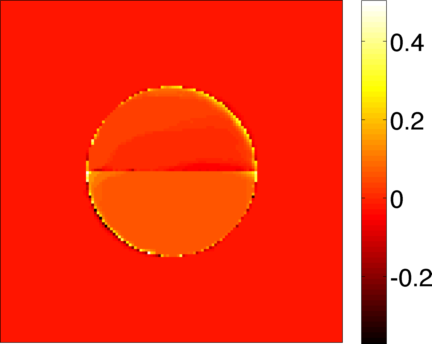}
    }
  \opt{siims}
  {
    \includegraphics[height=1.3in]{theta.png} &
    \includegraphics[height=1.3in]{p1Clean.png} &
    \includegraphics[height=1.3in]{p1Clean_minus_theta.png}
    }
  \end{tabular}
  \caption{Left: estimated label probabilities by the AMF model. Middle: noise-free label probabilities. Right: difference between the probabilities. Differences exist primarily at the segmentation boundaries, which is expected since the AMF model includes spatial regularization effects while the noise-free label probabilities are computed strictly locally. Overall, there is a good agreement between the probabilities.}
  \label{figSubtraction}
\end{figure}

A goal of AMF is to provide label probabilities from which the MAP
solution for the segmentation can be obtained, but which also allow
the assessment of segmentation uncertainty. To test this behavior, we generated
a highly ambiguous segmentation scenario, depicted in
\Fig{figAmbiguousImagesAndDistribution}. \mn{We start by assuming class conditional intensity distributions for the foreground and the background classes (\Fig{figAmbiguousImagesAndDistribution} right). Specifically, the class conditional intensity distributions were obtained as a mixture of Gaussians. We use three Gaussians with means $\mu=\{30,50,70\}$ and corresponding standard deviations $\sigma=\{5,10,5\}$ and mix the first two ($\mu=\{30,50\}$; $\sigma=\{5,10\}$) to obtain the background conditional intensity distribution and the last two ($\mu=\{50,70\}$; $\sigma=\{10,5\}$) to obtain the foreground conditional intensity distribution. In both cases, the mixing coefficients are $0.5$. The intensity distribution of the circle in the center of the image was chosen such that half of the circle has intensities that lie exactly in the
middle between the foreground and background. In particular, the \mn{intensity of the region outside the circle} is $\mu=30$, the intensity of the upper part of the circle is $\mu=50$, and the intensity of the lower part of the circle is $\mu=70$. Gaussian noise with mean zero and standard deviation of $\sigma=5$ was added to the background, $\sigma=10$ to the upper part of the circle, and $\sigma=5$ to the lower part of the circle. 
The results were obtained by assuming we know the \mn{conditional distributions} for the foreground and background classes;  likelihoods
were computed based on the noisy data. The regularization term was
weighted with $\lambda=5$.}

\Fig{figProbabilities} (\mn{left} two images) shows the local label probabilities for the noisy input image and for the noise-free image (that will not be available in practice). \Fig{figProbabilities} (\mn{right} two images) shows the label probabilities after running the AMF model (left) and after thresholding (binarization) at $P=0.5$ (right) that also corresponds to the MAP solution. \mn{Note that neither the foreground probability is one nor the foreground probability is zero due to the chosen class conditional intensity distributions: both the means of the background ($\mu=30$) and the foreground ($\mu=70$) have non-zero likelihood for background {\it and} foreground.} As desired, the AMF model captures the segmentation uncertainty by estimating the upper part of the circle at a probability close to $P=0.5$. At the same time, due to spatial regularization, the AMF model removes noise effects. The MAP solution captures the most likely foreground area, but completely loses the ambiguous area.

\Fig{figSubtraction} shows the estimated label probabilities and their true local counterparts along with a subtraction. The AMF method has effectively estimated the true label probabilities. Note that the true local label probabilities do not incorporate the effect of regularization. Hence, these two probabilities will slightly disagree at the segmentation boundaries.

\subsection{{Agreement with the Original Probability Model}}
\label{SecAgreementWithOriginalModel}

\begin{figure*}[ht]
  \centering
  \begin{tabular}{ccc}
     \includegraphics[width=0.3\textwidth]{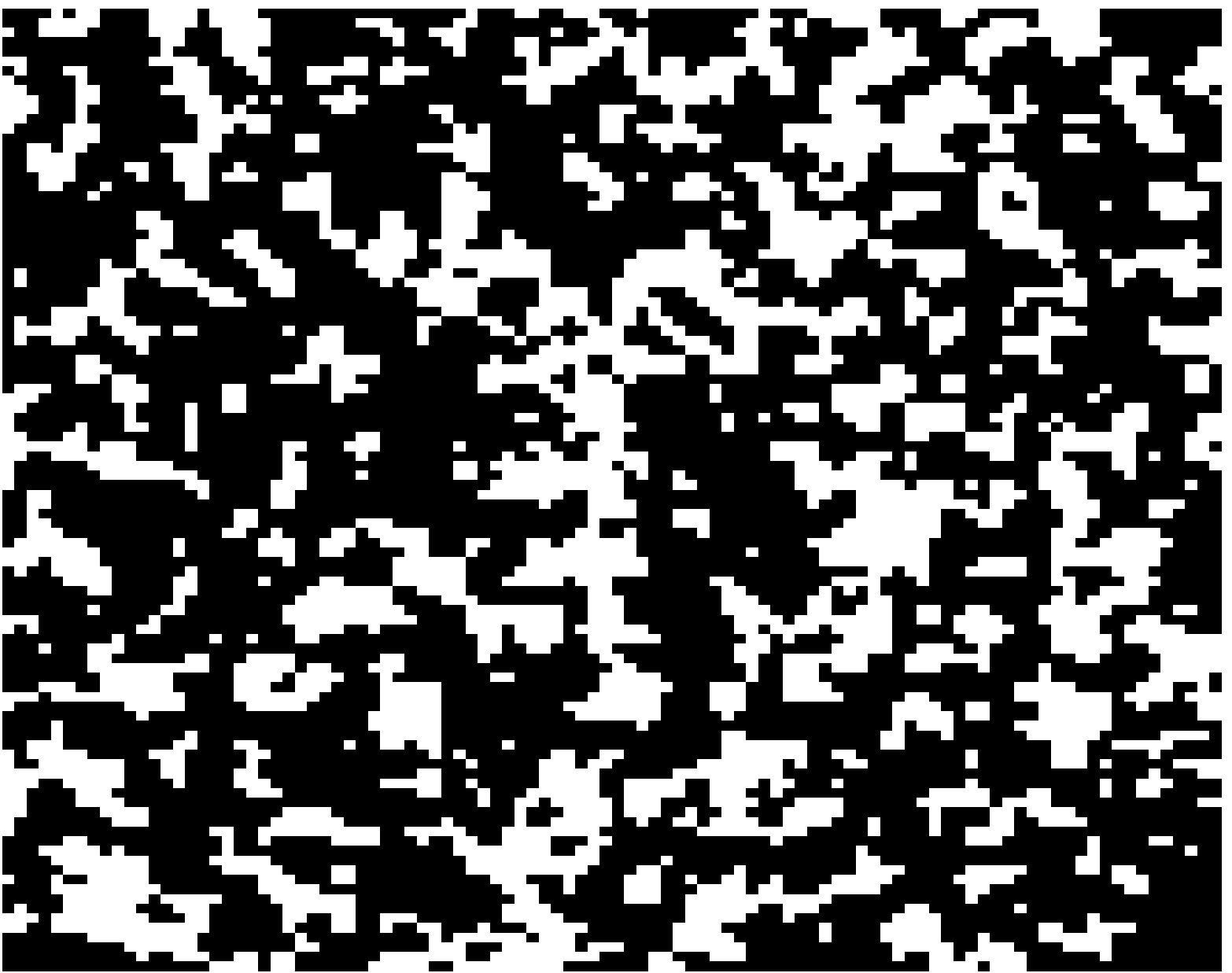} &
     \includegraphics[width=0.3\textwidth]{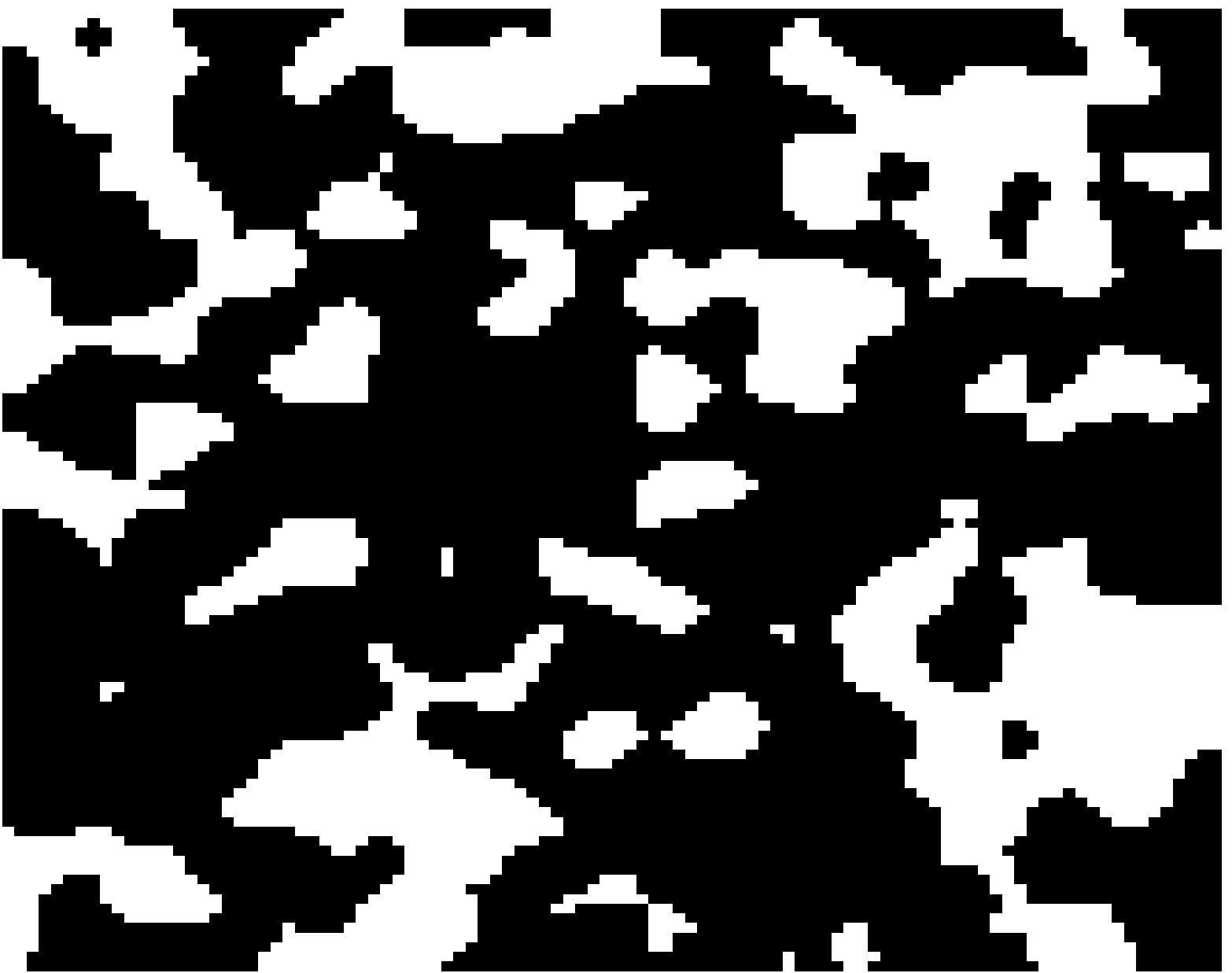} &
     \includegraphics[width=0.3\textwidth]{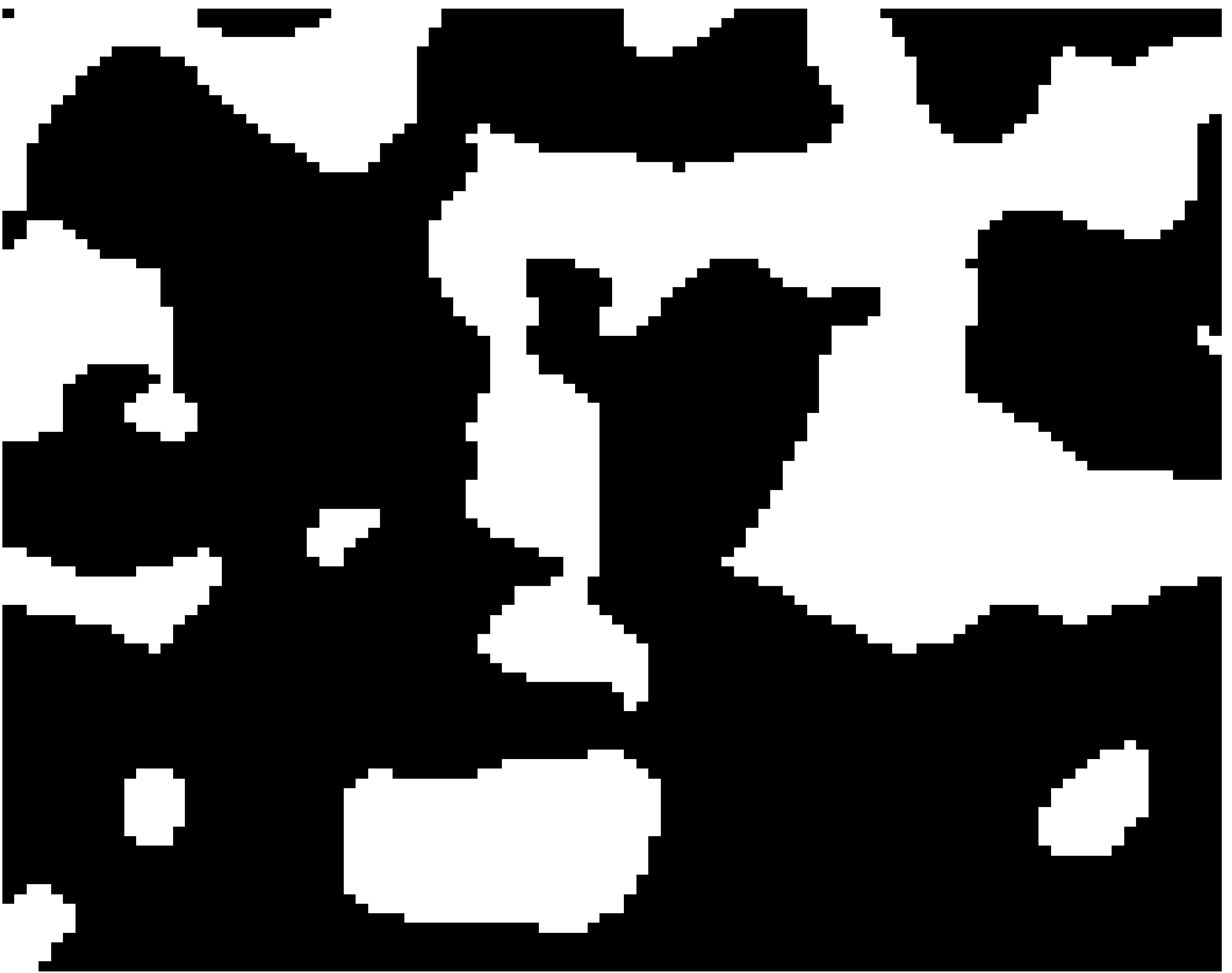} \\
  \opt{arxiv}
  {
     (a) True Label Map : $l=1$ &     (b) True Label Map : $l=3$ &     (c) True Label Map : $l=5$  \\
  }
  \opt{siims}
  {
     {\footnotesize (a) True Label Map : $l=1$} & {\footnotesize (b) True Label Map : $l=3$} & {\footnotesize (c) True Label Map : $l=5$}  \\
  }

     \includegraphics[width=0.3\textwidth]{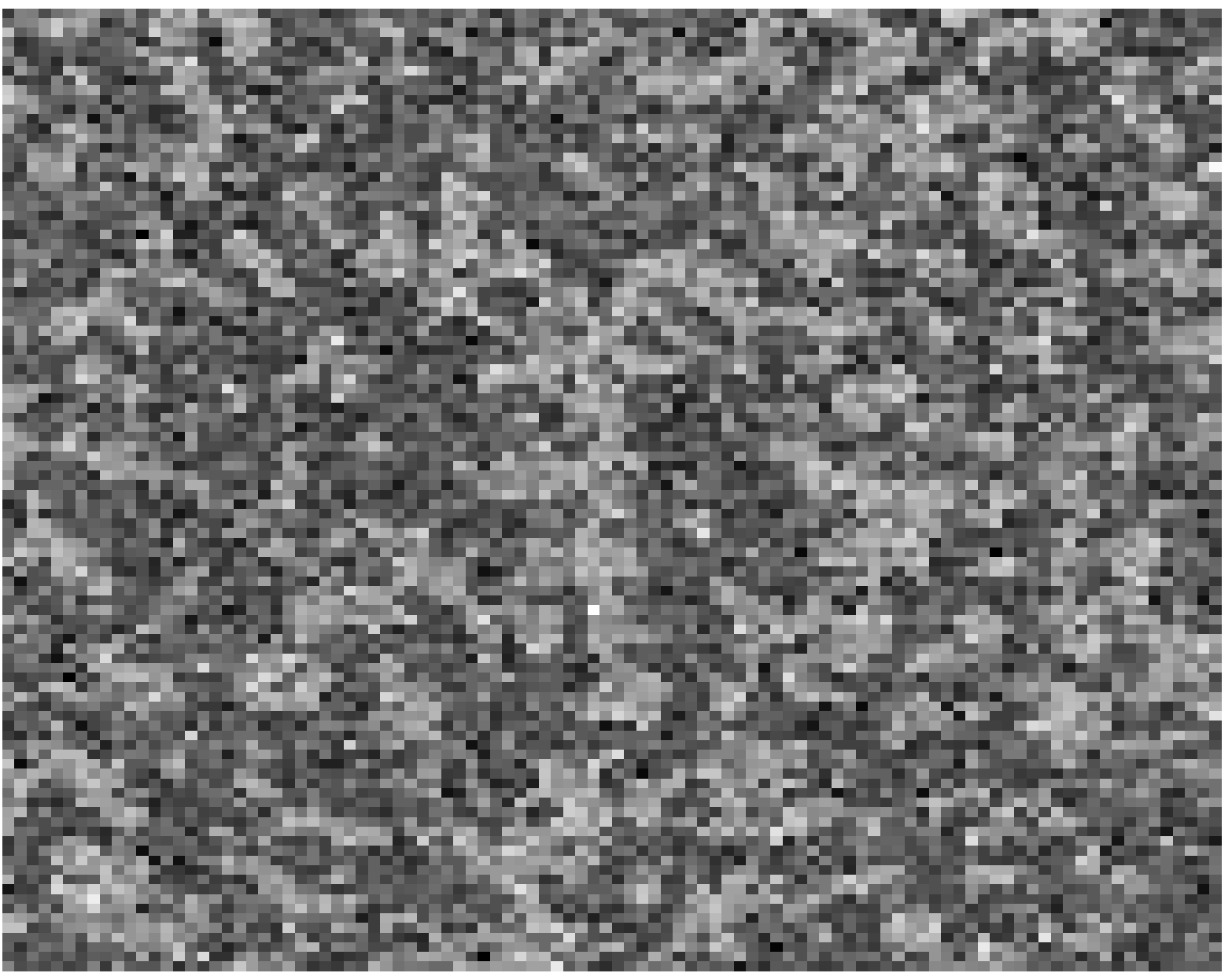} &
     \includegraphics[width=0.3\textwidth]{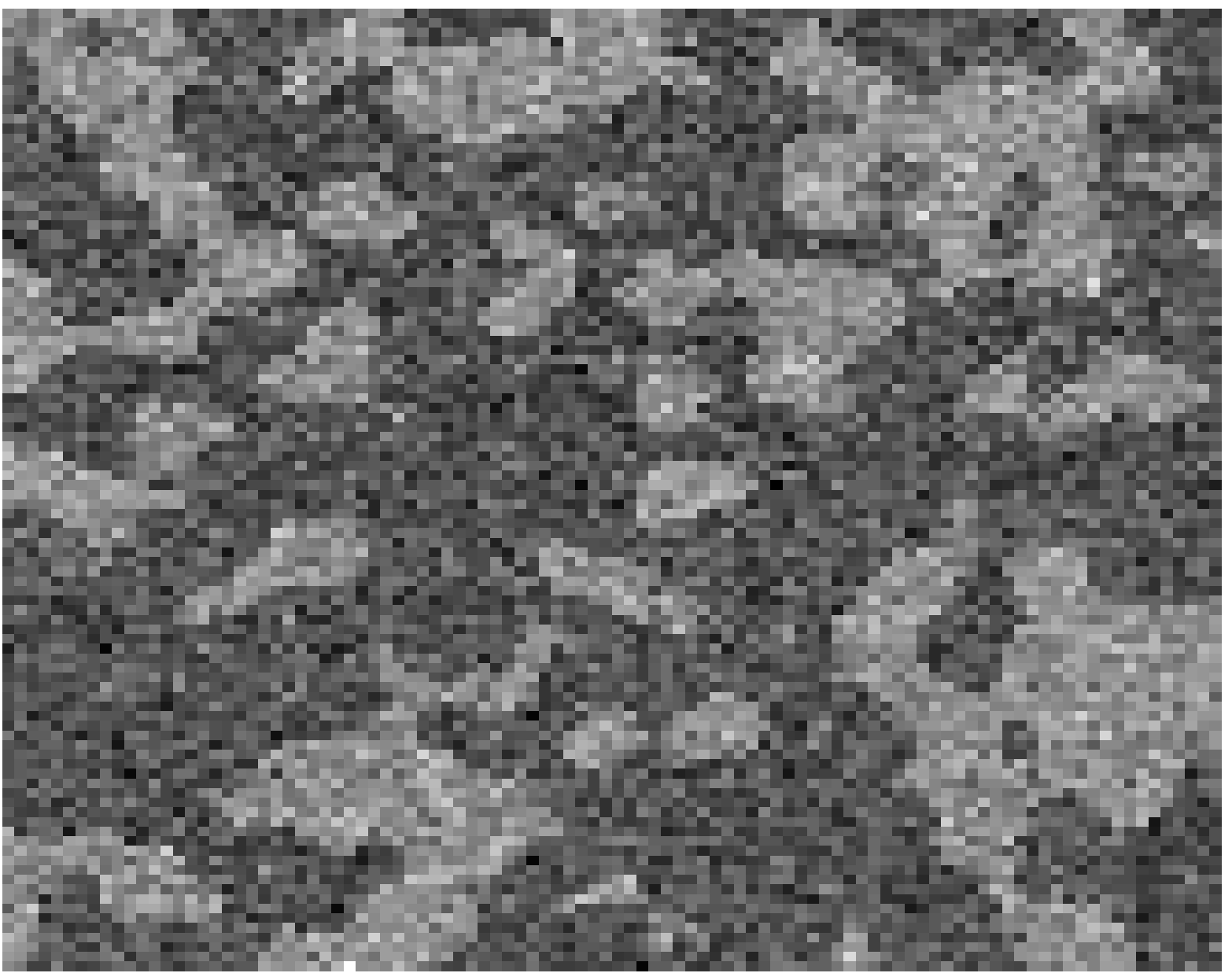} &
     \includegraphics[width=0.3\textwidth]{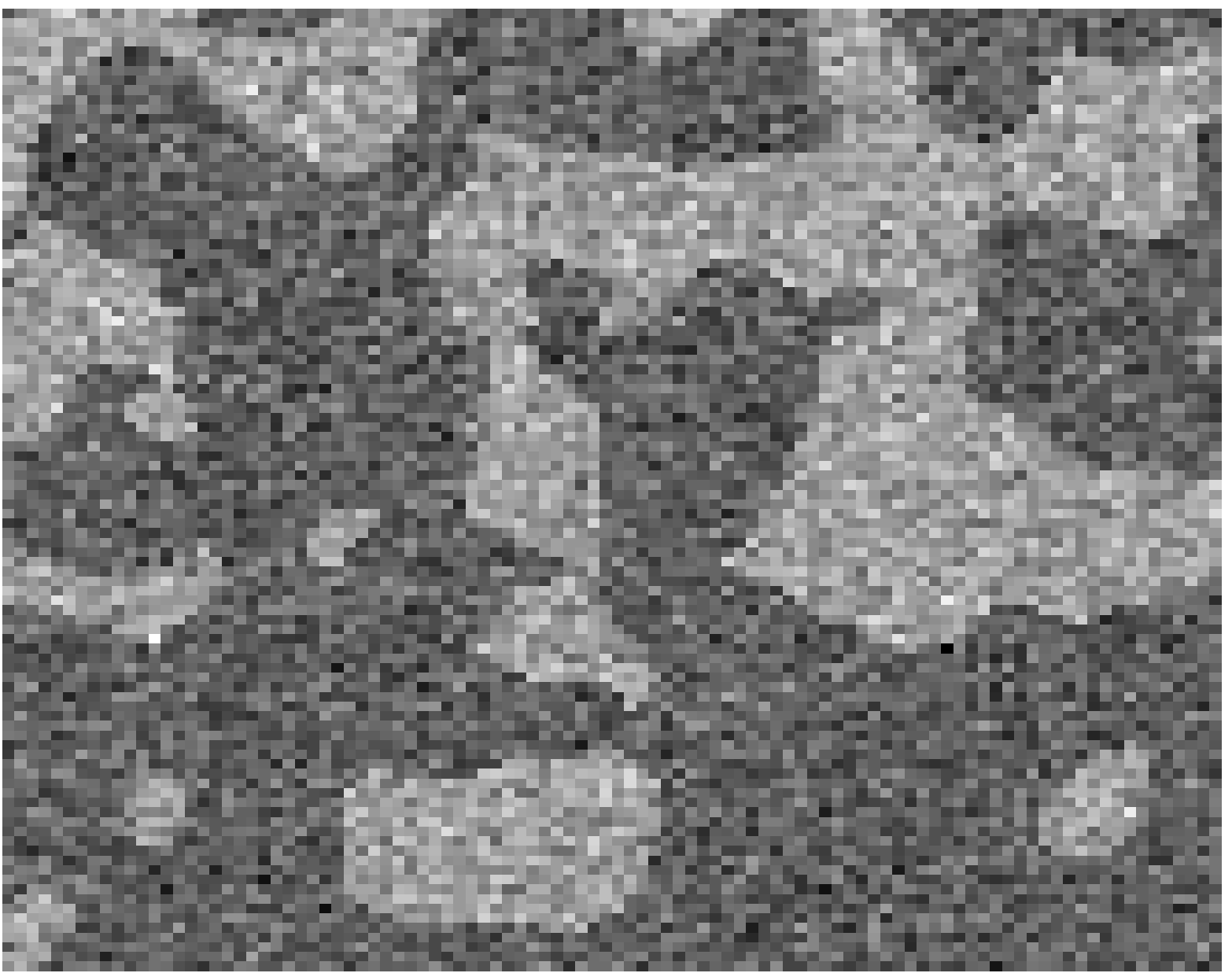} \\
  \opt{arxiv}
  {
     (d) Noisy Image: $l = 1, \sigma=0.4$ & (e) Noisy Image: $l = 3, \sigma=0.4$ & (f) Noisy Image: $l = 5, \sigma=0.4$ \\
  }
  \opt{siims}
  {
     {\footnotesize (d) Noisy Image: $l = 1, \sigma=0.4$} & {\footnotesize (e) Noisy Image: $l = 3, \sigma=0.4$} & {\footnotesize (f) Noisy Image: $l = 5, \sigma=0.4$} \\
  }

  \end{tabular}
  \caption{Single realizations of ground truth label maps and corresponding noisy images generated from Mat\'{e}rn  processes with length-scale parameter varied between 1, 3 and 5.}
  \label{fig:simulation}
\end{figure*}

In order to evaluate agreement between the original probability model, \Eqn{eq:Bayes-rule2}, \ie
\opt{arxiv}
{
  \begin{multline}\label{eqn:prob-orig}
   \ln P(\zvar | y) = v^{-1} \sum_{i\in\cX} \zvar_i \psi_i + \ln p(\y_i|0)\; d\x  \\ - \lambda L(\zvar)  + \const.
\end{multline}
}
\opt{siims}
{
  \begin{equation}\label{eqn:prob-orig}
   \ln P(\zvar | y) = v^{-1} \sum_{i \in \cX} \left[ \zvar_i \psi_i + \ln p(\y_i|0)\;\right]  - \lambda L(\zvar)  + \const.
\end{equation}
}
and {the AMF approximation, we conducted the following set of experiments on synthetic images.} A binary random field was generated by sampling on a 100$\times$100 grid from a Gaussian process with Mat\'{e}rn covariance function \cite{Cressie1991} with order parameter $p$ and scale parameter $l$, that provides fine-grained control over the smoothness of the field. This continuous valued image was then thresholded at a quantile value selected uniformly at random to create the \emph{ground truth binary label map} $\hat{\z}$ to which Gaussian noise is added to create a noisy image $\y$. For our experiments, we set the order parameter $p=1$ while varying the length scale parameter $l=1,3,5$ and the noise standard deviation $\sigma$ between $0.25$ and $0.4$. Increasing $l$ produces label maps with smoother boundaries and larger contiguous regions. Single realizations of $\hat{\z}$ for $l=1,3,5$ are shown in \Fig{fig:simulation}(a--c). Corresponding noisy images for $\sigma = 0.4$ are shown \Fig{fig:simulation}(d--f).

\begin{figure}
    \centering
    \includegraphics[trim=5cm 0cm 5cm 0cm,width=\textwidth]{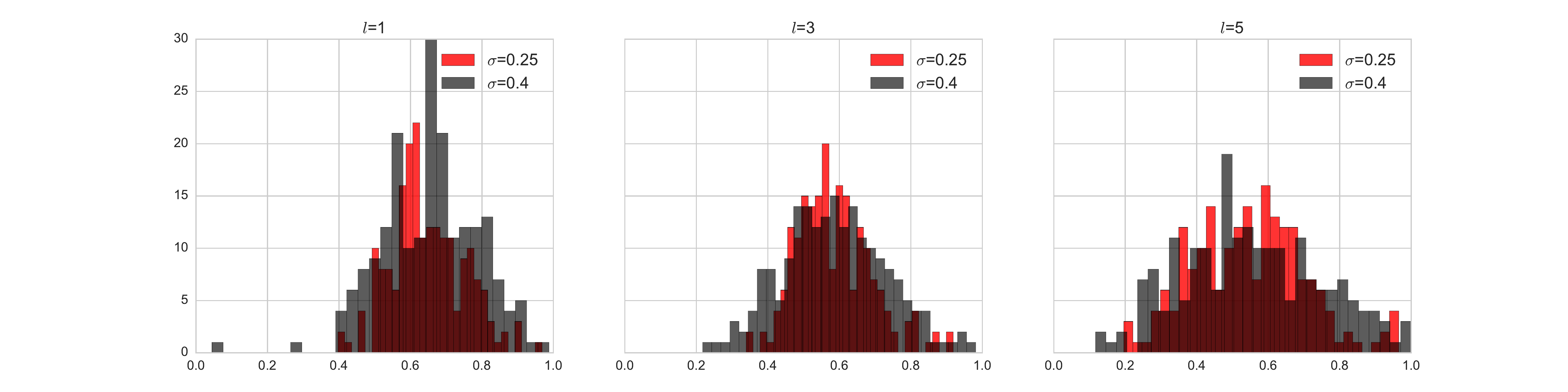}
    \caption{\mn{Histograms of the correlation coefficient between the posterior probability of the Gibbs samples as measured by $P$ and $Q$. Each histogram is across the realizations of the synthetic binary maps and noisy images, \ie, one correlation coefficient per pair, for the various settings of  Mat\'{e}rn  length scale parameter $l$ and image noise $\sigma$.}}
    \label{fig:materna_correlation}
    \includegraphics[trim=5cm 0cm 5cm 0cm,width=\textwidth]{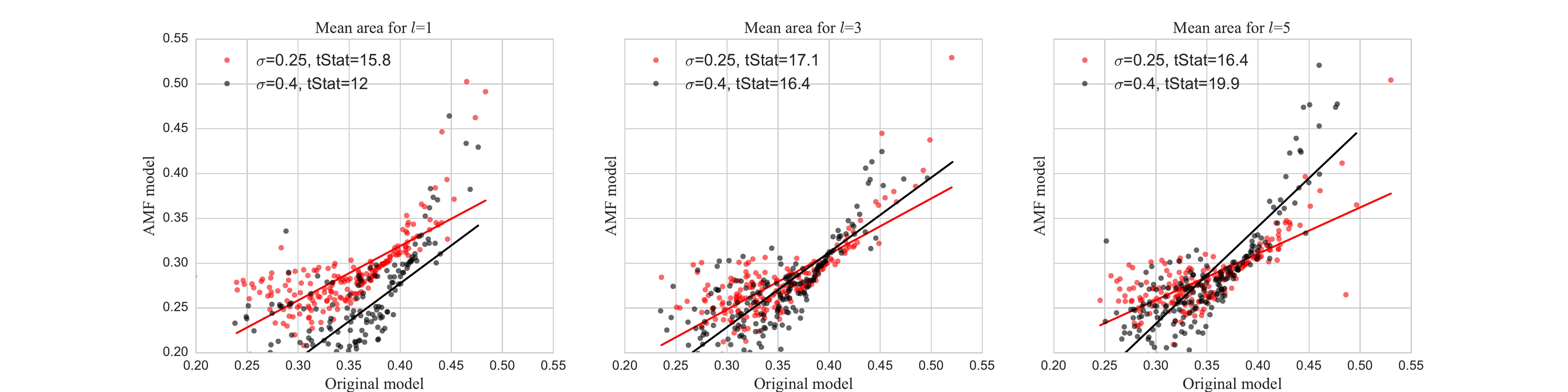}
    \caption{\mn{Scatter plots of the mean foreground area (as a fraction of total area) as measured under $P$ (via Gibbs sampling) and $Q$ (in closed form). Each point is one realization of the synthetic binary maps and noisy images for the various settings of  Mat\'{e}rn  length scale parameter $l$ and image noise $\sigma$.}}
    \label{fig:materna_mean}
    \includegraphics[trim=5cm 0cm 5cm 0cm,width=\textwidth]{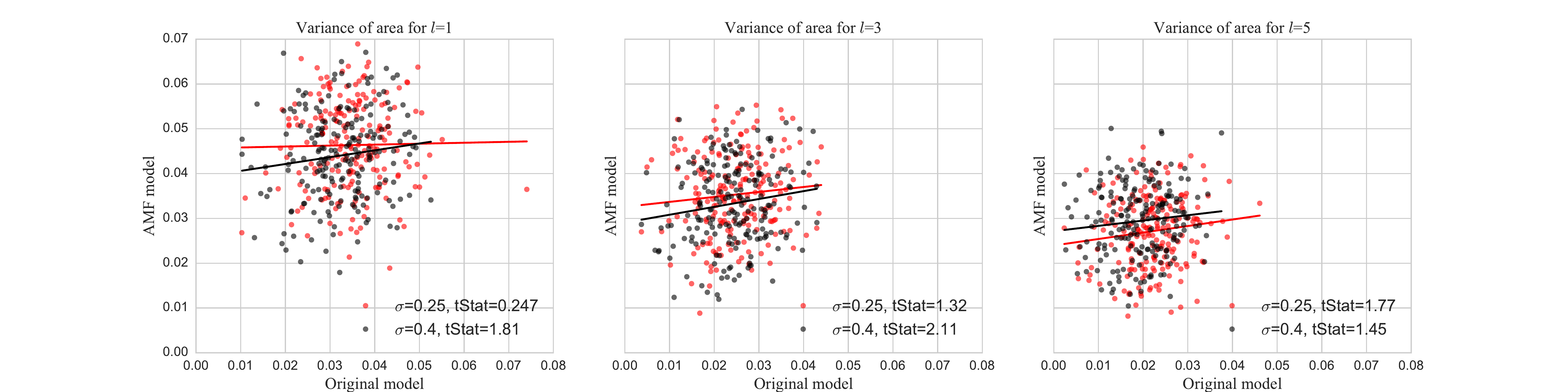}
    \caption{\mn{Scatter plots of the variance of the fractional foreground area as measured under $P$ (via Gibbs sampling) and $Q$ (in closed form). Each point is one realization of the synthetic binary maps and noisy images for the various settings of  Mat\'{e}rn  length scale parameter $l$ and image noise $\sigma$.}}
    \label{fig:materna_var}
\end{figure}

\mn{For each setting of Mat\'{e}rn length scale $l$ we generated 40 ground-truth binary label maps, and for each binary map we generated 5 noisy images at each noise level $\sigma$. Next, for every realized pair of binary and noisy images $(\zvar,\y)$, the AMF approximating distribution $Q$ was computed by solving the ROF equations on the logit maps of $\y$.} The original probability model $P$ of \Eqn{eqn:prob-orig} was also explored using Gibbs sampling with 5 chains of $N=10^5$ particles each, temperature $T=1$ and thinning factor=0.1.  \mn{The temperature parameter, which controls the scale of the sampling distribution, is needed because the probability distribution $P$ is known only up to to a scale factor (i.e. the partition function). Therefore, the Gibbs probability of $z_i = 1$ is $\exp(-e_1/T) /( \exp(-e_1/T)  +   \exp(-e_0/T)  )$, where $e_1$ and $e_0$ are the energies corresponding to $z_i=1$ and $z_i=0$ respectively.} Convergence was tested using the Gelman and Rubin diagnostic~\cite{gelman1992} resulting in approximately $2\times10^4$ particles being retained. \mn{Based on these Monte-Carlo particles, the following statistics were calculated for each realized image pair $(\zvar,\y)$:
\begin{itemize}
    \item {\it The correlation coefficient between the probability masses of each particle according to $P$ and $Q$.} Note that although both $P$ and $Q$ are known only up to scales, it does not affect the correlation coefficient computation. \mn{As shown in Fig.~\ref{fig:materna_correlation} we see a strong correlation between the label map probabilities as estimated by AMF and the original model. This implies that the AMF model is a good approximation to the original probability model. However the correlation seems to reduce with increase in $l$ and $\sigma$, implying that smoother images are harder to approximate - probably because of an increase of non-local interactions that cannot be well approximated by the mean-field distribution and that increasing noise causes greater mis-approximation.}
    \item {\it The mean area of the label map estimated for $P$ from the Gibbs samples and for $Q$ by closed-form evaluation.} \mn{As shown in Fig.~\ref{fig:materna_mean}  the AMF model appears to underestimate the foreground's mean-area when it is less than 50\% of the full image, but this underestimation improves as the foreground fraction increases. Nevertheless there is good agreement, in terms of trend, between the mean area as estimated by the AMF model (in closed form) and the original probability model (via Gibbs sampling).}
    \item {\it The variance in the area of the label map, again estimated for $P$ from the Gibbs samples and for $Q$ by closed-form evaluation.} \mn{As seen in Fig.~\ref{fig:materna_var}, the second order statistics are not captured well by the AMF when compared to the second order statistics of the original model (as assessed by Gibbs sampling); especially for images with larger levels of smoothness. This is not surprising given that the mean-field approximation does not capture higher order interactions of the random field.}
\end{itemize}
}
\mn{In summary, the posterior distribution of the AMF model correlates well with the posterior distribution obtained by Gibbs sampling. The obtained segmentation areas for the AMF model have the same trend as for Gibbs sampling, but tend to underestimate the segmentation area. As expected, higher-order statistics are not captured well due to the simplicity of the factorized variational distribution $Q$ of the AMF model.} 

\subsection{Assessment of AMF on Real Data}
\label{SecRealData}

\mn{We illustrate the behavior of the AMF approach on real images qualitatively in \Sec{SecQualitativeAMF} and quantitatively in \Sec{SecQuantitativeAMF}.} Our goal in this section is not to beat state-of-the art segmentation methods for our example segmentation applications (which may for example, use shape models or more sophisticated machine learning approaches to improve segmentation results), but to illustrate the AMF approach in the context of challenging datasets. Note, however, that the AMF model can be based on any foreground and background likelihood map. Therefore, it is able to augment other more sophisticated pre-processing to obtain foreground and background probabilities.

\subsubsection{Qualitative Assessment of AMF on Real Data}
\label{SecQualitativeAMF}

\mn{We use ultrasound images of the prostate and the heart as well as an image of Fabio~\cite{needell2013} to demonstrate the behavior of AMF under different levels of regularization.  We limit ourselves to simple intensity distributions for the Fabio and the heart ultrasound image. We use a classifier supporting probabilistic outputs based on image intensities for the prostate example. Image size for the prostate example is 257 by 521 pixels, for the heart example 314 by 350 pixels, and for the Fabio image 253 by 254 pixels.}

\begin{figure}
  \centering
  \begin{tabular}{ccc}
    \includegraphics[width=0.3\textwidth]{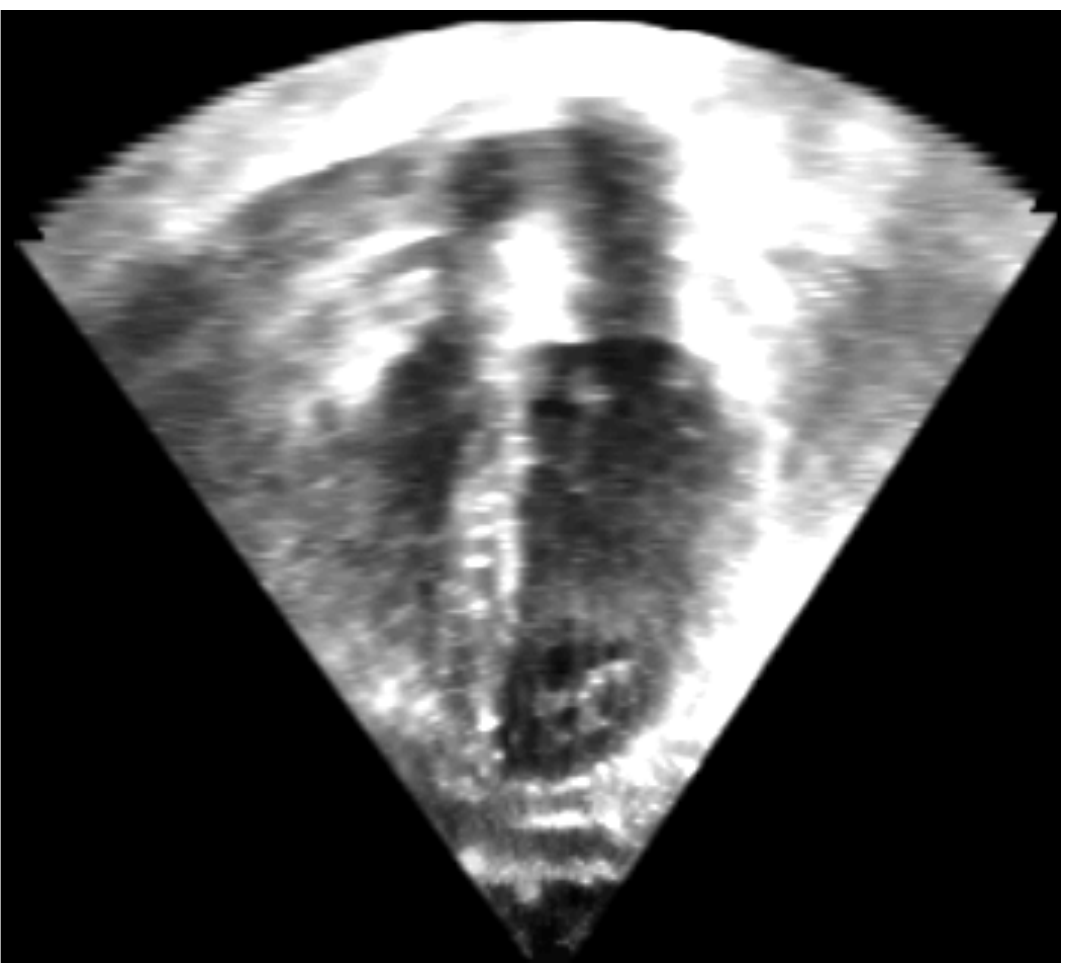} &
    \includegraphics[width=0.3\textwidth]{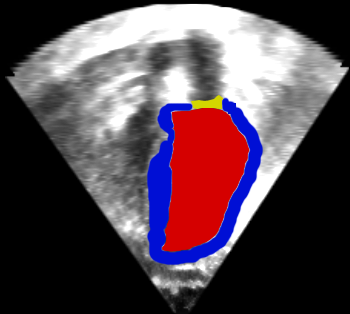} &
    \includegraphics[width=0.3\textwidth]{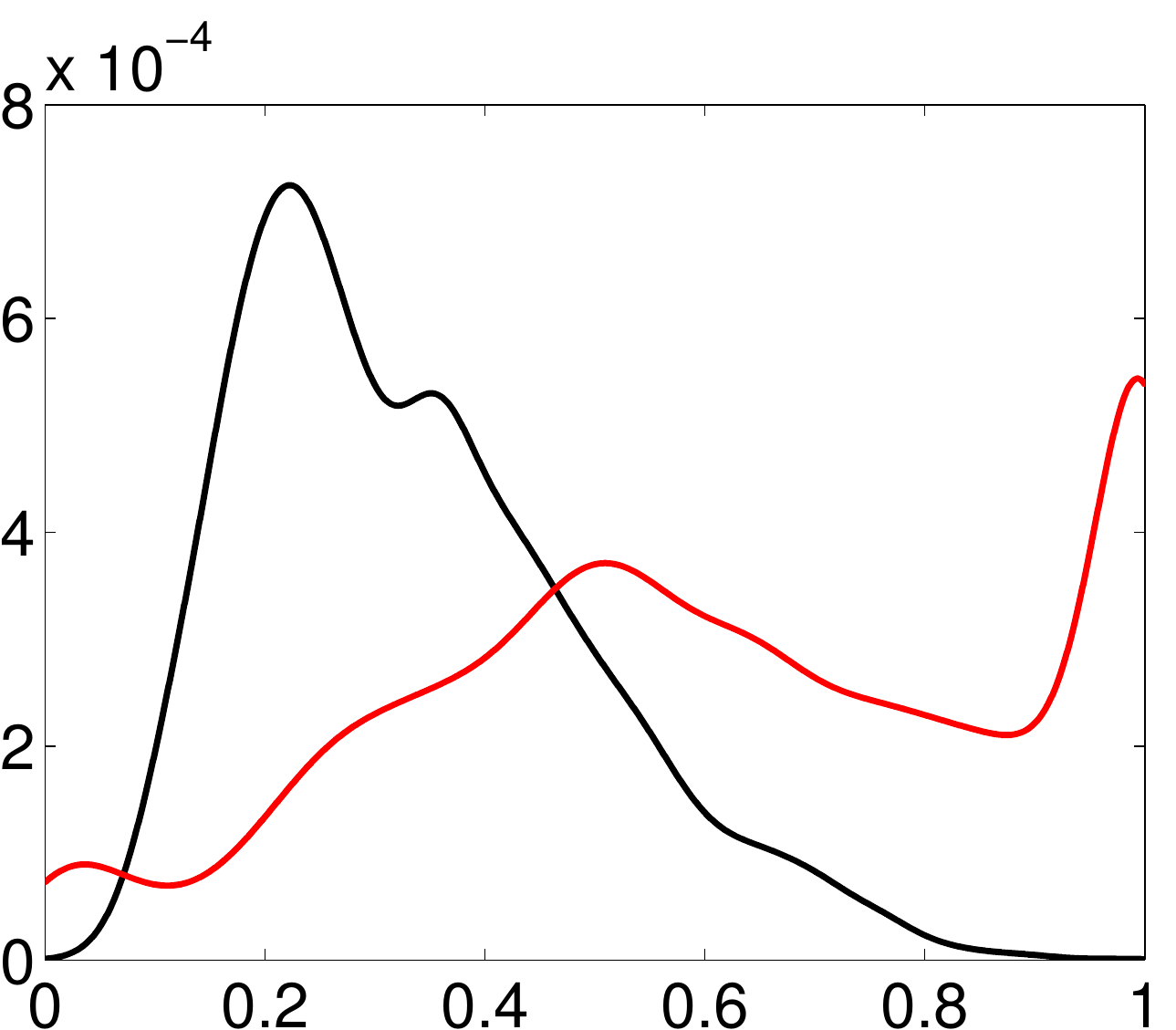}
  \end{tabular}
  \caption{{Left: ultrasound image of the heart. Middle: ultrasound image of the heart with overlaid expert segmentations of blood pool (red), myocardium (blue) and valves (yellow). Right: intensity distributions for the blood pool (red) and the areas outside of the heart (black) for the intensity-normalized image ($I\in[0,1]$). Intensity distributions clearly overlap making an intensity-only segmentation challenging.}}
  \label{FigHeartUS}
%
\renewcommand{\tabcolsep}{0pt}
  \centering
  \begin{tabular}{ccccc}
    \includegraphics[width=0.2\textwidth]{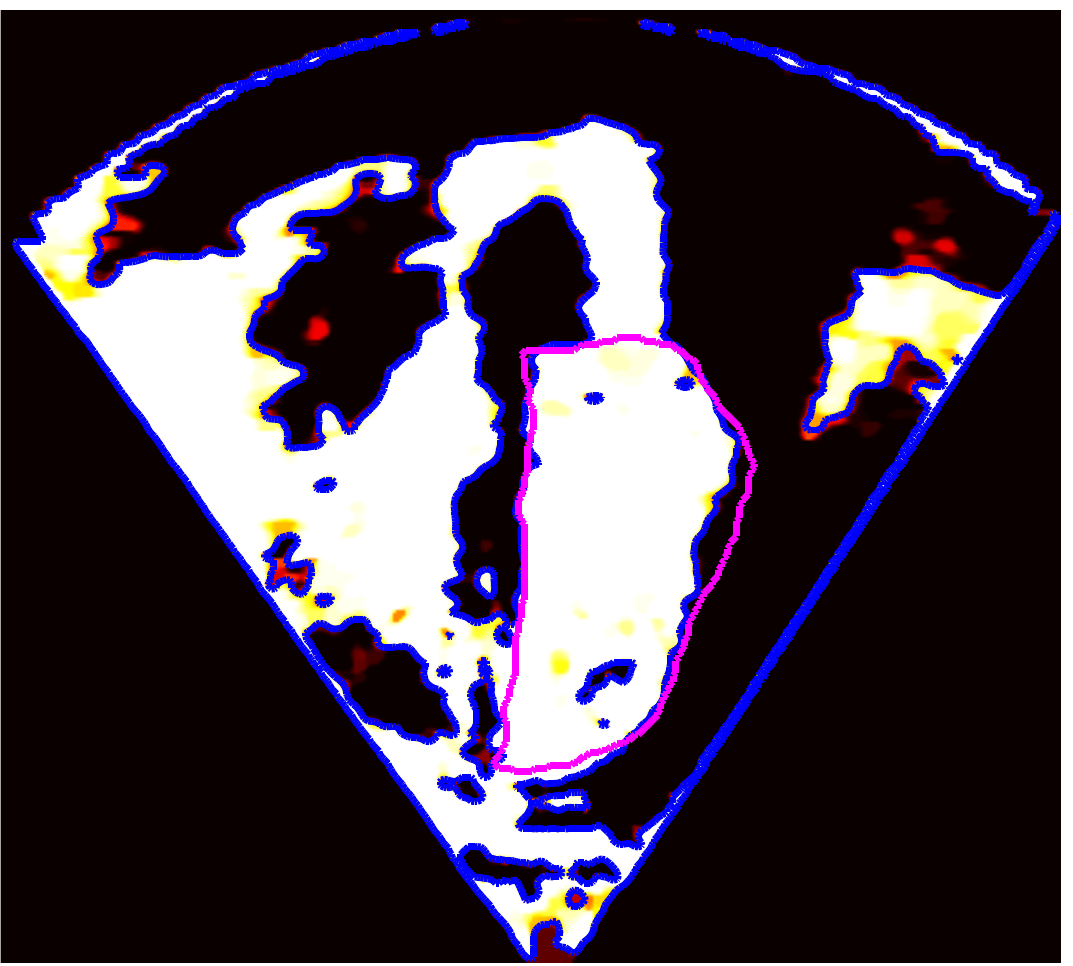} &
    \includegraphics[width=0.2\textwidth]{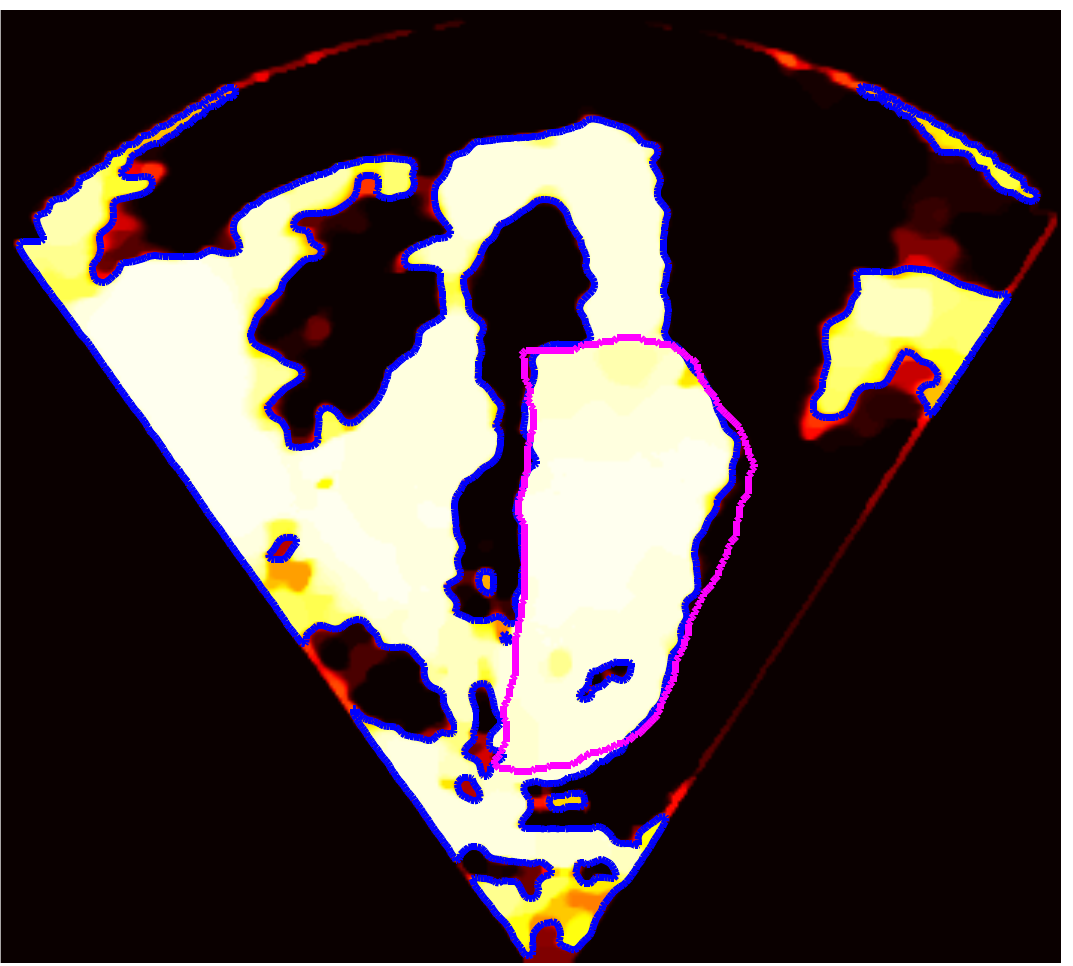} &
    \includegraphics[width=0.2\textwidth]{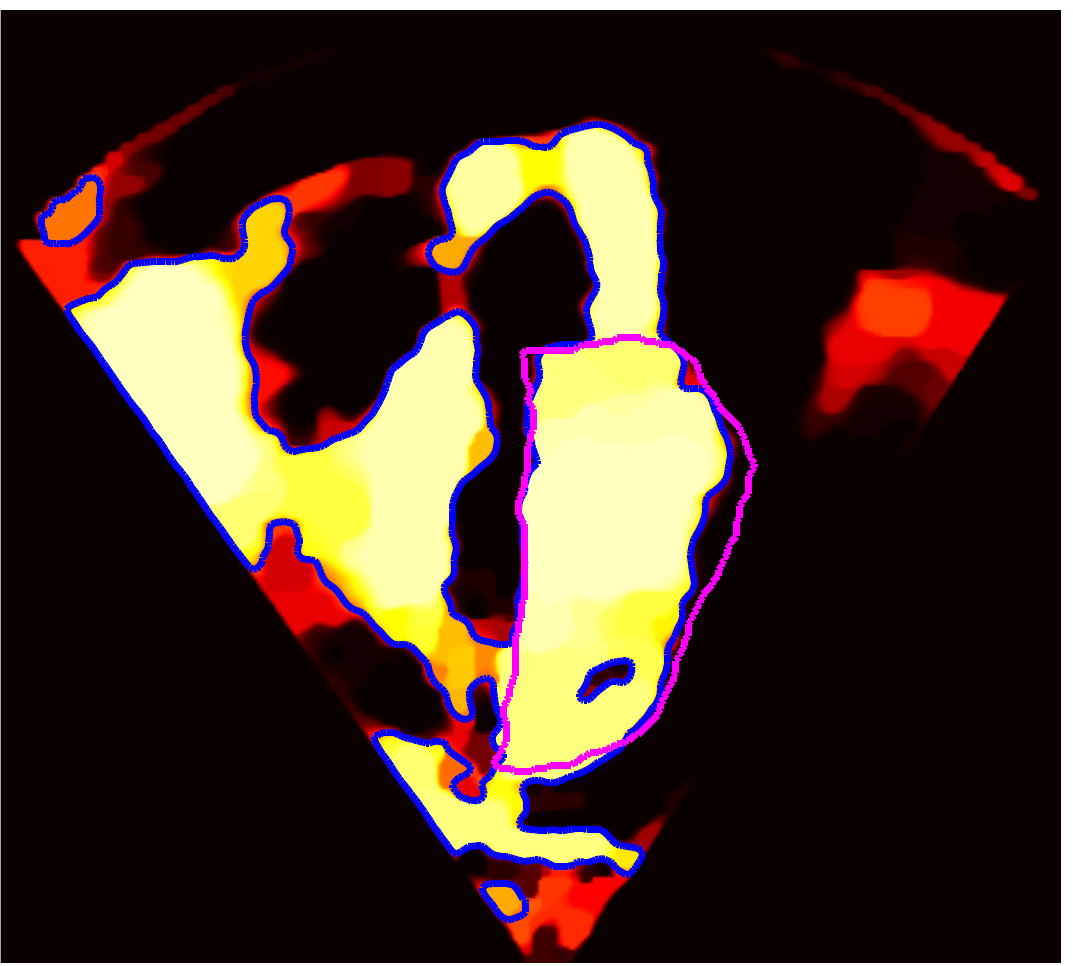} &
    \includegraphics[width=0.2\textwidth]{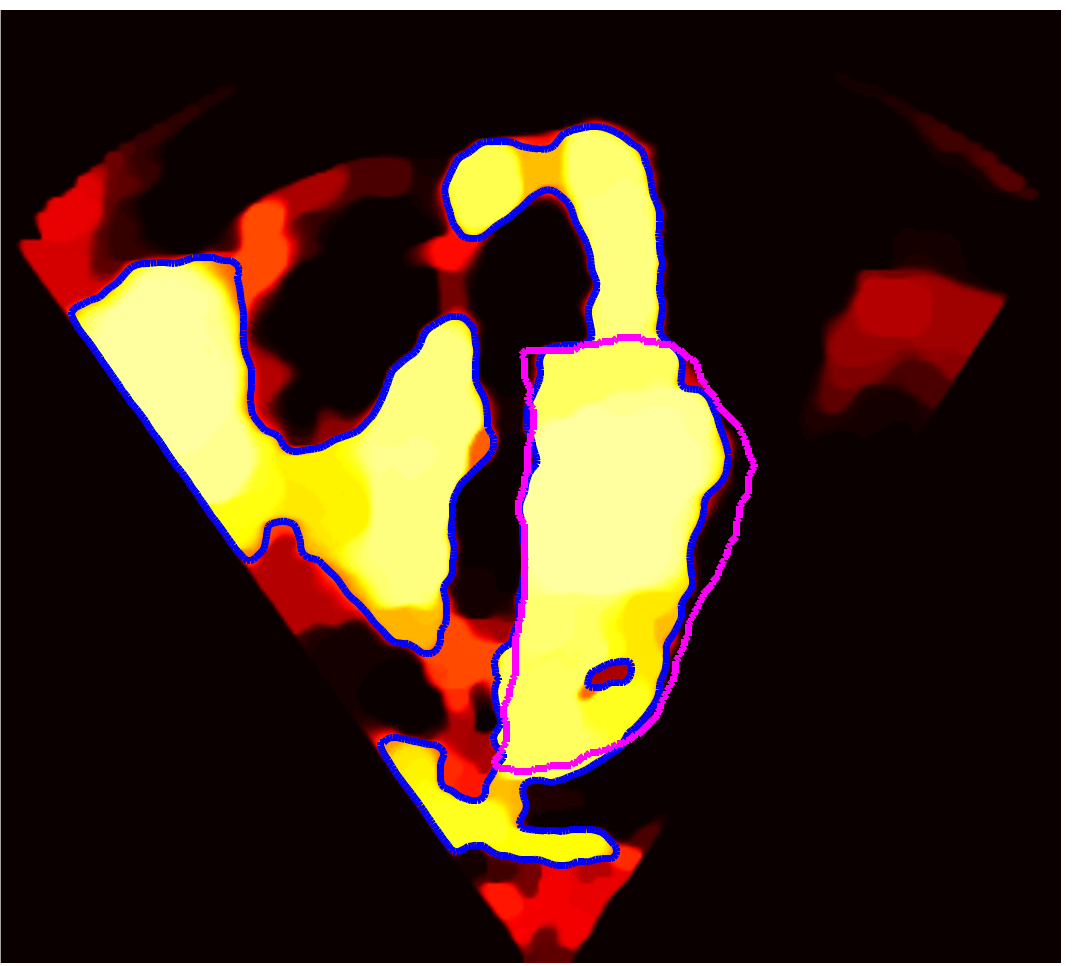} &
    \includegraphics[width=0.2\textwidth]{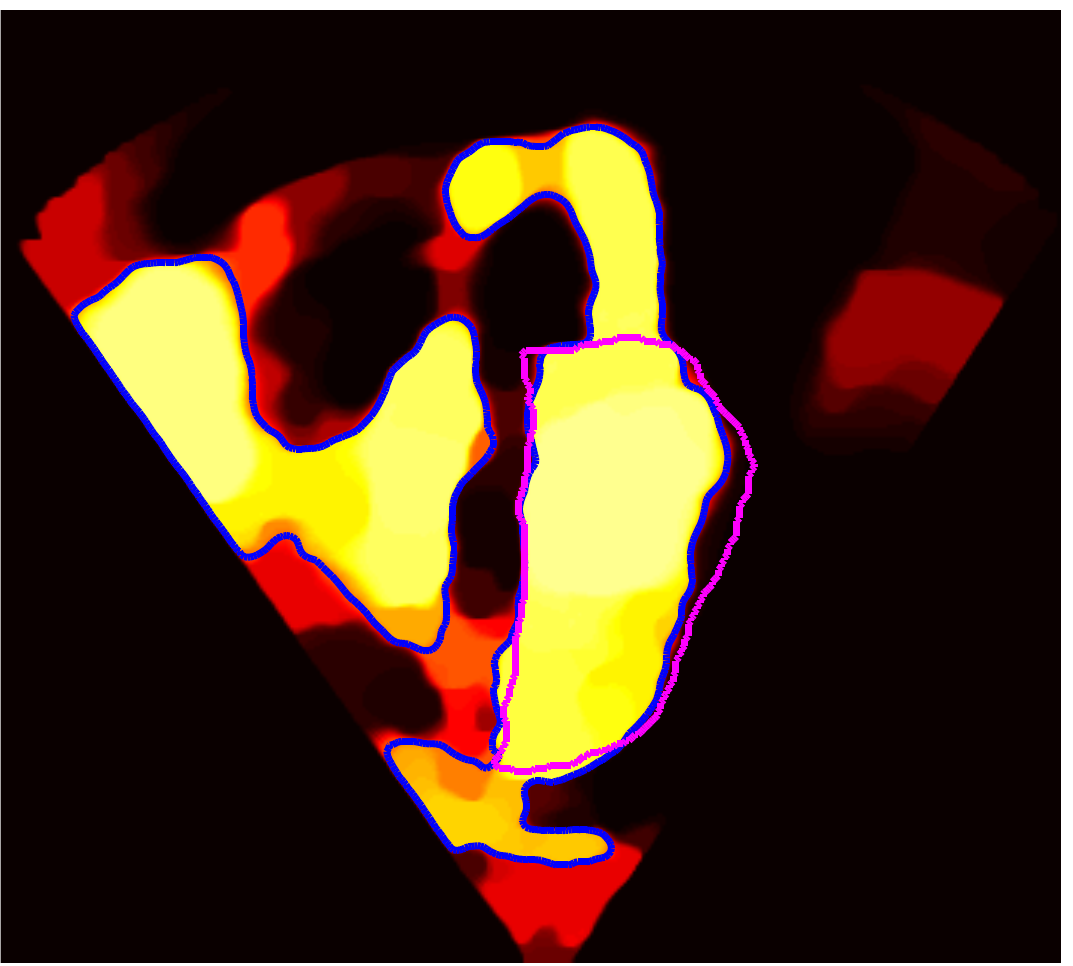} \\
    $\lambda=2$ & $\lambda=4$ & $\lambda=6$ & $\lambda=8$ & $\lambda=10$ \\
    \includegraphics[width=0.2\textwidth]{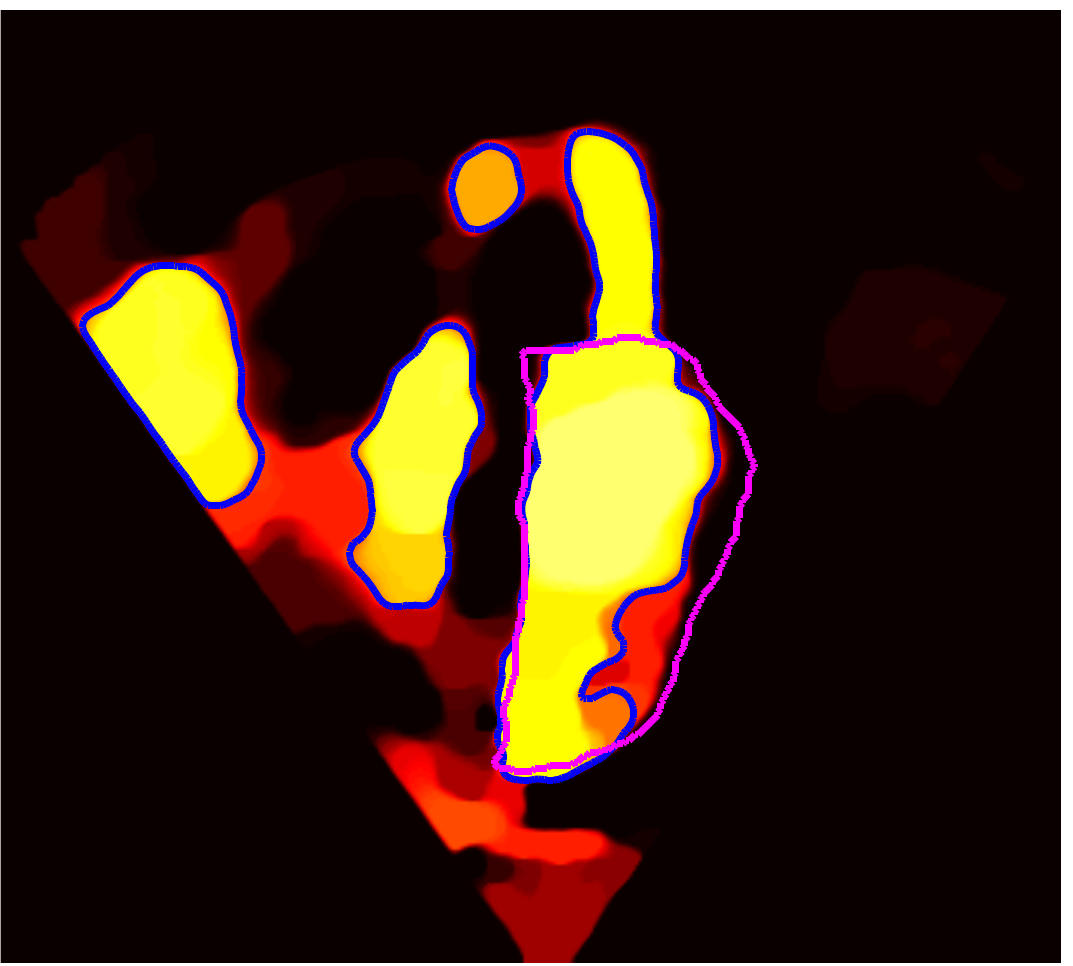} &
    \includegraphics[width=0.2\textwidth]{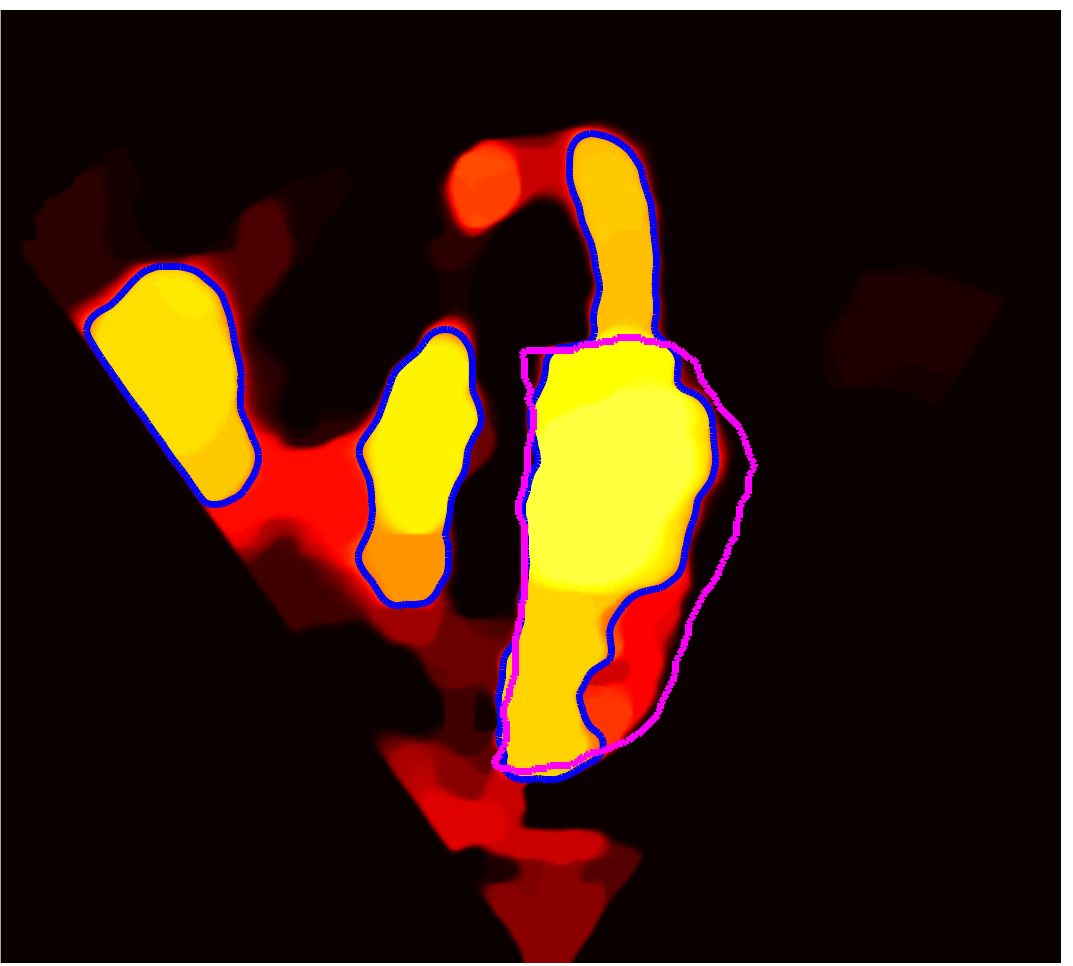} &
    \includegraphics[width=0.2\textwidth]{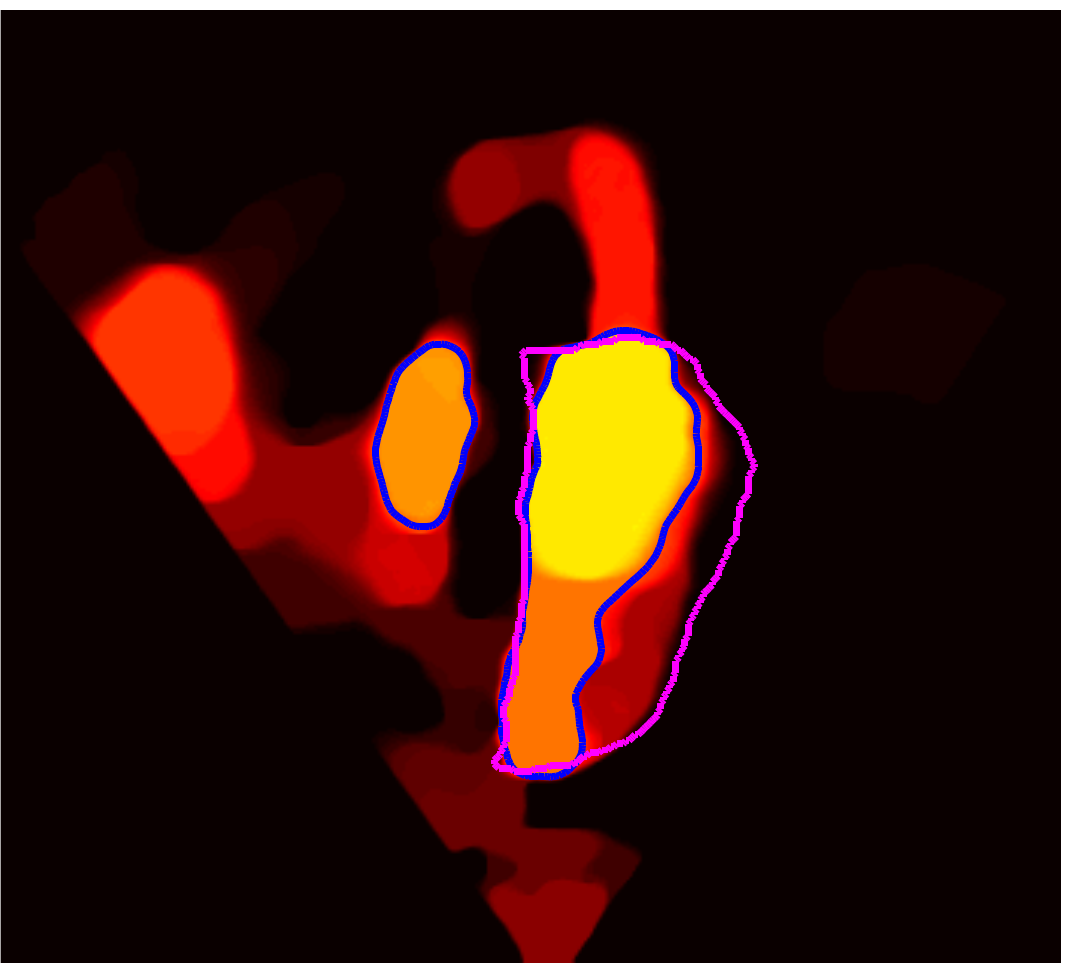} &
    \includegraphics[width=0.2\textwidth]{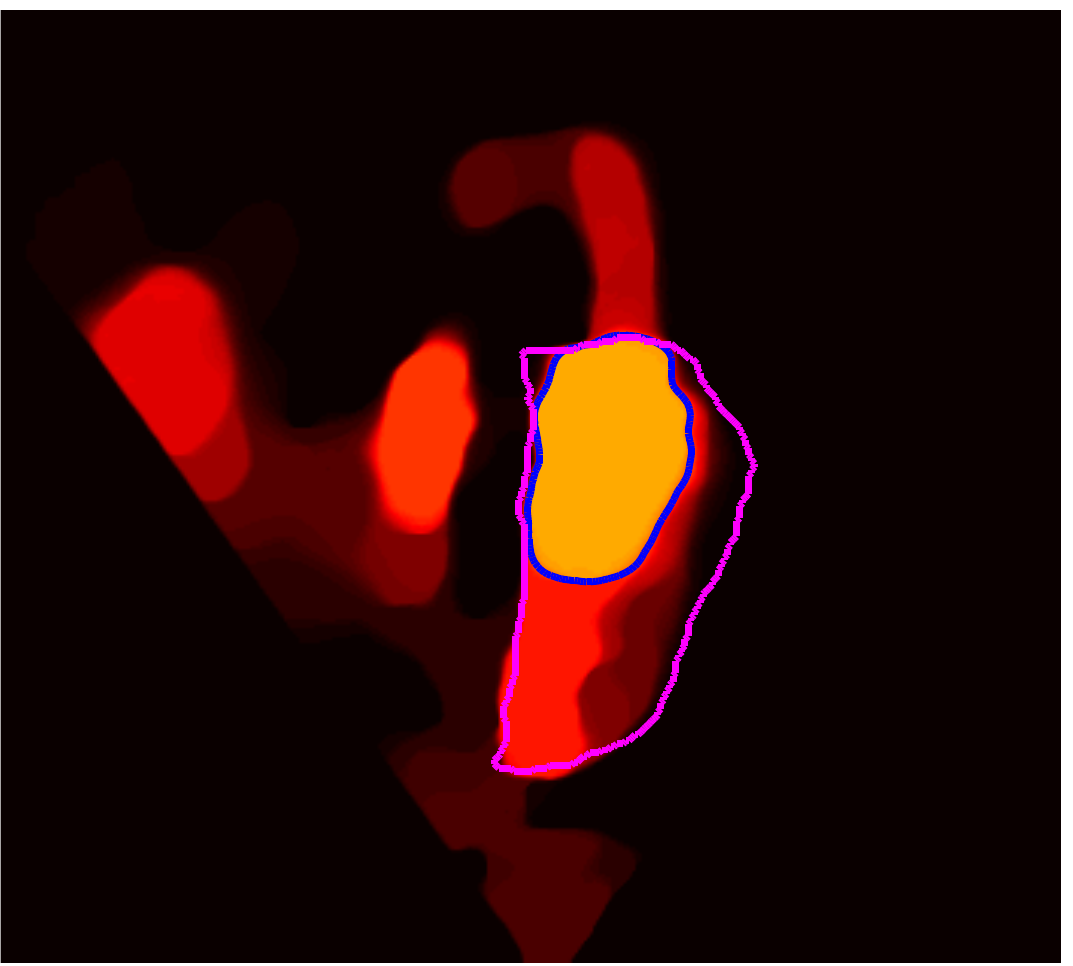} &
    \includegraphics[width=0.2\textwidth]{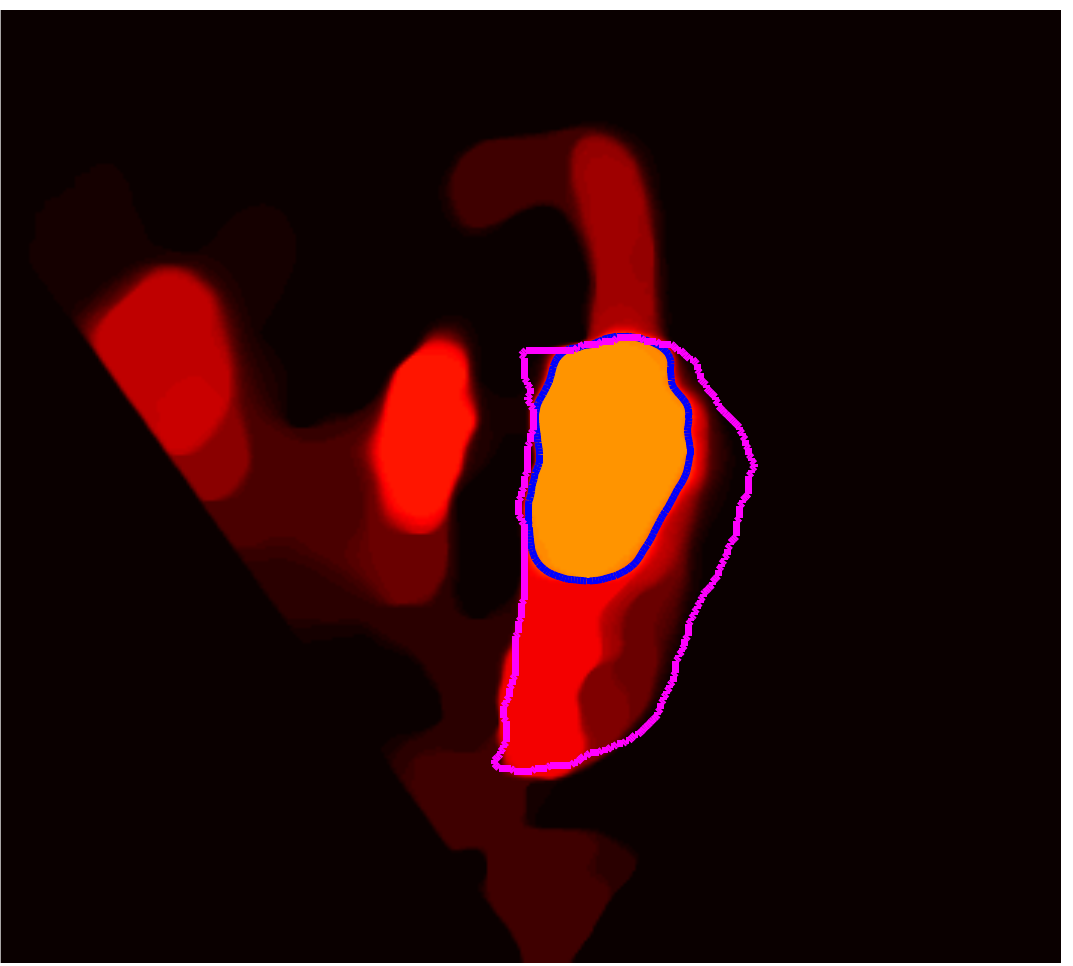} \\
    $\lambda=12$ & $\lambda=14$ & $\lambda=16$ & $\lambda=18$ & $\lambda=20$ \\
    \multicolumn{5}{c}{\includegraphics[width=0.4\textwidth]{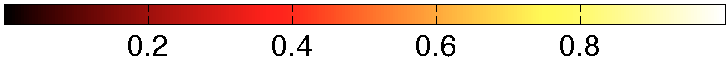}}
  \end{tabular}
  \caption{{Intensity-based segmentation results of the heart from an ultrasound image for the AMF model. Increased regularization captures increasingly consistent regions. Moderate to high regularization retains high probabilities of the blood pool while estimating low probabilities for the surroundings. Very large regularization yields ambiguous label probabilities throughout the complete image. Magenta contour indicates expert segmentation of the blood-pool, blue contour indicates the $0.5$ probability isocontour of the AMF solution.}}
  \label{FigAMFHeartSegmentationResults}
\end{figure}
\renewcommand{\tabcolsep}{6pt}

\Fig{FigHeartUS} shows an ultrasound image of the heart (left), an expert segmentation into blood pool, myocardium, and valves (middle) and the intensity distribution for the blood pool and outside the heart (right). These intensity distributions clearly overlap. We initialized the AMF model with this user-defined intensity distribution by sampling from the image followed by kernel-density estimation of the intensities. We re-estimated the intensity-distributions during the optimization. Specifically, given an intensity distribution, we compute the AMF solution, from that we obtain the binarized MAP solution that we use to re-estimate the intensity distributions using kernel-density estimation. We alternate AMF solution and density estimation to convergence. \Fig{FigAMFHeartSegmentationResults} shows the results of the AMF model for the estimation of label probabilities. The intensity ambiguity is captured in the  estimated label probabilities of the AMF model. Regularization behaves as expected: low regularization results in noisy label probability maps. Moderate to high regularization allows capturing of the blood pool (for the MAP solution) while declaring other regions ambiguous or low-probability. Very large regularization declares the full image ambiguous, as expected, because the model will, in this case, prefer overly large segmentation regions.

\begin{figure}
  \centering
  \begin{tabular}{ccc}
  \opt{arxiv}
  {
    \includegraphics[width=7.5cm,height=0.15\textheight]{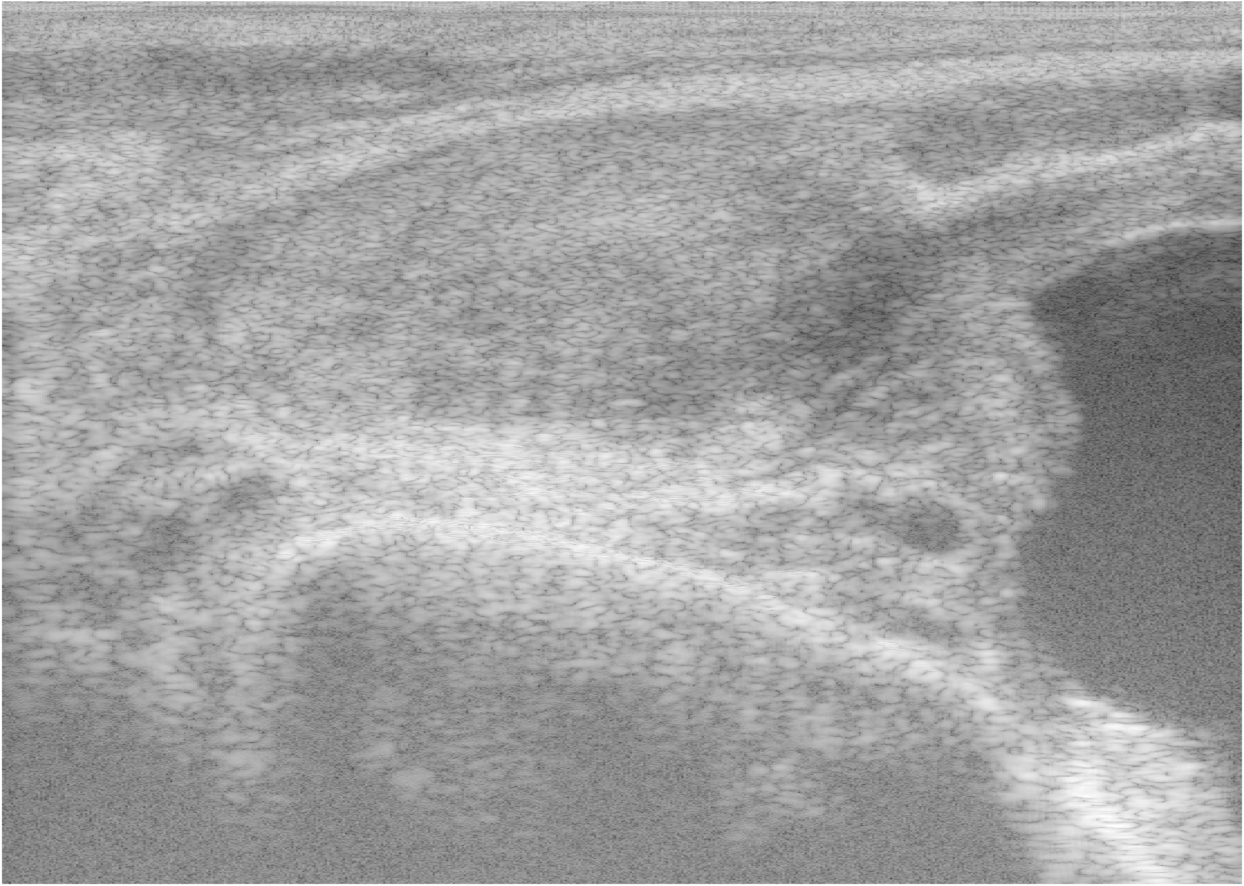} &
    \includegraphics[height=0.15\textheight]{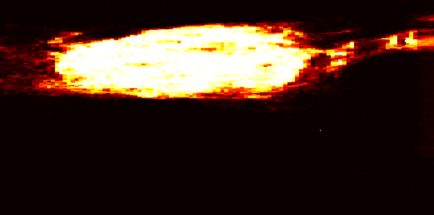} &
    \includegraphics[height=0.15\textheight]{colorbar}
    }
    \opt{siims}
    {
    \includegraphics[width=6cm,height=0.12\textheight]{prostate/US_fr45.png} &
    \includegraphics[height=0.12\textheight]{prostate/prostate_probability_map.png} &
    \includegraphics[height=0.12\textheight]{colorbar}
    }
  \end{tabular}
  \caption{{Left: ultrasound image of the prostate. Right: prostate probability map obtained by a machine-learning approach.}}
  \label{FigProstateUS}
%
  \centering
  \begin{tabular}{ccc}
\opt{arxiv}
{
    \includegraphics[width=0.225\textwidth]{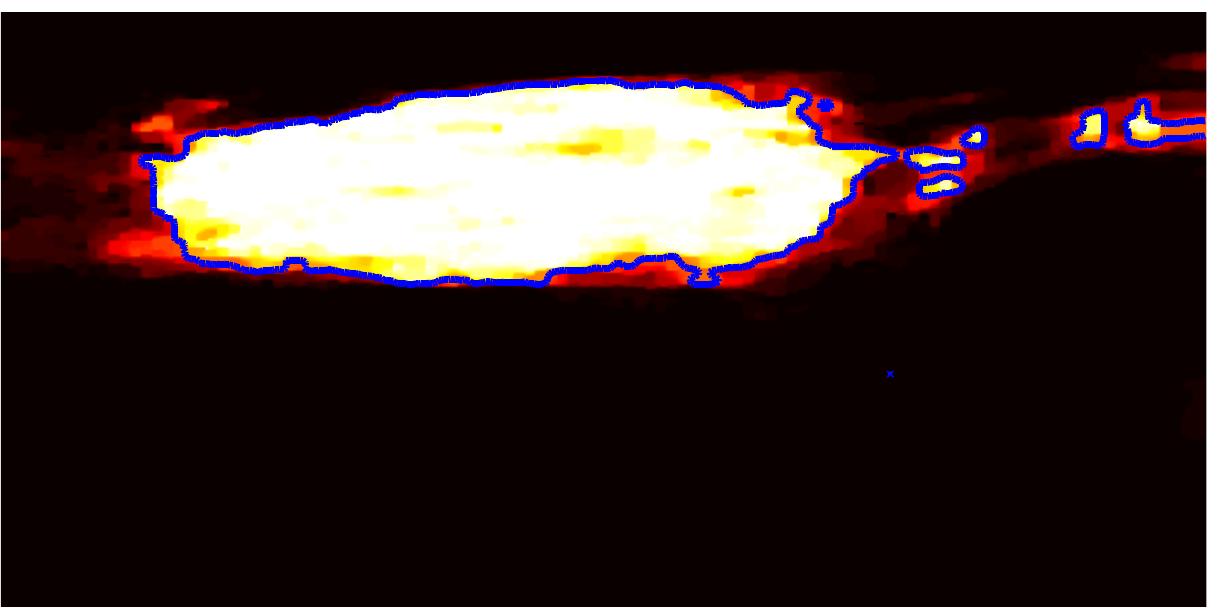} &
    \includegraphics[width=0.225\textwidth]{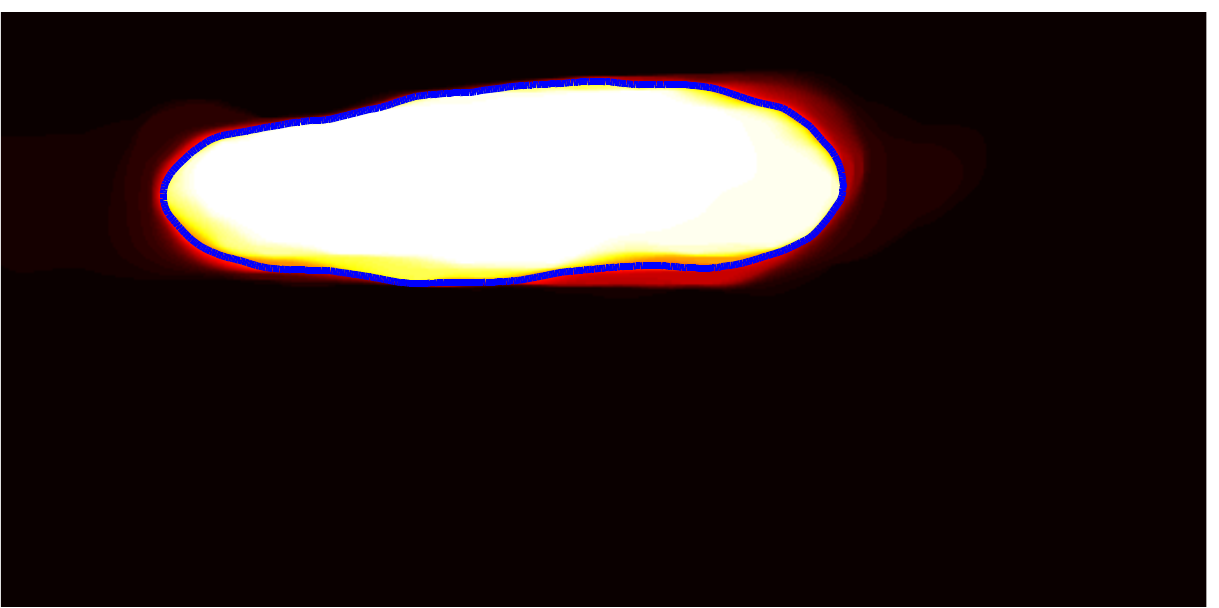} &
    \includegraphics[width=0.225\textwidth]{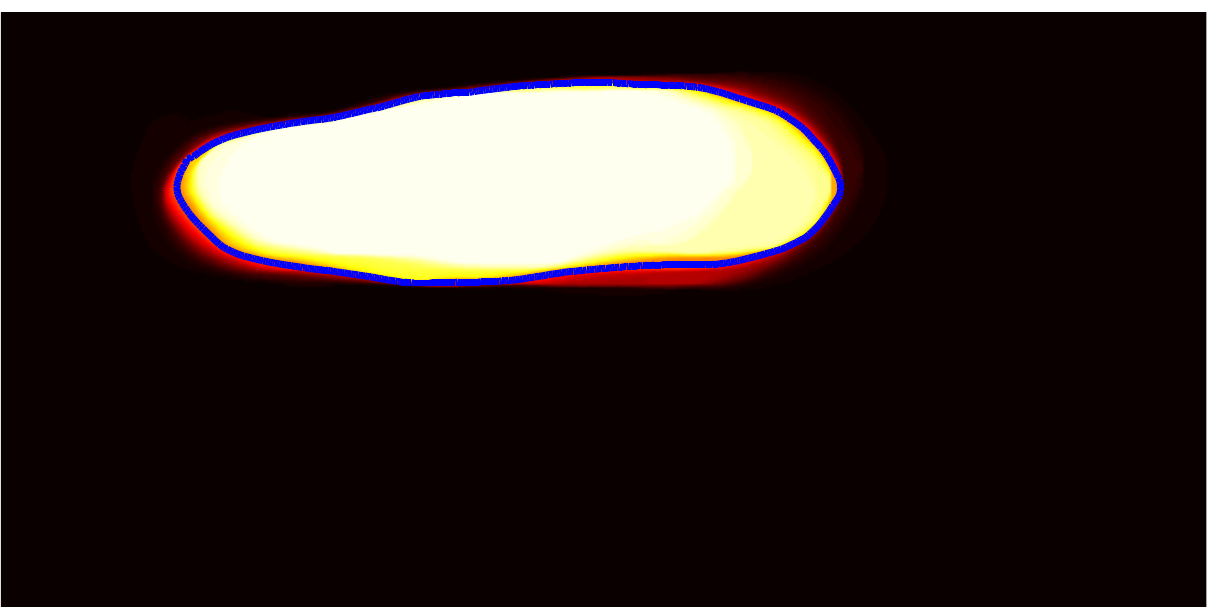} \\
    $\lambda=1$ & $\lambda=50$ & $\lambda=100$ \\
    \includegraphics[width=0.225\textwidth]{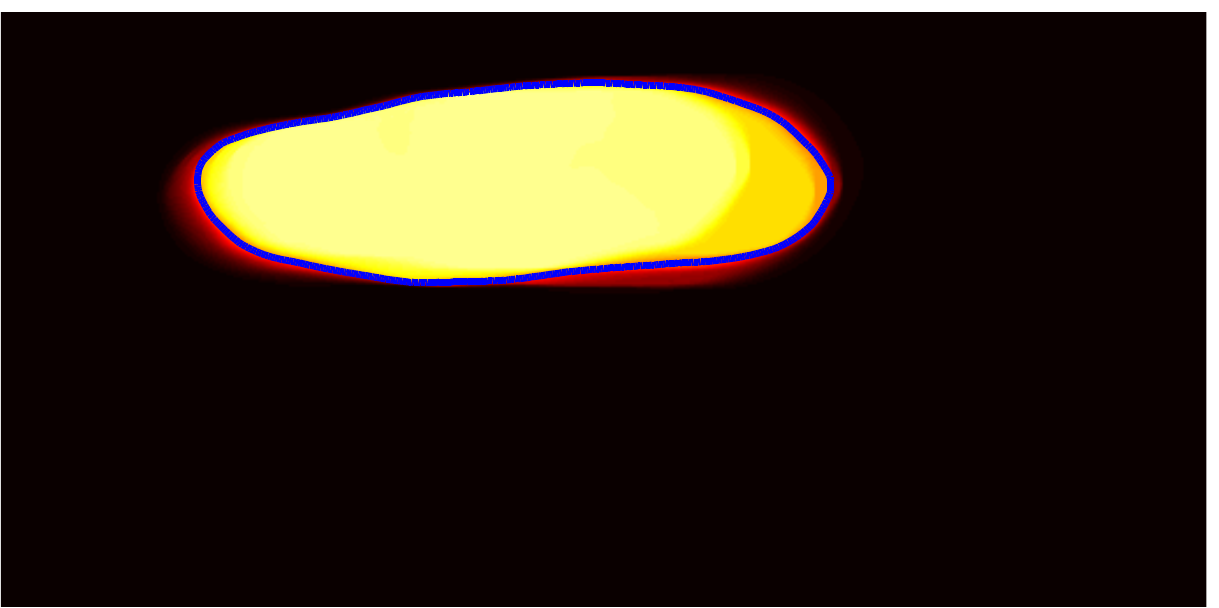} &
    \includegraphics[width=0.225\textwidth]{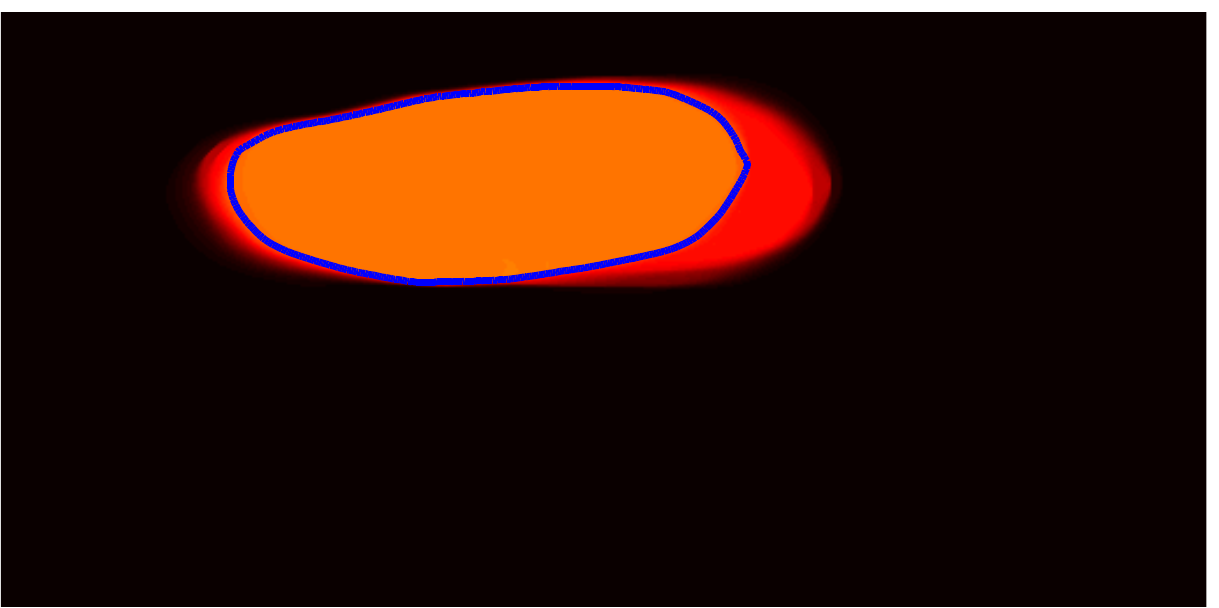} &
    \includegraphics[width=0.225\textwidth]{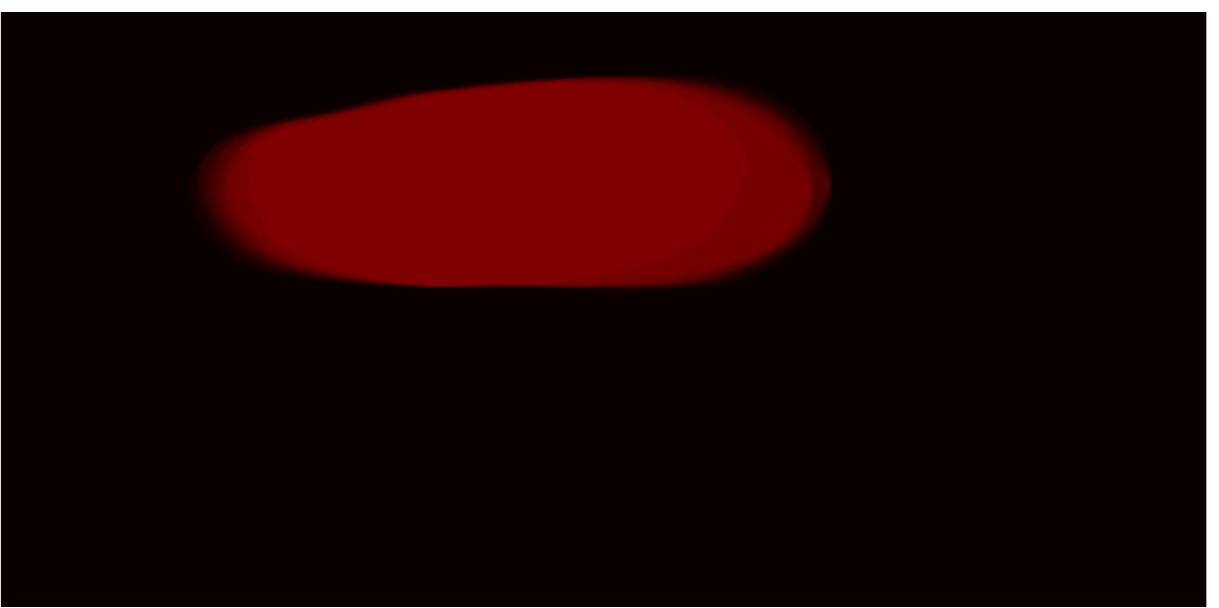} \\
    $\lambda=150$ & $\lambda=200$ & $\lambda=250$ \\
}
\opt{siims}
{
    \includegraphics[width=0.3\textwidth]{prostate/prostate_prob_map_gamma_auto_estimate_01} &
    \includegraphics[width=0.3\textwidth]{prostate/prostate_prob_map_gamma_auto_estimate_50} &
    \includegraphics[width=0.3\textwidth]{prostate/prostate_prob_map_gamma_auto_estimate_100} \\
    $\lambda=1$ & $\lambda=50$ & $\lambda=100$ \\
    \includegraphics[width=0.3\textwidth]{prostate/prostate_prob_map_gamma_auto_estimate_150} &
    \includegraphics[width=0.3\textwidth]{prostate/prostate_prob_map_gamma_auto_estimate_200} &
    \includegraphics[width=0.3\textwidth]{prostate/prostate_prob_map_gamma_auto_estimate_250} \\
    $\lambda=150$ & $\lambda=200$ & $\lambda=250$ \\
}
    \multicolumn{3}{c}{\includegraphics[width=0.4\textwidth]{colorbar_hot_horizontal}}
  \end{tabular}
  \caption{{Probability-map-based segmentation results of the prostate from an ultrasound image for the AMF model. \mn{Input to the AMF is the prostate probability map of \Fig{FigProstateUS}(right).} Increased regularization captures increasingly consistent regions. Moderate to high regularization retain high probabilities of the prostate while estimating low probabilities for the surroundings. Very large regularizations yield ambiguous label probabilities throughout the complete image. Blue contour indicates the $0.5$ probability isocontour of the AMF solution.}}
  \label{FigAMFProstateSegmentationResults}
\end{figure}

\mn{\Fig{FigProstateUS} shows an ultrasound image of the prostate (left) and the corresponding results of an experimental prostate segmentation system (right). The prototype system analyzed Radio Frequency (RF) ultrasound data using deep learning and random forest classification to generate label probabilities.  Alternating optimization, as in the heart example, was not used. \Fig{FigAMFProstateSegmentationResults} shows the results of the AMF model. The same conclusions as for the heart example apply. More regularity yields cleaner looking probability images as the AMF smooths the probability field as expected because of the connection to the ROF model. Changes are not as drastic as for the heart example as the initial probability map is already substantially more regular.}

\begin{figure}
  \centering
  \begin{tabular}{ccc}
    \includegraphics[width=0.3\textwidth]{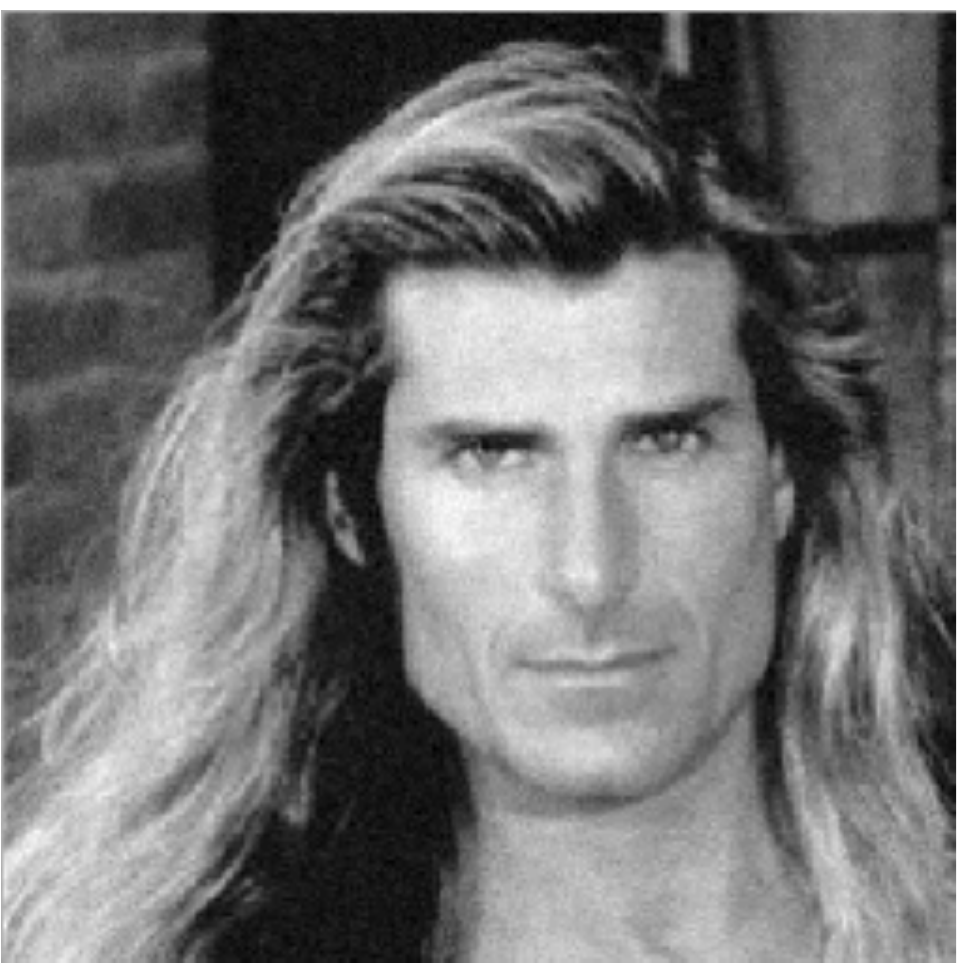} &
    \includegraphics[width=0.3\textwidth]{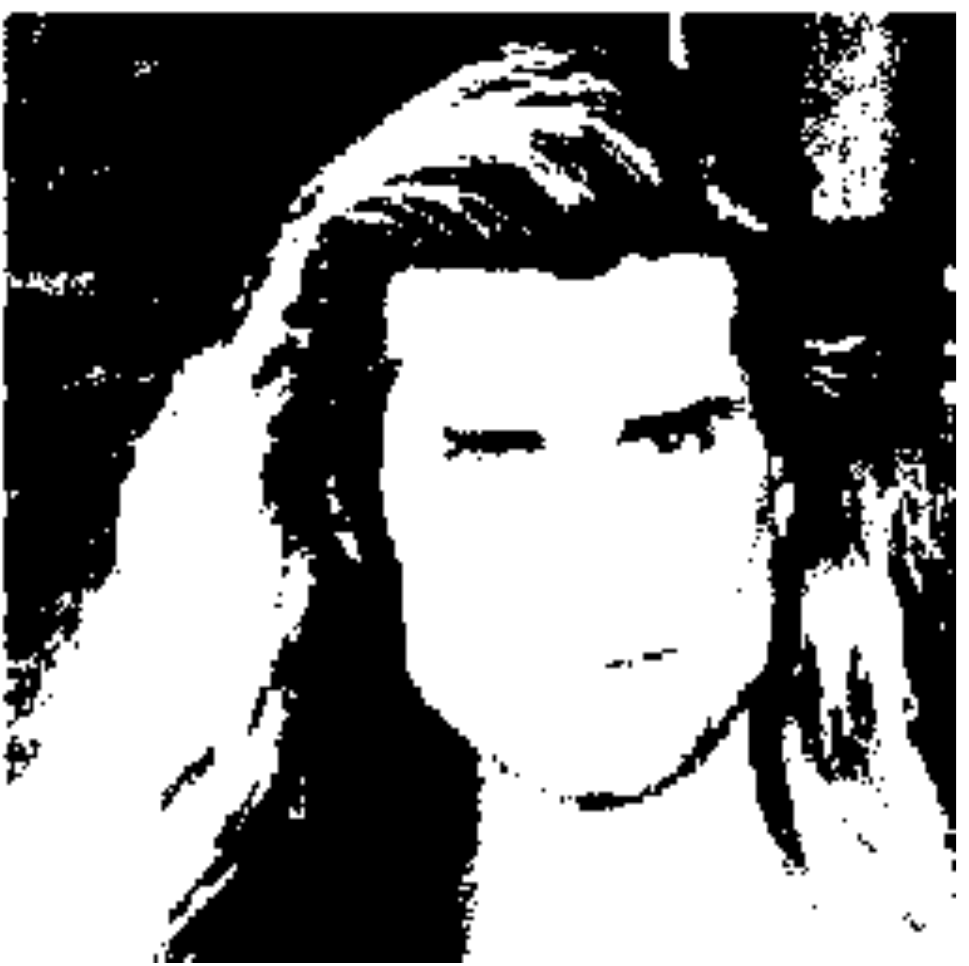} &
    \includegraphics[width=0.3\textwidth]{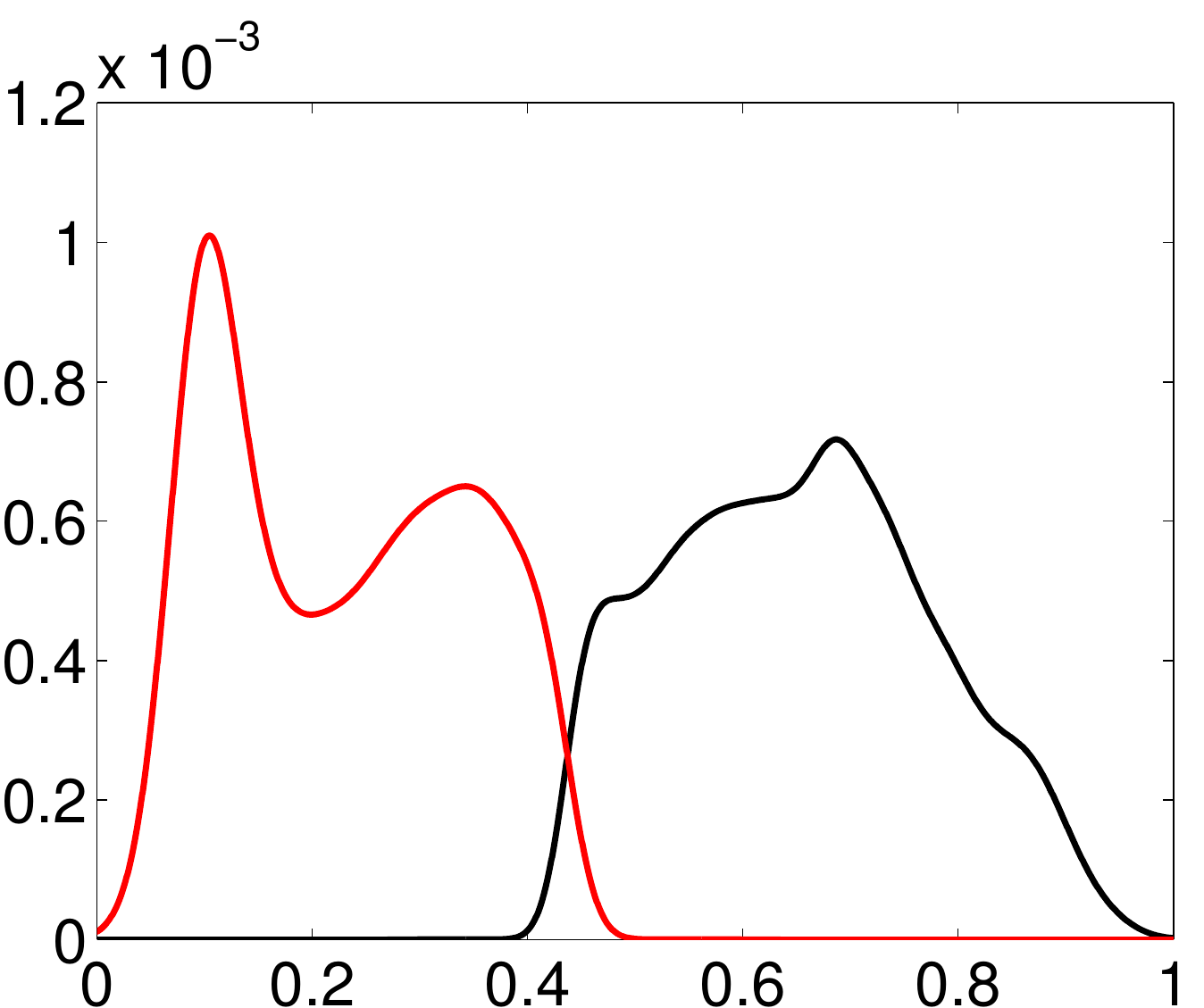}
  \end{tabular}
  \caption{{Left: Fabio image. Middle: Otsu-thresholded Fabio image. Right: intensity distributions for the intensity-normalized image ($I\in[0,1]$) based on the classes determined by Otsu thresholding.}}
  \label{FigFabioResults}
%
\renewcommand{\tabcolsep}{0pt}
  \centering
  \begin{tabular}{ccccc}
    \includegraphics[width=0.2\textwidth]{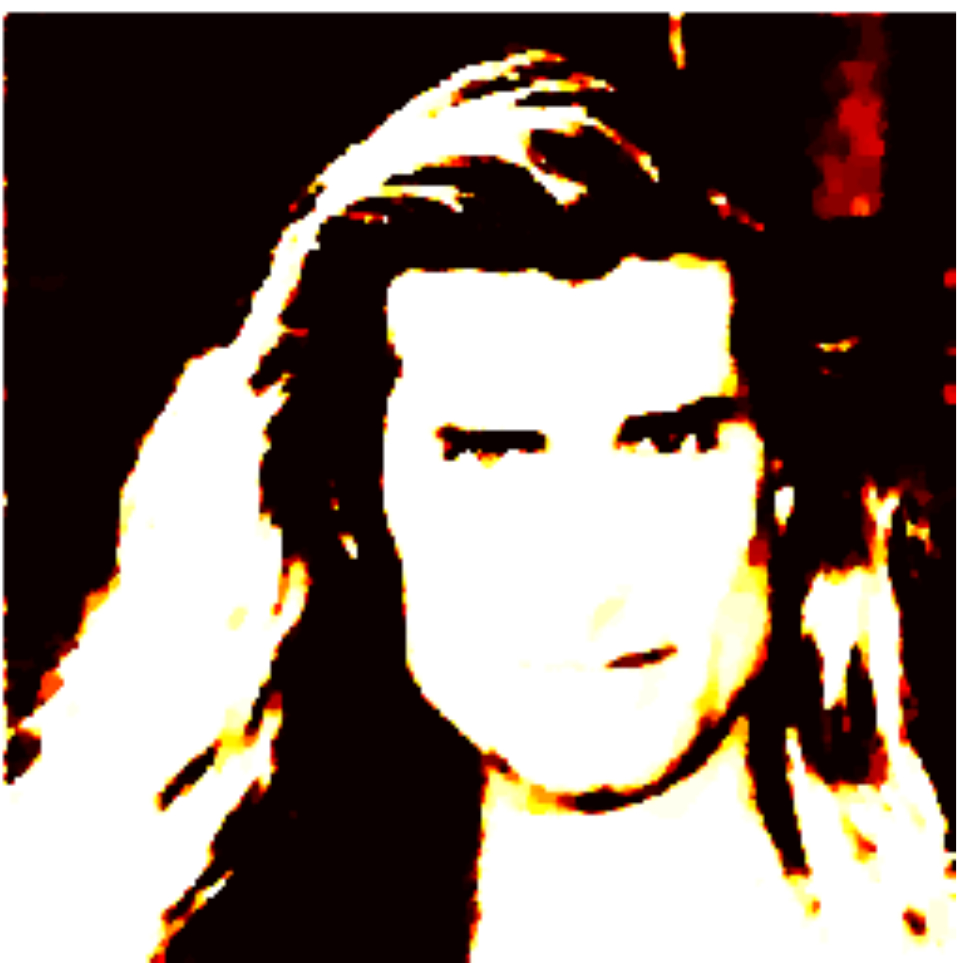} &
    \includegraphics[width=0.2\textwidth]{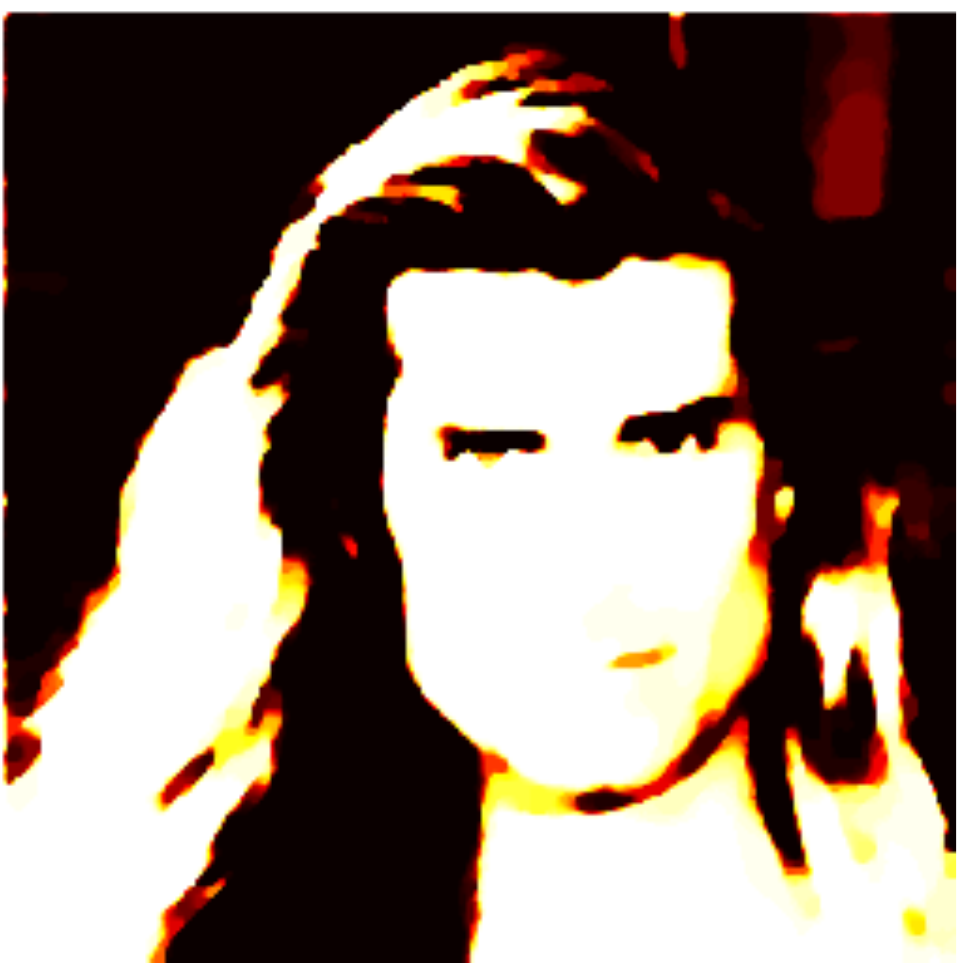} &
    \includegraphics[width=0.2\textwidth]{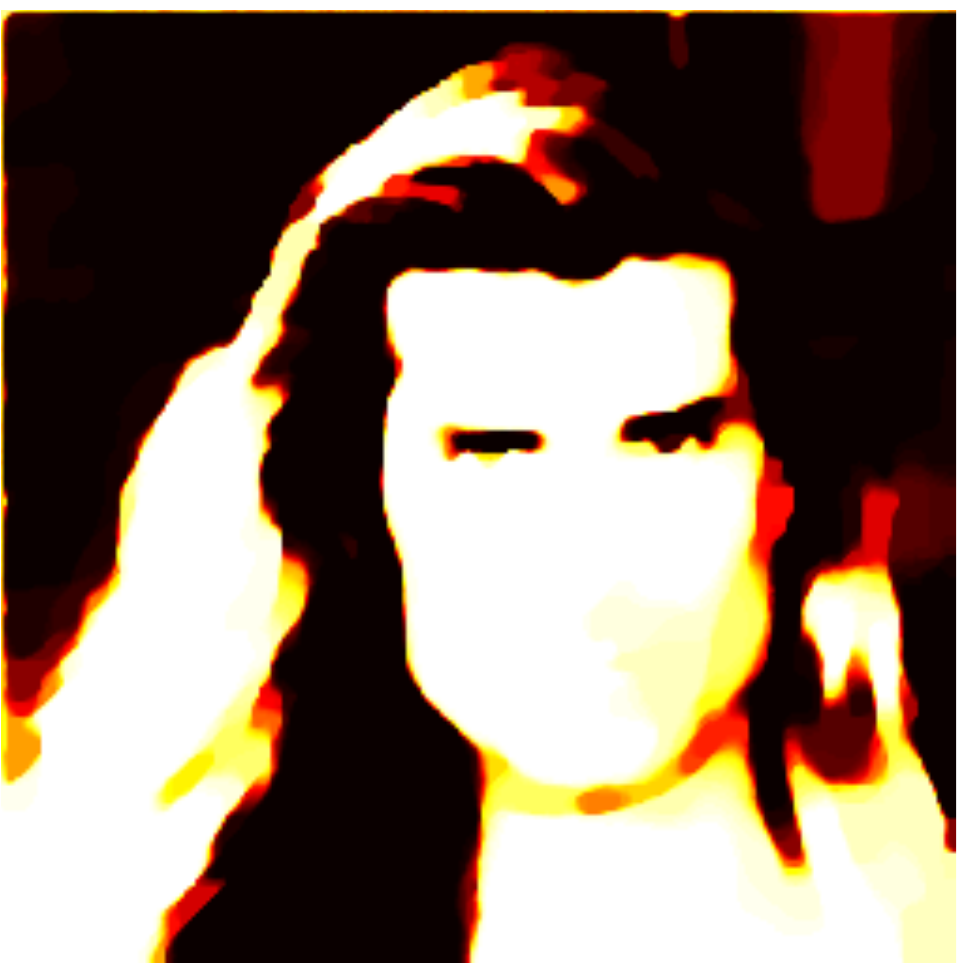} &
    \includegraphics[width=0.2\textwidth]{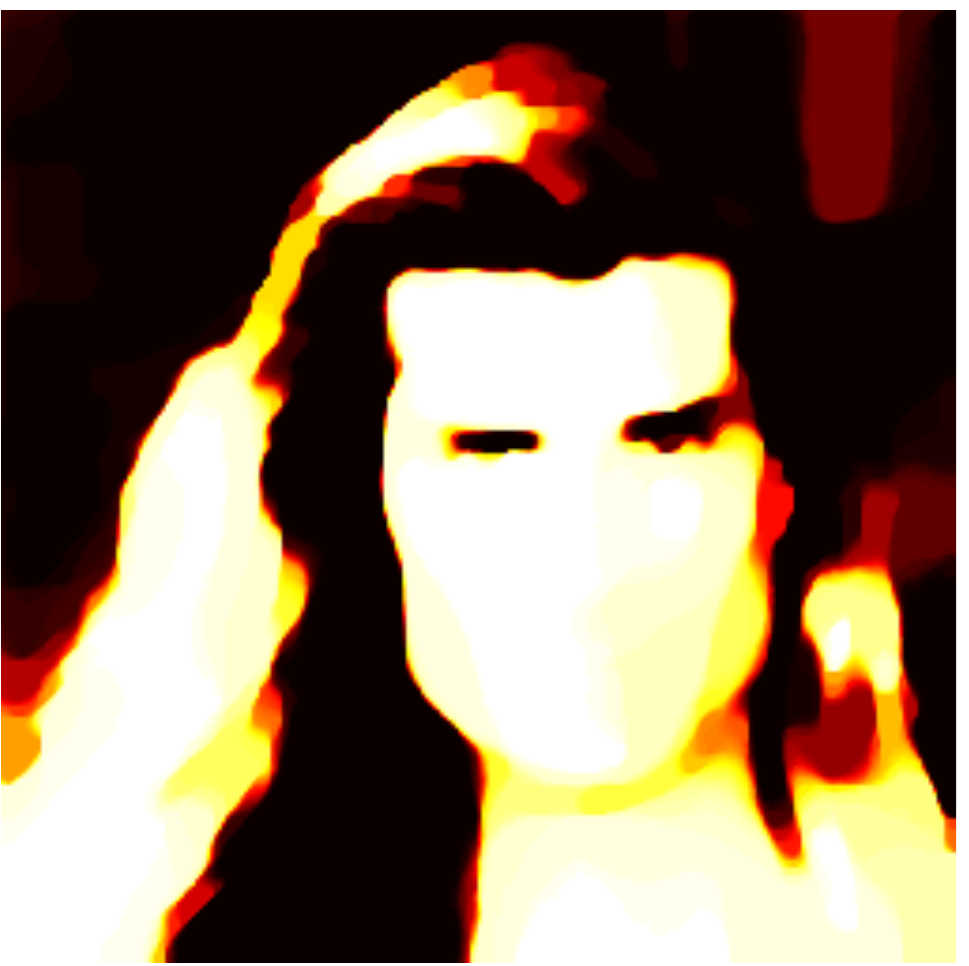} &
    \includegraphics[width=0.2\textwidth]{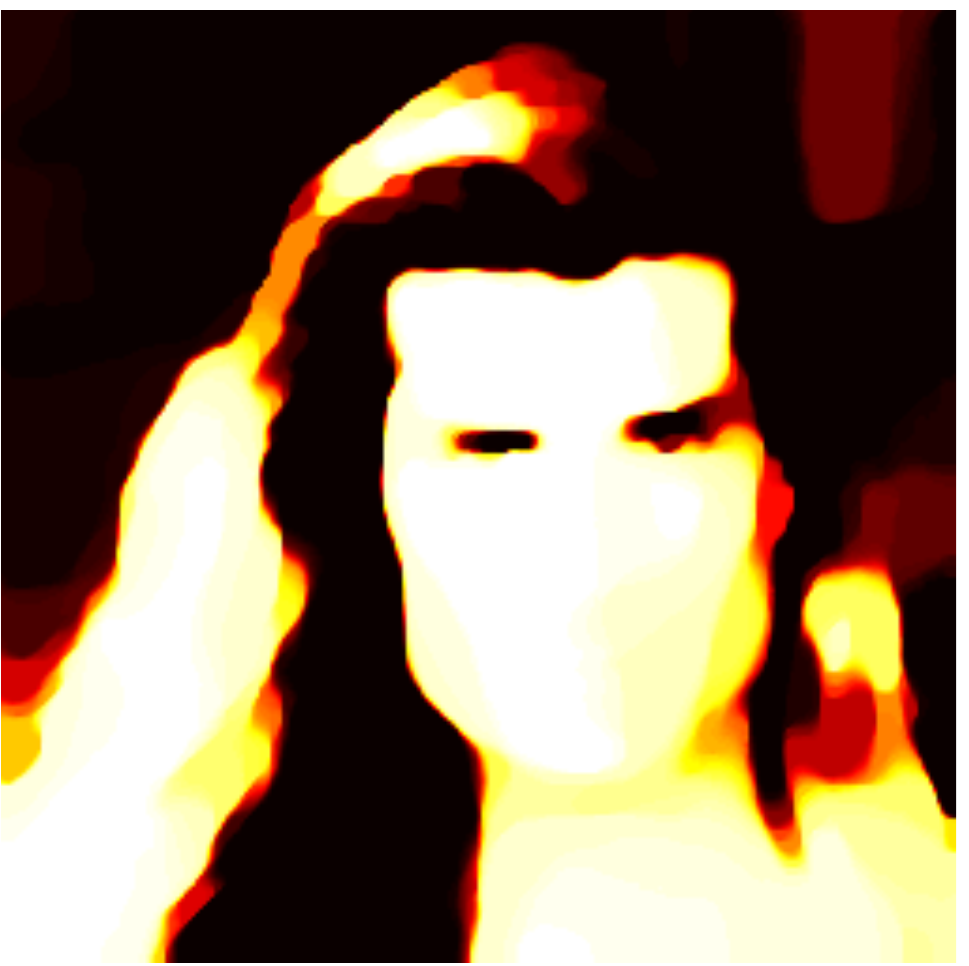} \\
    $\lambda=2$ & $\lambda=4$ & $\lambda=6$ & $\lambda=8$ & $\lambda=10$ \\
    \includegraphics[width=0.2\textwidth]{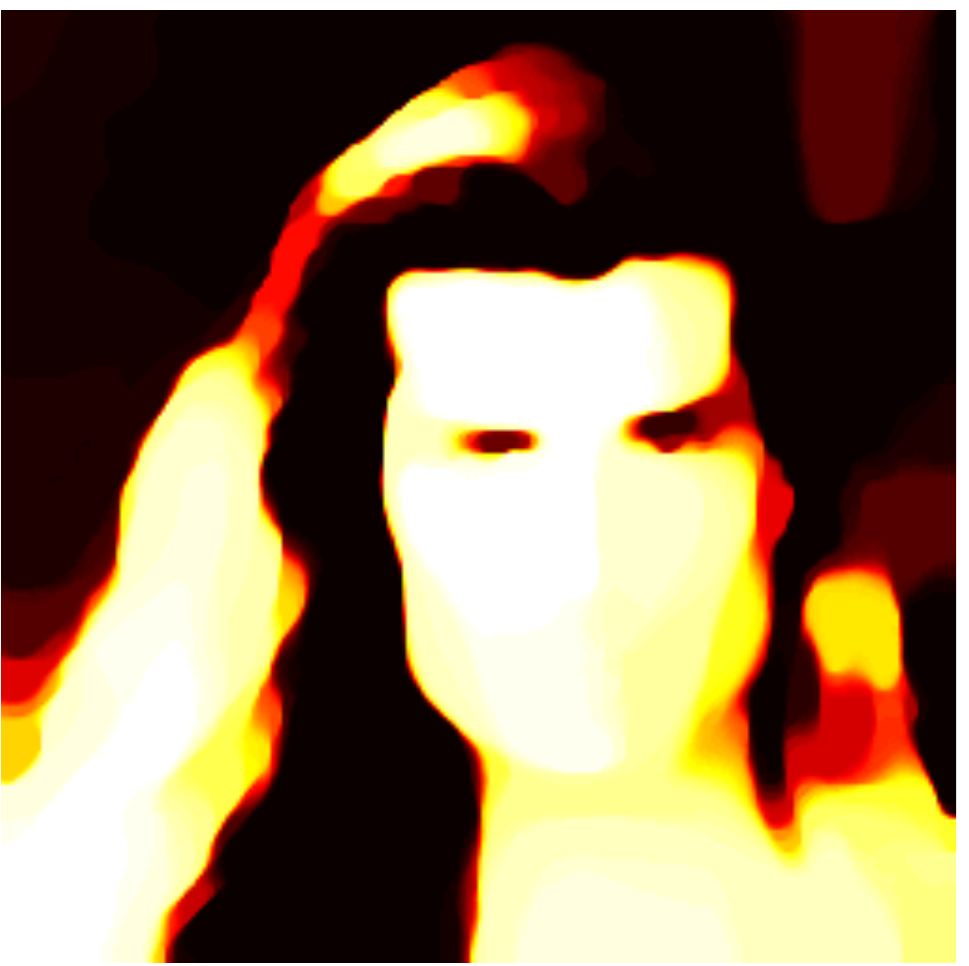} &
    \includegraphics[width=0.2\textwidth]{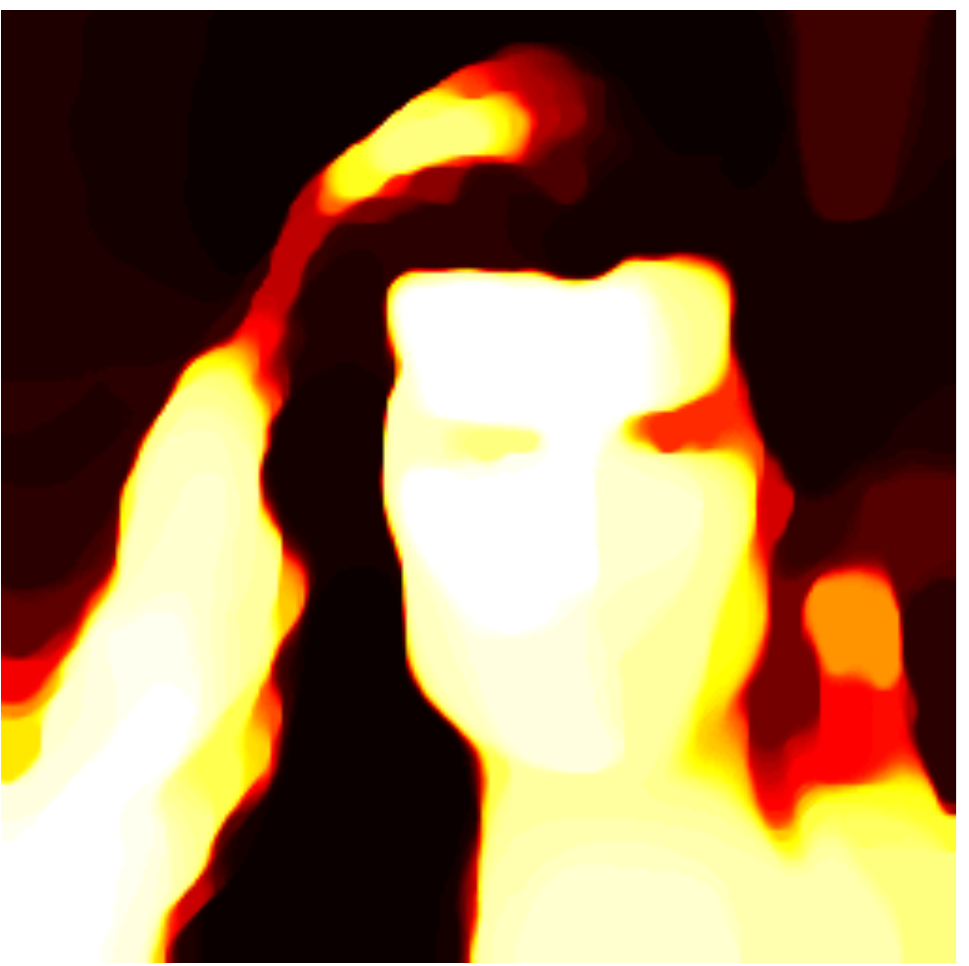} &
    \includegraphics[width=0.2\textwidth]{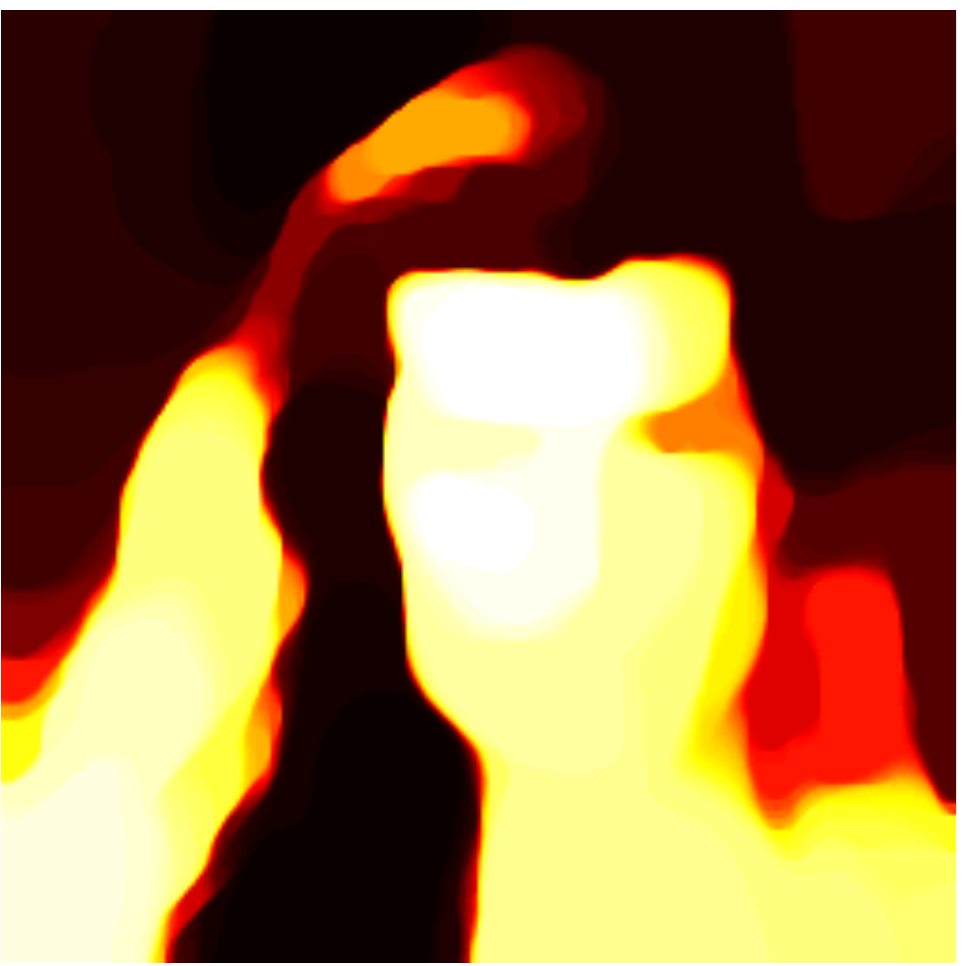} &
    \includegraphics[width=0.2\textwidth]{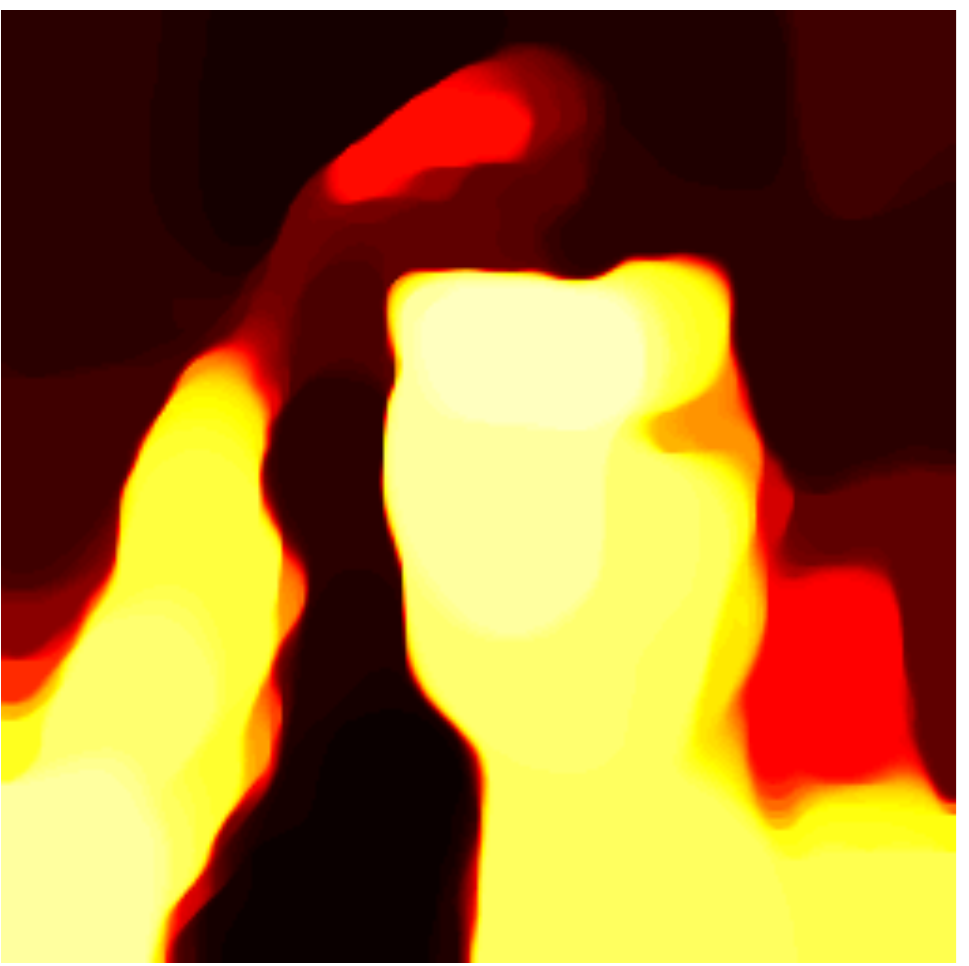} &
    \includegraphics[width=0.2\textwidth]{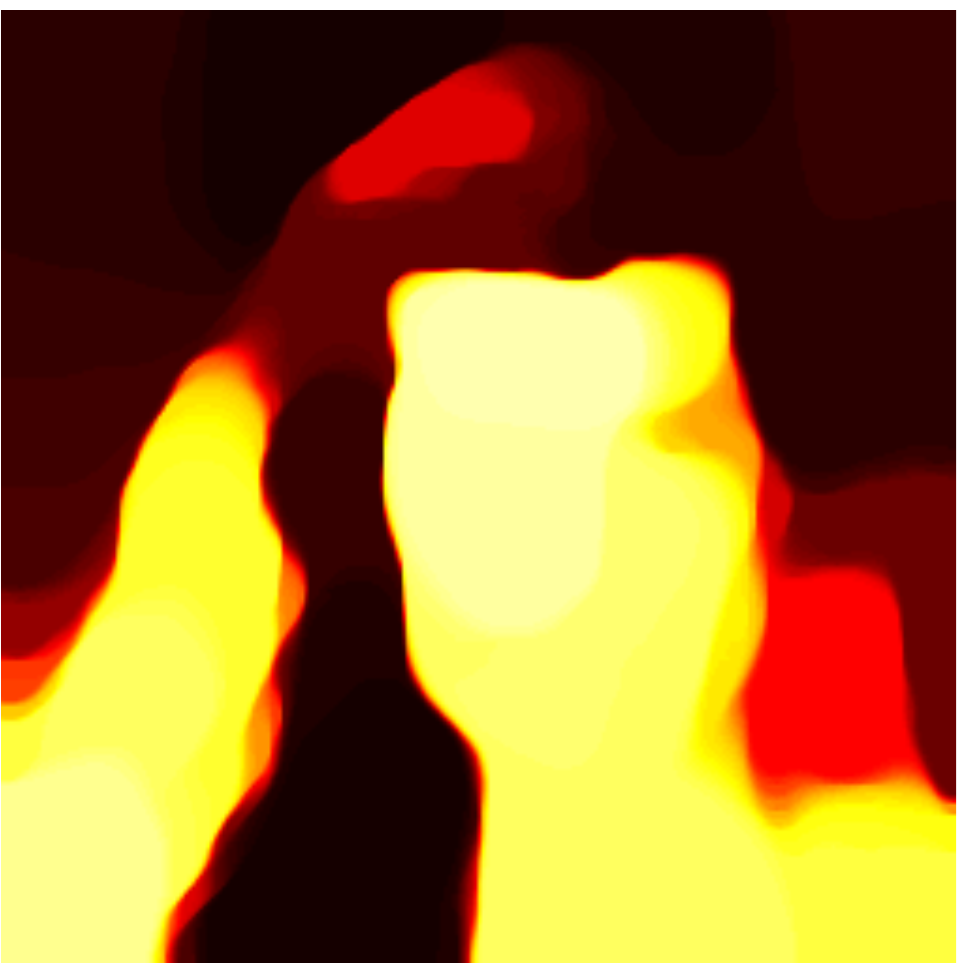} \\
    $\lambda=12$ & $\lambda=14$ & $\lambda=16$ & $\lambda=18$ & $\lambda=20$ \\
    \multicolumn{5}{c}{\includegraphics[width=0.4\textwidth]{colorbar_hot_horizontal}}
  \end{tabular}
  \caption{{Intensity-based segmentation results for the Fabio image for the AMF model.}
}
  \label{FigAMFFabio_SegmentationResults}
\end{figure}

\mn{\Fig{FigFabioResults} show the original Fabio image including its segmentations based on a modified version of Otsu thresholding (where foreground and background classes can have distinct means {\it and} standard deviations) and the corresponding intensity histogram. This image can be separated reasonably well using intensity information alone. \Fig{FigAMFFabio_SegmentationResults} shows the corresponding AMF results. We obtained these results by initializing AMF using the modified Otsu-thresholding procedure and then followed the same alternating optimization approach as for the heart ultrasound segmentation. Clearly, larger values for the regularization parameter $\lambda$ put the emphasis on larger image structures.}

\mn{These experiments show that the AMF model (i) results in label probabilities which are spatially smooth (as expected due to the connection to the ROF model), (ii) exhibits a balancing effect between local label likelihood and spatial regularization, and (iii)  tends to more uncertain label assignments for strong spatial regularization.}

\renewcommand{\tabcolsep}{6pt}

\subsubsection{Quantitative Assessment of AMF on Real Data}
\label{SecQuantitativeAMF}

\mn{For quantitative analysis, we applied AMF to  the segmentation benchmark data (\texttt{icgbench}) of Santner et al.~\cite{santner2010}. This benchmark dataset consists of 158 natural color images (391 by 625 pixels). For each image a manual segmentation is available. Furthermore, each image contains seed regions for the objects to be segmented. In total there are 262 seed regions and 887 objects.} \mn{As proposed by Santner et al.~\cite{santner2010}, we train a random forest \mn{(using Matlab's \texttt{TreeBagger} function)} for each image given pixel color information in image areas defined by user-provided seed locations dilated by a disk structural element of radius of 9 pixels. Each random forest consists of 100 trees, $\lambda$ was set to $10$ for all the experiments, and was trained on local CIElab color features. Once trained on the seeds, the resulting random forest classifier is applied to the full images generating noisy label probabilities. The mean computation time for an AMF segmentation was 22.1s for the RGB color images of the \texttt{icgbench} dataset using a \mn{Matlab CPU implementation on a 2GHz Intel Xeon, E5405. The computer had 8 cores, but the code was not explicitly multi-threaded (beyond what Matlab multi-threads automatically).} As \texttt{icgbench} is a dataset for multi-label segmentation but our current AMF model only supports binary segmentation tasks\footnote{\mn{A multi-label extension is likely possible, but it remains to be investigated if connections to the ROF and the CV models can still be made.}}, we investigate two different segmentation approaches:
\begin{itemize}
        \item {\it Individual Binary Segmentations:} For a given image we create binary segmentations by considering one class as the foreground and all other classes as the background.
        \item {\it Quasi-Multi-Label Segmentation:} Individual binary segmentations do not guarantee that local label probabilities over all classes sum up to one. Hence, we project the local label probabilities obtained from the individual binary segmentations onto the probability simplex. We used the standard Euclidean projection~\cite{chen2011,michelot1986} onto the simplex though other approaches could be used as well~\cite{pohl2007using,andrews2014}.
\end{itemize}
}


\begin{figure}
        \centering
        \begin{tabular}{cc}
               \includegraphics[height=0.3\textheight]{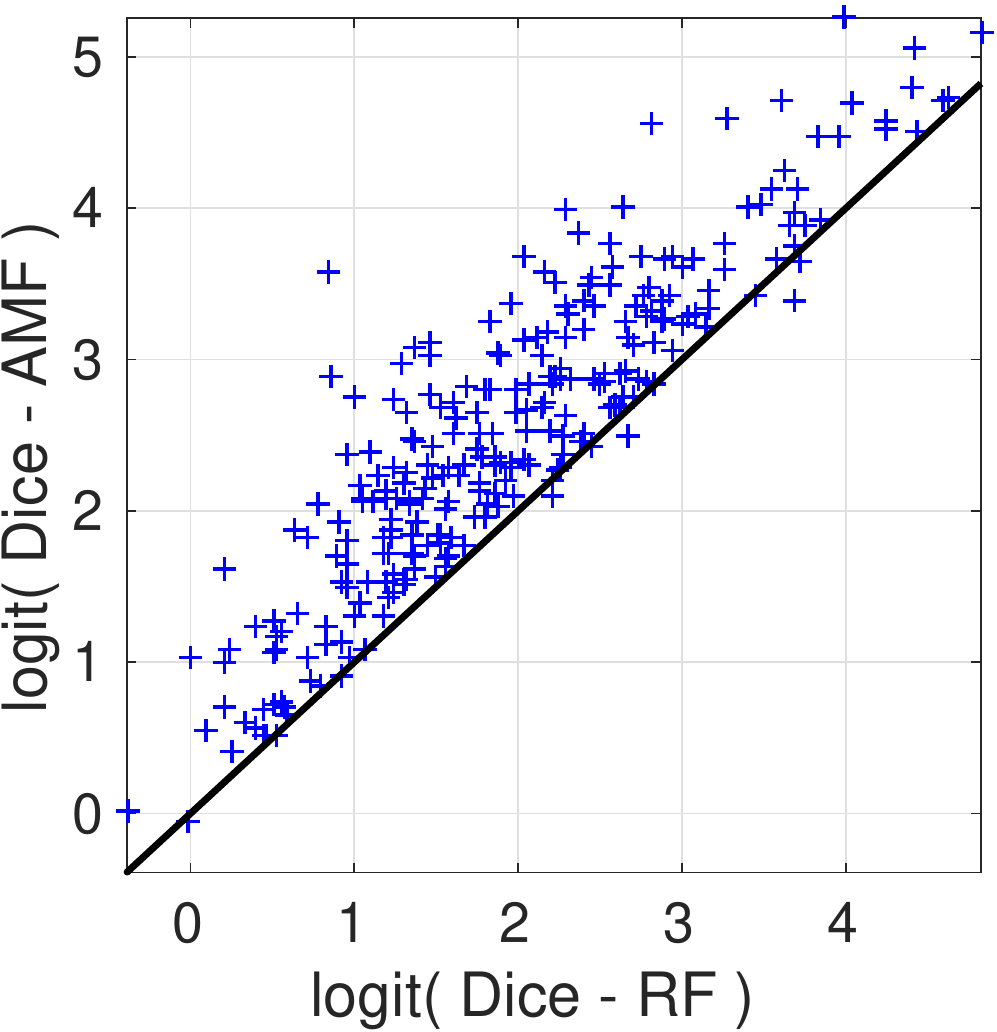} &
               \includegraphics[height=0.3\textheight]{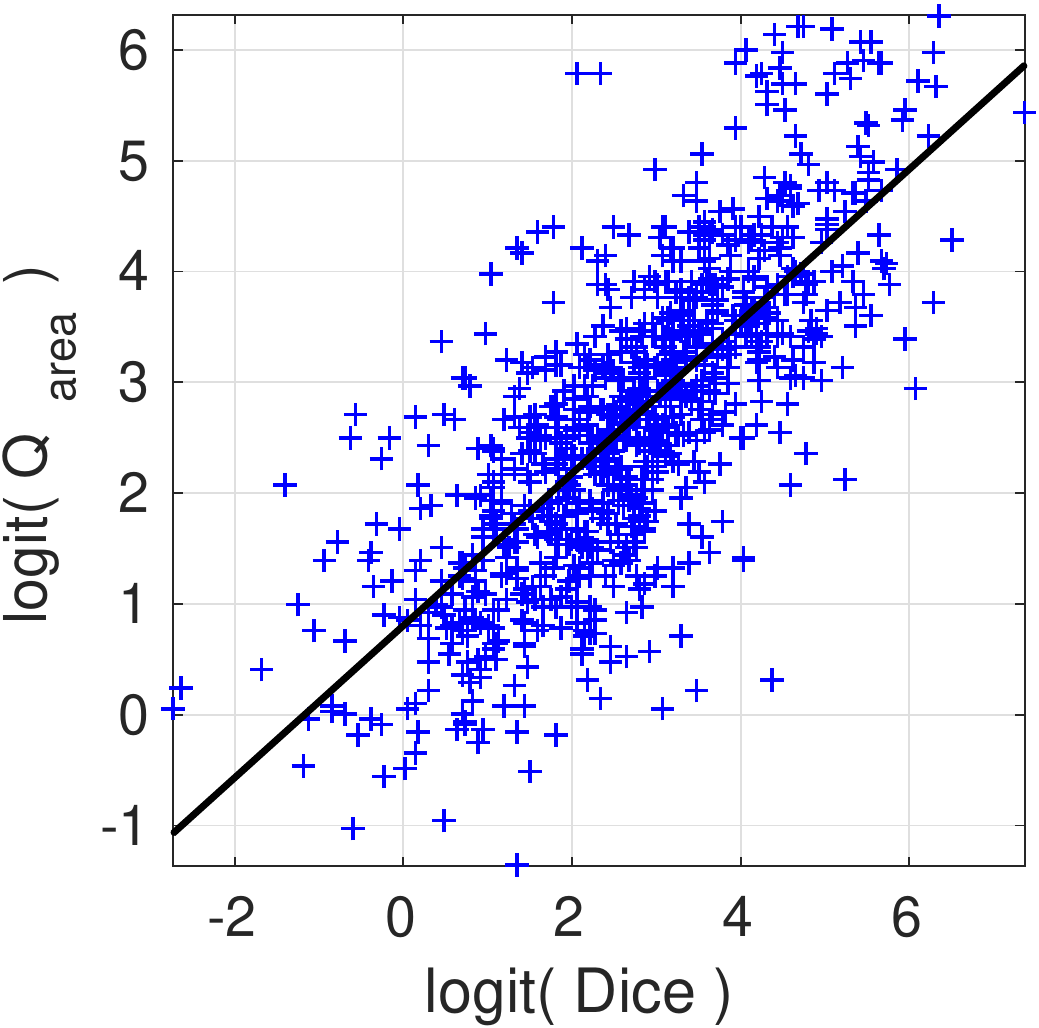}
        \end{tabular}
        \caption{\mn{Left: Scatter plot for Dice segmentation scores for all the objects of the \texttt{icgbench} database. Comparison between obtained Dice scores through the random forest (RF) and after applying the AMF model (AMF). Values are logit transformed before plotting for better visualization: $\text{logit}(p) = \ln (p/(1-p))$. In the vast majority of the cases, AMF improves the Dice score. Line indicates equal values for RF and AMF model, \ie, values above the line indicate a better performance of the AMF model compared to the RF. Right: Scatter plot between Dice segmentation score and area-normalized posterior approximation $Q$ of the AMF. Values are also logit transformed for better visualization. High Dice scores are generally related to high $Q$ values. A clear linear trend is visible for the logit-transformed variables. Line is a least-squares fit to the logit transformed $Q/Dice$ value pairs. Sample $(p,\text{logit}(p))$ pairs are as follows: $(0.01,-4.60)$, $(0.25,-1.10)$, $(0.5,0)$, $(0.75,1.10)$, $(0.9,2.20)$, $(0.99,4.60)$.}}
        \label{fig:dice_and_normalizedQ}
\end{figure}

\mn{\Fig{fig:dice_and_normalizedQ}(left) compares the obtained Dice scores over all 887 individual object segmentations based on the random forest and based on AMF applied to the random forest label probabilities. The Dice score between two sets $S_1$ and $S_2$ is defined as
\begin{equation}
        \text{Dice}(S_1,S_2) \triangleq \frac{2|S_1\cap S_2|}{|S_1| + |S_2|}.
\end{equation}
To evaluate image segmentations, \mn{$S_1$ and $S_2$ correspond to sets of object pixels which are the most likely for a given object class label (\ie, foreground and background).}
AMF clearly improves the segmentations generated by the random forest. The mean Dice score (with standard deviation in parentheses) for the individual segmentations over all images is 0.82(0.18) for the random forest, which are significantly worse \mn{($p<10^{-10}$)} according to a one-sided paired t-test)
than individual binary AMF segmentations, whose mean Dice score is 0.88(0.15). The quasi-multi label AMF approach further improves the mean Dice score to 0.89(0.14). Computing multi-label Dice-scores\footnote{\mn{We compute the multi-label Dice score as the mean over the {\it individual} Dice scores for the individual binary segmentations for an image. Hence, we obtain {\it one} multi-label Dice score per {\it image}, but as many individual Dice scores as there are objects in an image.}} for all the images results in a mean Dice score of 0.84(0.11) for the random forest and 0.90(0.09) for the quasi-multi-label AMF segmentation, which is significantly better ($p<1e-10$ according to a one-sided paired t-test) and matches the Dice score obtained by Santner et al.~\cite{santner2010} when using the same features.}

\mn{Not only is our method simpler than the approach by Santner et al.~\cite{santner2010}, which uses a sophisticated random forest implementation coupled with a true multi-label segmentation approach (\ie, all labels are jointly considered during the segmentation and not in a one-versus-all-other classes fashion as in our approach), but our method also complements the MAP solution with posterior label probabilities, which can be used to quantitatively assess the confidence in the segmentation. A possible confidence measure is to compute an {\it area-normalized} approximation of the posterior 
\mn{\begin{eqnarray}
        Q(\zvar;~{\theta}) & \triangleq & \prod_\indx {\theta}_i^{\zvar_i} \left(1 - {\theta}_i\right)^{1 - \zvar_i} \\
        &=& \exp\left\{ \sum_{i\in\{i:\zvar_i=1\}}  \ln(\theta_i) + \sum_{i\in\{i:\zvar_i=0\}}\ln(1-\theta_i)\right\}.
\end{eqnarray}}

\mn{Area-normalization is useful as object sizes in the \texttt{icgbench} dataset vary greatly. Specifically, we define the area-normalized form of $Q(\cdot;\cdot)$ as}
\begin{equation}
        Q_{\text{area}}(\zvar;~{\theta}) \triangleq \exp\left\{ \frac{1}{|\{i:\zvar_i=1\}|}\sum_{i\in\{i:\zvar_i=1\}}  \ln(\theta_i) + \frac{1}{|\{i:\zvar_i=0\}|}\sum_{i\in\{i:\zvar_i=0\}}\ln(1-\theta_i)\right\}.
\end{equation}
}
\mn{We use the MAP solution of the AMF, $\zvar_{map}$, to evaluate $Q_{\text{area}}$.}
\mn{For a binary $\theta$, \ie, no uncertainty in the inferred binary segmentation,
$Q_{\text{area}}(\zvar_{map};~{\theta}) = 1$. The value of $Q_{\text{area}}(\zvar_{map};~{\theta})$ decreases if $\theta$ is not binary indicating uncertainty in the inferred segmentation. In comparison to the approximate posterior $Q$, this area-normalized measure allows us to assess uncertainty of objects independent of their size. For $Q_{\text{area}}(\zvar_{map};~{\theta})$ to be a useful measure of segmentation quality, it should be high for high Dice scores and conversely low for low Dice scores. The scatter plot between Dice scores and $Q_{\text{area}}(\zvar_{map};~{\theta})$ in \Fig{fig:dice_and_normalizedQ}(right) shows that this is indeed frequently the case. Hence $Q_{\text{area}}(\zvar_{map};~{\theta})$ can serve as a measure of segmentation confidence in the absence of manual segmentations.}

\renewcommand{\tabcolsep}{0pt}
\begin{figure}
\begin{tabular}{cccc}
        \includegraphics[width=0.25\textwidth,frame]{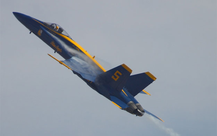} &
        \includegraphics[width=0.25\textwidth,frame]{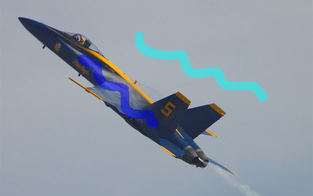} &
        \includegraphics[width=0.25\textwidth,frame]{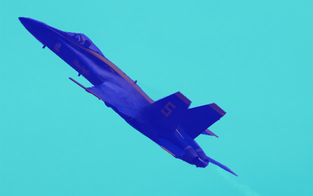} &
        \includegraphics[width=0.25\textwidth,frame]{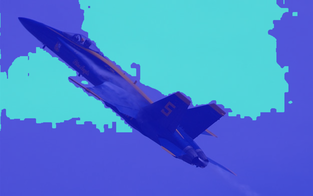} \\
        (a) & (b) & (c) & (d) \\
        \includegraphics[width=0.25\textwidth,frame]{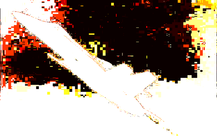} &
        \includegraphics[width=0.25\textwidth,frame]{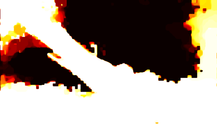} &
        \includegraphics[width=0.25\textwidth,frame]{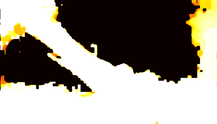} &
        \includegraphics[width=0.25\textwidth,frame]{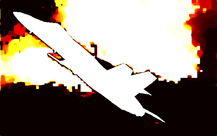} \\
        (e) & (f) & (g) & (h) \\
        \multicolumn{4}{c}{\includegraphics[width=0.4\textwidth]{colorbar_hot_horizontal}}
\end{tabular}
\caption{\mn{Unusual case: Segmentation with high confidence, but low Dice score indicating a segmentation of low quality. (a) Original Image; (b) Seeds to train the random forest; (c) Expert segmentation; (d) AMF segmentation; (e) label probabilities for plane object computed by random forest ; (f) AMF-computed label probabilities for the plane object; (g) masked AMF-computed label probabilities, only showing areas where the plane object is most probable ; (h) AMF-computed label probabilities for the correct expert labels at each location (white image, $\theta=1$ would be a perfect result). As the color values for the plane seeds (dark blue) are similar to regions in the sky, the random forest classifier (e) is overly confident from which the AMF (f) cannot recover. Hence, there is poor overlap between the resulting segmentation (d and h) and the expert labeling (c). At the same time, the overall confidence for this example is high due to the high certainty of the random forest approach. The Dice score for the quasi-multi-label AMF and for the binary AMF is 0.49. The mean $Q_{\text{area}}$ score for both approaches is $0.94$.}}
\label{fig:plane}
\end{figure}

\begin{figure}
\begin{tabular}{cccc}
        \includegraphics[width=0.25\textwidth,frame]{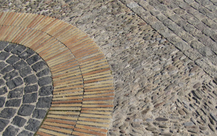} &
        \includegraphics[width=0.25\textwidth,frame]{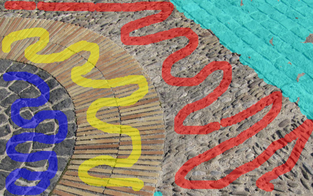} &
        \includegraphics[width=0.25\textwidth,frame]{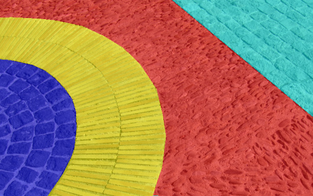} &
        \includegraphics[width=0.25\textwidth,frame]{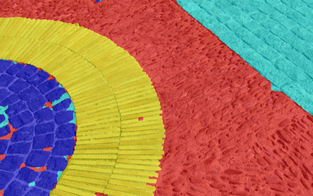} \\
        (a) & (b) & (c) & (d) \\
        \includegraphics[width=0.25\textwidth,frame]{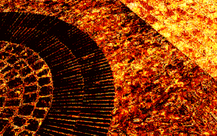} &
        \includegraphics[width=0.25\textwidth,frame]{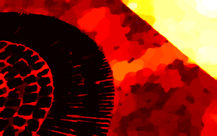} &
        \includegraphics[width=0.25\textwidth,frame]{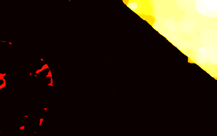} &
        \includegraphics[width=0.25\textwidth,frame]{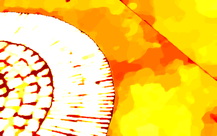} \\
        (e) & (f) & (g) & (h) \\
        \multicolumn{4}{c}{\includegraphics[width=0.4\textwidth]{colorbar_hot_horizontal}}
\end{tabular}
\caption{\mn{Unusual case: Segmentation with low confidence, but high Dice score indicating a segmentation of high quality. (a) Original Image; (b) Seeds to train the random forest; (c) Expert segmentation; (d) AMF segmentation; (e) random forest label probabilities for cobble-stone object on the top-right; (f) corresponding label probabilities computed by AMF; (g) masked AMF-computed label probabilities, only showing areas where the cobble-stone object is the most probable; (h) AMF-computed label probabilities for the correct expert labels at each location (white image, $\theta=1$ would be a perfect result). In this example, the seed points for the cobble-stone object (b; cyan) essentially fully segment the object of interest. However as the color values are ambiguous with respect to the other classes (in particular the red seed label) the overall segmentation result is not highly confident (f  and  g) resulting in a lower $Q_{\text{area}}$ score. However, the most probable labelings also agree with the experts' opinion (d and h). The Dice score for the quasi-multi-label AMF is 0.97 and for the binary AMF 0.93. The mean $Q_{\text{area}}$ scores are 0.71 and 0.67 respectively.}}
\label{fig:floor}
\end{figure}
\renewcommand{\tabcolsep}{6pt}

\begin{figure}
        \begin{tabular}{cccc}
                \includegraphics[width=0.25\textwidth,frame]{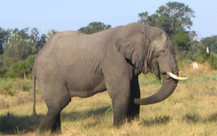} &
                \includegraphics[width=0.25\textwidth,frame]{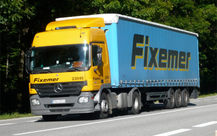} &
                \includegraphics[width=0.25\textwidth,frame]{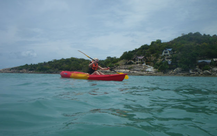} &
                \includegraphics[width=0.25\textwidth,frame]{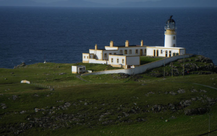} \\
                \includegraphics[width=0.25\textwidth,frame]{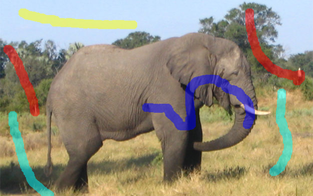} &
                \includegraphics[width=0.25\textwidth,frame]{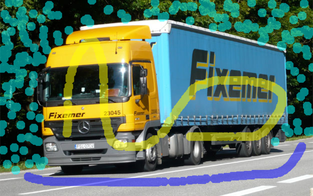} &
                \includegraphics[width=0.25\textwidth,frame]{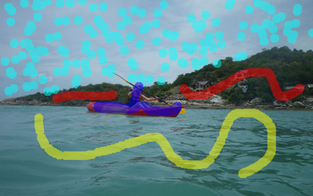} &
                \includegraphics[width=0.25\textwidth,frame]{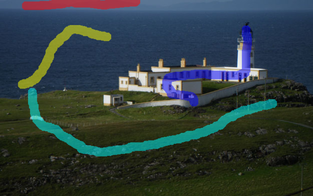} \\
                \includegraphics[width=0.25\textwidth,frame]{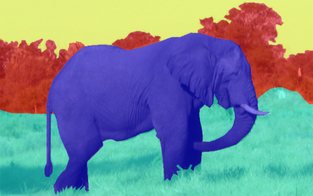} &
                \includegraphics[width=0.25\textwidth,frame]{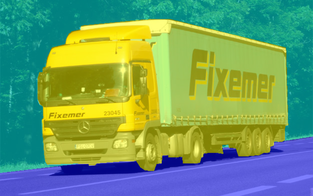} &
                \includegraphics[width=0.25\textwidth,frame]{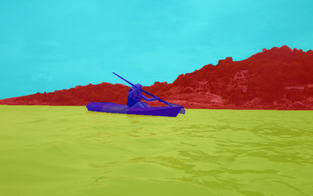} &
                \includegraphics[width=0.25\textwidth,frame]{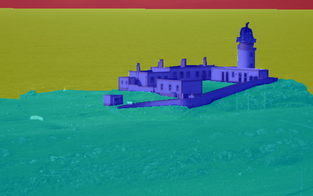} \\
                \includegraphics[width=0.25\textwidth,frame]{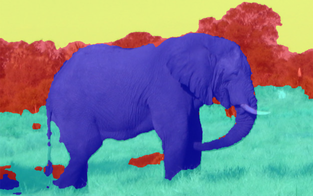} &
                \includegraphics[width=0.25\textwidth,frame]{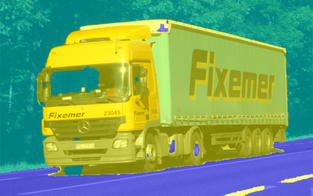} &
                \includegraphics[width=0.25\textwidth,frame]{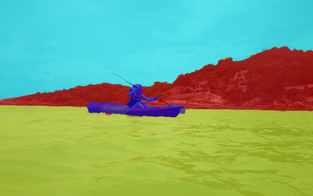} &
                \includegraphics[width=0.25\textwidth,frame]{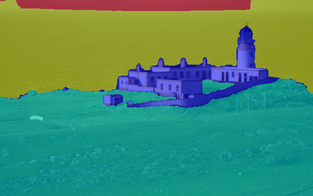} \\
                Dice = 0.95/0.95 & Dice = 0.93/0.93 & Dice = 0.97/0.97 & Dice = 0.93/0.92 \\
                $Q_{\text{area}}=0.94/0.94$ & $Q_{\text{area}}=0.98/0.98$ & $Q_{\text{area}}=0.98/0.97$ & $Q_{\text{area}}=0.97/0.96$ \\
                \includegraphics[width=0.25\textwidth,frame]{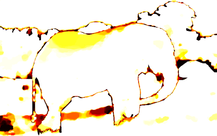} &
                \includegraphics[width=0.25\textwidth,frame]{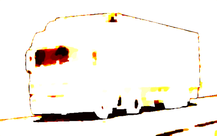} &
                \includegraphics[width=0.25\textwidth,frame]{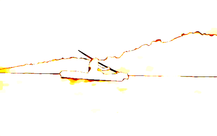} &
                \includegraphics[width=0.25\textwidth,frame]{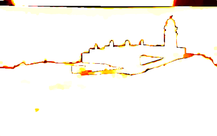} \\
                (a) & (b) & (c) & (d) \\
                \multicolumn{4}{c}{\includegraphics[width=0.4\textwidth]{colorbar_hot_horizontal}}
        \end{tabular}
        \caption{\mn{Sample segmentation results for highly confident high quality segmentations. Top row: original images; 2nd row: seeds used for training the random forest; 3rd row: expert segmentations; 4th row: AMF segmentation result; last row: label probabilities computed  by AMF with respect to the object segmented by the expert (\ie, given an expert label the corresponding probability for that label as computed by the AMF is displayed; a perfect result would be a totally white image). Dice scores for the quasi-multi label approach applied to the AMF segmentation, majority voting (first value), and the mean for all binary segmentations for a given image respectively. $Q_{\text{area}}$ scores are the means over all the segmented objects for the quasi-multi-label approach (first value) and the binary segmentation approach (second value).}}
        \label{fig:successful_segmentations_1}
\end{figure}
\begin{figure}
        \begin{tabular}{cccc}
                \includegraphics[width=0.25\textwidth,frame]{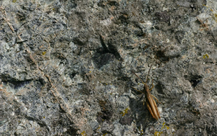} &
                \includegraphics[width=0.25\textwidth,frame]{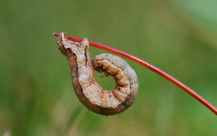} &
                \includegraphics[width=0.25\textwidth,frame]{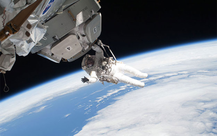} &
                \includegraphics[width=0.25\textwidth,frame]{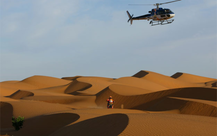} \\
                \includegraphics[width=0.25\textwidth,frame]{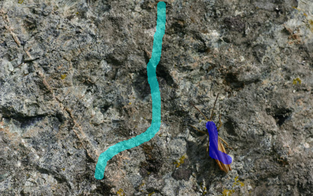} &
                \includegraphics[width=0.25\textwidth,frame]{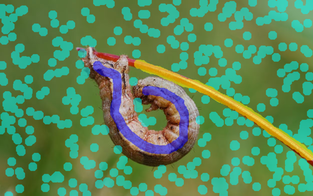} &
                \includegraphics[width=0.25\textwidth,frame]{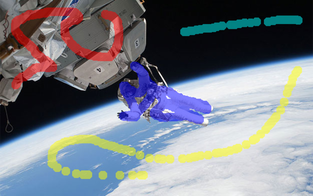} &
                \includegraphics[width=0.25\textwidth,frame]{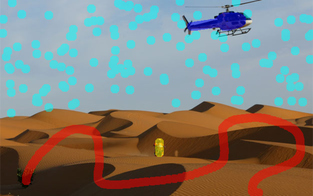} \\
                \includegraphics[width=0.25\textwidth,frame]{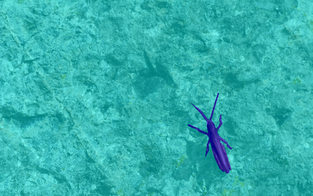} &
                \includegraphics[width=0.25\textwidth,frame]{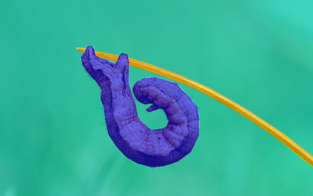} &
                \includegraphics[width=0.25\textwidth,frame]{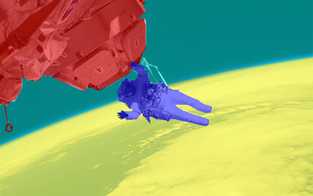} &
                \includegraphics[width=0.25\textwidth,frame]{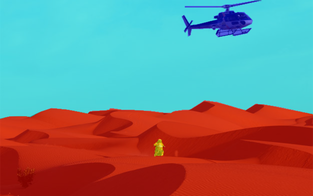} \\
                \includegraphics[width=0.25\textwidth,frame]{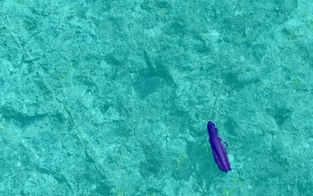} &
                \includegraphics[width=0.25\textwidth,frame]{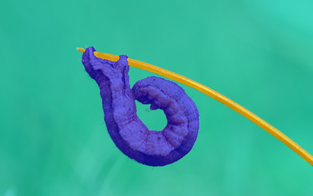} &
                \includegraphics[width=0.25\textwidth,frame]{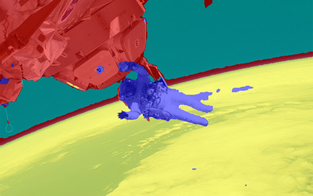} &
                \includegraphics[width=0.25\textwidth,frame]{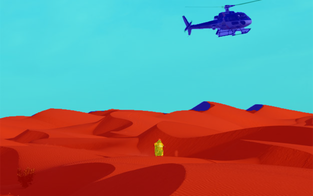} \\
                Dice = 0.90/0.90 & Dice = 0.98/0.98 & Dice = 0.93/0.95 & Dice = 0.94/0.93 \\
                $Q_{\text{area}}=0.95/0.95$ & $Q_{\text{area}}=0.97/0.96$ & $Q_{\text{area}}=0.86/0.85$ & $Q_{\text{area}}=0.86/0.81$ \\
                \includegraphics[width=0.25\textwidth,frame]{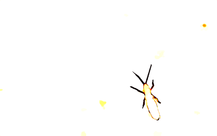} &
                \includegraphics[width=0.25\textwidth,frame]{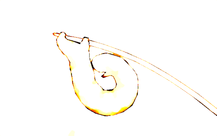} &
                \includegraphics[width=0.25\textwidth,frame]{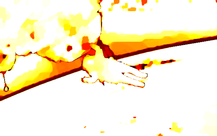} &
                \includegraphics[width=0.25\textwidth,frame]{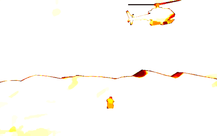} \\
                (a) & (b) & (c) & (d) \\
                \multicolumn{4}{c}{\includegraphics[width=0.4\textwidth]{colorbar_hot_horizontal}}
        \end{tabular}
        \caption{\mn{Sample segmentation results for highly confident high quality segmentations. Top row: original images; 2nd row: seeds used for training the random forest only; 3rd row: expert segmentation; 4th row: AMF segmentation result; last row: label probabilities of AMF with respect to the object segmented by the expert (\ie, given an expert label the corresponding probability for that label as computed by the AMF is displayed; a perfect result would be a totally white image). Dice scores for the quasi-multi label approach applied to AMF segmentation, majority voting (first value) and the mean for all binary segmentations for a given image respectively. $Q_{\text{area}}$ scores are the means over all the segmented objects for the quasi-multi-label approach (first value) and the binary segmentation approach (second value).}}
        \label{fig:successful_segmentations_2}
\end{figure}

\mn{To gain a deeper understanding of the $Q_{\text{area}}$  measure, it is instructive to review cases where $Q_{\text{area}}$ seems unrelated to the Dice score. \Fig{fig:plane} shows a case with very high $Q_{\text{area}}$, but low Dice score, caused by a very confident, but incorrect output of the random forest from which the AMF cannot recover. \Fig{fig:floor} shows a case with very low $Q_{\text{area}}$ but high Dice score. Here, the segmentation is good, but our approach is not confident as other regions have similar color. \Fig{fig:successful_segmentations_1} and \Fig{fig:successful_segmentations_2} show examples where segmentations receive both high Dice and $Q_{\text{area}}$ scores, indicating high quality segmentations which also have high segmentation confidence according to $Q_{\text{area}}$.}

\mn{In summary, the AMF model shows good segmentation performance across a large set of natural images. Furthermore, the posterior distribution on labels carries useful information as it can provide a proxy for likely segmentation quality.}

\renewcommand{\tabcolsep}{0pt}

\renewcommand{\tabcolsep}{6pt}


\section{Conclusions}
\label{SecConclusions}

We described a method for binary image segmentation which allows efficient estimation of  approximate label probabilities through a \mn{ VMF} approximation. We carefully analyzed the theoretical properties of the model and tested its behavior on synthetic and real datasets. A particularly useful feature of our model is that it has strong connections to the Chan-Vese segmentation model and the ROF image-denoising model. \mn{Our method can therefore be  implemented using off-the-shelf solvers of the ROF model. This simple and efficient way to compute solutions makes AMF an attractive alternative to Chan-Vese-like approaches, which, unlike AMF, do not compute posterior \mn{distributions on labels}.} \mn{A current drawback of our method is its binary formulation. Nevertheless, our approach can be used for multi-label segmentation by converting multi-label problems to multiple binary segmentations. A truly multi-label formulation of AMF is outside the scope of this paper, but should be investigated in future work. It will be interesting to see if connections to the Chan-Vese and the ROF model can also be established in a multi-label VMF approach.}
\section{Acknowledgements}
\label{SecAcknowledgements}

This work was supported by NIH grants P41RR013218, P41EB015898, P41EB015902, NIAID R24 AI067039, R01CA111288, R01 HL127661, K05 AA017168 and NSF grants EECS-1148870, EECS-0925875. We also gratefully acknowledge Tassilo Klien and Andrey Fedorov for providing prostate data and probability maps, and Dr. Theodore Abraham for the heart data.

\appendix

\section*{Appendix: Comparison of Approximate and Exact Distributions}

We compare the approximated and exact distributions,
 $Q(\zvar ; \theta)$ and $P(\zvar|y)$, respectively, for general realizations of z.
Because the normalizer for $P(\zvar|y)$ is not available, and for convenience, we will compare
$\ln\frac{P(\zvar|y)}{P(\zvar_0|y)}$ and
$\ln\frac{Q(\zvar;\theta)}{ Q(\zvar_0 ;\theta)}$, \mn{where $\zvar_0$ is the most probable realization under $Q$.}

For calculating the log probability ratio of $P$, we return to the original probability model.
From \mn{\Eqn{eq:likely}} - \Eqn{eq:Bayes-rule},
\begin{align}
   \ln P(\zvar | y) =& \sum_\indx \zvar_i \psi_i - \lambda L(\zvar)  + \const \\ & \approx v^{-1}  \int_{\cX} \zvar(\x) \psi(\x) d\x- \lambda L(\zvar)  + \const.
\end{align}

Then the log probability ratio for the exact posterior, $P$, is
\[
 \ln\frac{P(\zvar|y)}{P(\zvar_0|y)}
 \approx  v^{-1} \int_{\cX} (\zvar(\x) - \zvar_0(\x)) \psi(\x) d\x  - \lambda L(\zvar) + \lambda L(\zvar_0) \PD
\]

Working towards the probability ratio for the AMF approximate posterior, $Q$,
using \mn{\Eqn{eqn:mean-field-finite}}
and using a similar technique,
\[
 \ln Q(\zvar;\theta)\approx   v^{-1} \int_{\cX} \zvar(\x) \phi(\x) dx + \const \PD
\]  Here it is easy to see that the most probable realization under $Q$ is bounded by the zero level-set of $\phi$.

We may now write the log probability ratio for $Q$,
\[
 \ln\frac{Q(\zvar;\theta)}{ Q(\zvar_0 ;\theta)} \approx v^{-1} \int_{\cX} (\zvar(\x) - \zvar_0(\x)) \phi(\x) d\x \PD
\]

\begin{figure}
  \centering
  \includegraphics[width=0.75\columnwidth]{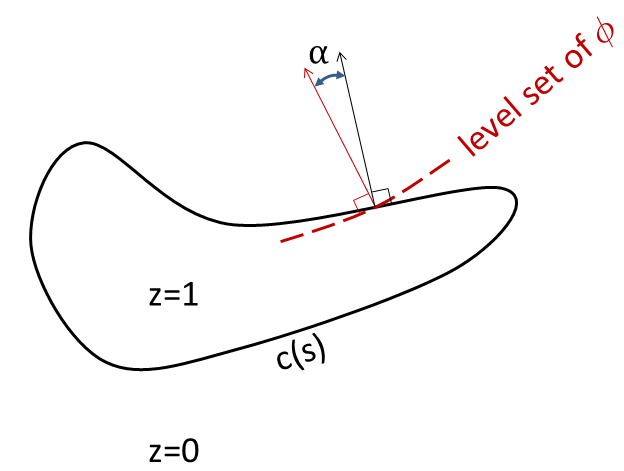}
  \caption{Level sets and normals.}
  \label{FigGeom}
\end{figure}

Subtracting the two probability ratios,
\begin{multline}
  \ln\frac{P(\zvar|y)}{P(\zvar_0|y)} - \ln\frac{Q(\zvar;\theta)}{ Q(\zvar_0 ;\theta)} \\ =
  v^{-1} \int_{\cX} (\zvar(\x) - \zvar_0(\x))(\psi(\x) - \phi(\x)) d\x - \lambda L(\zvar) + \lambda L(\zvar_0) \PD
\end{multline}
We now make use of the AMF equation, $\phi(\x) - \psi(\x) - v \lambda \area \kappa(\phi(\x)) = 0$, to establish relationships among the log probability ratios of $p$ and $q$. We obtain
\begin{eqnarray}
    \ln\frac{P(\zvar|y)}{P(\zvar_0|y)} &-& \ln\frac{Q(\zvar;\theta)}{ Q(\zvar_0 ;\theta)} \\
    &~&\hspace{-2cm}=
    \lambda \left[- \int_{\cX} (\zvar(\x) - \zvar_0(\x))\kappa(\phi(\x))d\x - L(\zvar) + L(\zvar_0) \right]\\
    &~&\hspace{-2cm}=
    \lambda \Big[- \int_{\cX: \zvar(\x) = 1} \nabla \cdot \left(\frac{\nabla \phi(\x)}{|\nabla \phi(\x)|}\right) d\x \\
      &~&\hspace{-2cm}~~~+
     \int_{\cX:\zvar_0(\x) = 1} \nabla \cdot \left(\frac{\nabla \phi(\x)}{|\nabla \phi(\x)|}\right) d\x
    - L(\zvar) + L(\zvar_0) \Big]\\
    &~&\hspace{-2cm}=
    \lambda \Big[- \int_{c(s)} N(\x) \cdot \left(\frac{\nabla \phi(\x)}{|\nabla \phi(\x)|}\right) ds \\
      &~&\hspace{-2cm}~~~+
        \int_{c_0(s)} N(\x)\cdot \left(\frac{\nabla \phi(\x)}{|\nabla \phi(\x)|}\right) ds
    - L(\zvar) + L(\zvar_0) \Big].
\end{eqnarray}
The last two lines use the divergence theorem; $c(s)$ is the boundary of $\zvar(\x)$ oriented so that the outward normal points
from $\zvar(\x) = 1$ towards $\zvar(\x) = 0$, and similarly for $c_0(\x)$ and $\zvar_0(\x)$.
$N(\x)$ is the outward normal vector to the curve in question (see \Fig{FigGeom}).

Then
\opt{arxiv}
{
\begin{multline}
    \ln\frac{P(\zvar|y)}{P(\zvar_0|y)} - \ln\frac{Q(\zvar;\theta)}{ Q(\zvar_0 ;\theta)} \\ =
    \lambda \left[\int_{c(s)} \beta(\x) ds - \int_{c_0(s)} \beta(\x) ds
    - L(\zvar) + L(\zvar_0) \right] \CM
\end{multline}
}
\opt{siims}
{
\begin{equation}
    \ln\frac{P(\zvar|y)}{P(\zvar_0|y)} - \ln\frac{Q(\zvar;\theta)}{ Q(\zvar_0 ;\theta)} =
    \lambda \left[\int_{c(s)} \beta(\x) ds - \int_{c_0(s)} \beta(\x) ds
    - L(\zvar) + L(\zvar_0) \right] \CM
\end{equation}
}
where
\[
  \beta(\x) \doteq  N(\x) \cdot \left(\frac{- \nabla \phi(\x)}{|\nabla \phi(\x)|}\right)
\]
is the dot product of two unit vectors, the outward normal to the curve and the negative of the direction
of the gradient of $\phi$.

On the curve $c_0$, $\beta(\x) = 1$,
because the boundary of $\zvar_0$ is a  level-set of $\phi(\x)$.  In that case
the second and fourth terms cancel.  Re-writing the third term as an integral over $c$,
\opt{arxiv}
{
\begin{multline}
    \ln\frac{P(\zvar|y)}{P(\zvar_0|y)} - \ln\frac{Q(\zvar;\theta)}{ Q(\zvar_0 ;\theta)} =
    \lambda \left[ \int_{c(s)} \beta(\x) ds - \int_{c(s)} 1 ds \right] \\ =
    \lambda  \int_{c(s)} (\beta(\x)- 1) ds \PD
\end{multline}
}
\opt{siims}
{
\begin{equation}
    \ln\frac{P(\zvar|y)}{P(\zvar_0|y)} - \ln\frac{Q(\zvar;\theta)}{ Q(\zvar_0 ;\theta)} =
    \lambda \left[ \int_{c(s)} \beta(\x) ds - \int_{c(s)} 1 ds \right] =
    \lambda  \int_{c(s)} (\beta(\x)- 1) ds \PD
\end{equation}
}
Because $\beta(\x)$ is the dot product of two unit vectors, we may write
$\beta(\x) = \cos(\alpha(\x))$, where $\alpha$ is the angle between the normal to
the curve and the negative of the gradient direction of $\phi(\x)$ (see \Fig{FigGeom}).  Then, using $\cos(\alpha) - 1 = -2 \sin^2(\frac{\alpha}{2})$,
\[
    \ln\frac{P(\zvar|y)}{P(\zvar_0|y)} - \ln\frac{Q(\zvar;\theta)}{ Q(\zvar_0 ;\theta)} =
    -2 \lambda  \int_{c(s)} \sin^2\left(\frac{\alpha(\x)}{2}\right) ds \mn{\PD}
\]

Summarizing the comparison of the probability
ratios of the exact and approximate
distributions, $P$ and $Q$, respectively we see the following:
\begin{itemize}
\item For realizations that are bounded by level-sets of $\phi$, $\alpha$ is zero, so the probability ratios agree.
\item For realizations whose boundaries  are in direction ``close'' to level-sets of $\phi$,
  the probability ratios approximately agree (the disagreement is quadratic in $\alpha$).
\item For curves where $\alpha$ is not small, the probability ratio for $Q$ will be larger
than for $P$, \ie, $Q$ underestimates the length penalty of $P$.
\end{itemize}

We saw above that the zero level-set of $\phi$ is the boundary of the
most probable realization under the approximate distribution,
$Q(\zvar;\theta)$ (and it is unique). \mn{Since the probability ratios agree for $\zvar_0$ (a level set of $\phi$), and the $Q$ ratio upper-bounds the $P$ ratio, we conclude that it is also the boundary of the MAP realization under $P(\zvar|y)$.} In summary, $\zvar_0$, whose boundary is
the zero level-set of $\phi$, satisfies
\[
  \zvar_0 \doteq H(\phi(\x)) = \arg\max_z Q(\zvar;\theta) = \arg\max_z P(\zvar|y) \PD
\]

\bibliographystyle{siam}
\bibliography{references}

\end{document}